\documentclass[10pt,twocolumn,twoside,english]{IEEEtran}

\global\long\def\mat#1{\boldsymbol{#1}}

\global\long\def\optr{\mbox{tr}}

\global\long\def\opdiag{\mbox{diag}}

\newtheorem{theorem}{Theorem}
\newtheorem{proposition}{Proposition}
\newtheorem{lemma}{Lemma}

\newtheorem{example}{Example}

\global\long\def\t#1{\widetilde{#1}}
\global\long\def\h#1{\widehat{#1}}
\global\long\def\abs#1{\left\lvert #1\right\rvert }
\global\long\def\norm#1{\lVert#1\rVert}

\global\long\def\set#1{\left\{  #1\right\}  }
\global\long\def\bydef{\overset{\text{def}}{=}}

\global\long\def\EE{\mathbb{E}\,}
\global\long\def\EEk{\mathbb{E}_{k}\,}

\global\long\def\R{\mathbb{R}}
\global\long\def\E{\mathbb{E}}

\global\long\def\va{\boldsymbol{a}}
\global\long\def\vb{\boldsymbol{b}}
\global\long\def\vc{\boldsymbol{c}}

\global\long\def\vm{\boldsymbol{m}}

\global\long\def\vp{\boldsymbol{p}}
\global\long\def\vq{\boldsymbol{q}}
\global\long\def\vr{\boldsymbol{r}}

\global\long\def\vs{\boldsymbol{s}}

\global\long\def\vu{\boldsymbol{u}}
\global\long\def\vv{\boldsymbol{v}}
\global\long\def\vw{\boldsymbol{w}}
\global\long\def\vx{\boldsymbol{x}}
\global\long\def\vy{\boldsymbol{y}}
\global\long\def\vz{\boldsymbol{z}}

\global\long\def\mA{\boldsymbol{A}}
\global\long\def\mB{\boldsymbol{B}}

\global\long\def\mD{\boldsymbol{D}}
\global\long\def\mE{\boldsymbol{E}}

\global\long\def\mG{\boldsymbol{G}}

\global\long\def\mI{\boldsymbol{I}}
\global\long\def\mJ{\boldsymbol{J}}
\global\long\def\mK{\boldsymbol{K}}
\global\long\def\mL{\boldsymbol{L}}
\global\long\def\mM{\boldsymbol{M}}

\global\long\def\mP{\boldsymbol{P}}
\global\long\def\mQ{\boldsymbol{Q}}
\global\long\def\mR{\boldsymbol{R}}
\global\long\def\mS{\boldsymbol{S}}

\global\long\def\mU{\boldsymbol{U}}

\global\long\def\mV{\boldsymbol{V}}
\global\long\def\mW{\boldsymbol{W}}
\global\long\def\mX{\boldsymbol{X}}

\global\long\def\mZ{\boldsymbol{Z}}

\global\long\def\mLa{\boldsymbol{\Lambda}}
\global\long\def\mOm{\boldsymbol{\Omega}}

\global\long\def\a{\alpha}

\global\long\def\s{\sigma}
\global\long\def\e{\epsilon}

\global\long\def\T{\top}

\global\long\def\nt{\left\lfloor nt\right\rfloor }
\global\long\def\ns{\left\lfloor ns\right\rfloor }
\global\long\def\nT{\left\lfloor nT\right\rfloor }

\global\long\def\Qkn{\mQ_k^{(n)}}
\global\long\def\vqn{\vq^{(n)}}
\global\long\def\tvqn{\t \vq^{(n)}}
\global\long\def\hvqn{\h \vq^{(n)}}
\global\long\def\vrn{\vr^{(n)}}
\global\long\def\vmn{\vm^{(n)}}
\global\long\def\PP{\mathbb{P}}

\global\long\def\normF#1{\norm{#1}_{\text F}}

\usepackage{dsfont}
\newcommand{\charfn}{\mathds{1}}

\usepackage{xr}
\usepackage{cite}
\usepackage[T1]{fontenc}
\usepackage[latin9]{inputenc}
\usepackage{mathtools}
\usepackage{amsmath}
\usepackage{amssymb}
\usepackage{commath}
\usepackage{latexsym}
\let\labelindent\relax
\usepackage{enumitem}
\usepackage{algorithmic}
\usepackage{algorithm}
\usepackage{subfigure}
\makeatletter

\providecommand{\eref}[1]{\eqref{eq:#1}}  
\providecommand{\sref}[1]{Section~\ref{sec:#1}}
\providecommand{\fref}[1]{Figure~\ref{fig:#1}}

\pdfpageheight\paperheight
\pdfpagewidth\paperwidth

\newcommand{\cip}{\overset{\mathcal{P}}{\longrightarrow}}
\newcommand{\cw}{\overset{\text{weakly}}{\longrightarrow}}

\usepackage{color}
\definecolor{hellgelb}{rgb}{1,1,0.85}
\definecolor{colKeys}{rgb}{0,0,1}
\definecolor{colIdentifier}{rgb}{0,0,0}
\definecolor{colComments}{rgb}{1,0,0}
\definecolor{colString}{rgb}{0,0.5,0}

\DeclareMathOperator*{\argmin}{arg\,min}
\DeclareMathOperator*{\argmax}{arg\,max}
\DeclareMathOperator{\diag}{diag}

\makeatother

\usepackage{babel}

\begin{document}

\title{Subspace Estimation from Incomplete Observations: A High-Dimensional Analysis}

\author{\IEEEauthorblockN{Chuang Wang,
Yonina C. Eldar,~\IEEEmembership{Fellow,~IEEE}
and Yue M. Lu,~\IEEEmembership{Senior Member,~IEEE}}
\thanks{C. Wang is with the John A. Paulson School of Engineering and Applied Sciences, Harvard University, Cambridge, MA 02138, USA (e-mail: chuangwang@g.harvard.edu).}
\thanks{Y. C. Eldar is with the Department of EE, Technion, Israel Institute of Technology, Haifa, 32000, Israel (e-mail: yonina@ee.technion.ac.il).}
\thanks{Y. M. Lu is with the John A. Paulson School of Engineering and Applied Sciences, Harvard University, Cambridge, MA 02138, USA (e-mail: yuelu@seas.harvard.edu). }
\thanks{The work of C. Wang and Y. M. Lu was supported in part by the US Army Research Office under contract W911NF-16-1-0265 and in part by the US National Science Foundation under grants CCF-1319140 and CCF-1718698. The work of Y. Eldar was supported in part by the European Union's Horizon 2020 Research and Innovation Program under Grant 646804-ERCCOG-BNYQ. Preliminary results of this work was presented at the Signal Processing with Adaptive Sparse Structured Representations (SPARS) workshop in 2017.}
}

\maketitle
\begin{abstract}
We present a high-dimensional analysis of three popular
algorithms, namely, Oja's method, GROUSE and PETRELS, for subspace estimation from streaming and highly incomplete observations.  We show that, with proper time scaling, the time-varying
 principal angles between the true subspace and its estimates given by the algorithms
 converge weakly to deterministic processes when the ambient dimension $n$ tends to infinity. Moreover, the limiting processes can be exactly characterized as the unique solutions of certain ordinary differential equations (ODEs). A finite sample bound is also given, showing that the rate of convergence towards such limits is $\mathcal{O}(1/\sqrt{n})$. In addition to providing asymptotically exact predictions of the dynamic performance of the algorithms, our high-dimensional analysis yields several insights, including an asymptotic equivalence between Oja's method and GROUSE, and a precise scaling relationship linking the amount of missing data to the signal-to-noise ratio. By analyzing the solutions of the limiting ODEs, we also establish phase transition phenomena associated with the steady-state performance of these techniques.


\end{abstract}



\section{Introduction}

Subspace estimation is a key task in many signal processing applications. Examples include source localization in array processing, system identification, network monitoring, and image sequence analysis, to name a few. The ubiquity of subspace estimation comes from the fact that a low-rank subspace model can conveniently capture the intrinsic, low-dimensional structures of many large datasets.

In this paper, we consider the problem of estimating and tracking an unknown subspace from \emph{streaming measurements} with many \emph{missing entries}. The streaming setting appears in applications (\emph{e.g.} video surveillances) where high-dimensional data arrive sequentially over time at high rates. It is especially relevant in dynamic scenarios where the underlying subspace to be estimated can be time-varying. Missing data is also  very common  in practice. Incomplete observations may result from a variety of reasons, such as the limitations of the sensing mechanisms, constraints on power consumption or communication bandwidth, or a deliberate design feature that protects  privacy of individuals by removing partial records.

GROUSE \cite{Balzano:2010} and PETRELS \cite{Chi2013} as well as the classical 
Oja's method \cite{Oja1982} are three popular algorithms for solving the subspace estimation problem. They are all streaming algorithms in the sense that they provide instantaneous, \emph{on-the-fly} updates to their subspace 
estimates upon the arrival of a new data point. 
The three differ in their update rules: Oja's method and GROUSE perform 
first-order incremental gradient descent on the Euclidean space and 
the Grassmannian, respectively, whereas PETRELS can be interpreted as 
a second-order stochastic gradient descent scheme. These algorithms 
have been shown to be highly effective in practice, but their performance 
depends on careful choice of algorithmic parameters such as 
the step size (for GROUSE and Oja's method) and the discount parameter (for PETRELS). 
Various convergence properties of these techniques have been studied in 
\cite{Chi2013, Balzano:2015, ZhangB:2016, Zhang2015,gonen2016subspace}, 
but a precise analysis of their performance is still an open problem. Moreover, the important question of how 
the signal-to-noise ratios (SNRs), the amount of missing data, and various other algorithmic parameters affect the estimation performance is not fully understood.

As the main objective of this work, we present a tractable and asymptotically exact analysis of the dynamic performance of 
Oja's method, GROUSE and PETRELS in the high-dimensional regime. Our contribution is mainly threefold:

1. \emph{Precise analysis via scaling limits.} We show in Theorem~\ref{thm:ode-d1} and Theorem~\ref{thm:limit} that the time-varying trajectories of the estimation errors, measured in terms of the principal angles between the true underlying subspace and the estimates given by the algorithms, converge weakly to deterministic processes, as the ambient dimension $n \to \infty$. Moreover, these deterministic limits can be characterized as the unique solutions of certain ordinary differential equations (ODEs). In addition, we provide a finite-size guarantee in Theorem~\ref{thm:finite}, showing that the convergence rate towards the limits is $\mathcal{O}(1/\sqrt{n})$. Numerical simulations verify the accuracy of our asymptotic predictions. The main technical tool behind our analysis is the weak convergence theory of stochastic processes (see \cite{Meleard:1987, Sznitman:1991, EthierK:85, billingsley2013convergence, Jacod:2010} for  mathematical foundations and \cite{WangL:16, Wang2017, Wang2017c} for recent applications in related estimation problems).

2. \emph{Insights regarding the algorithms.} In addition to providing asymptotically exact predictions of the dynamic performance of the three subspace estimation algorithms, our high-dimensional analysis leads to several insights. First, the result of Theorem~\ref{thm:ode-d1} implies that, despite their different update rules, Oja's methods and GROUSE are asymptotically equivalent, with both converging to the \emph{same} deterministic process as the dimension increases. Second, the characterization given in Theorem~\ref{thm:limit} shows that PETRELS can be examined within a common framework that incorporates all three algorithms, with the difference being that PETRELS uses an adaptive scheme to adjust its effective step sizes. Third, our limiting ODEs also reveal an (asymptotically) exact scaling relationship that links the amount of missing data to the SNR. See the discussion in \sref{insights} for details.

3. \emph{Fundamental limits and phase transitions.} Analyzing the 
limiting ODEs also reveals phase transition phenomena associated with the steady-state performance of these algorithms. Specifically, we provide in Propositions~\ref{prop:phase_Oja_GROUSE} and \ref{prop:phase_PETRELS} critical thresholds for setting key algorithm parameters (as a function of the SNR and the subsampling ratio), beyond which the algorithms converge to ``noninformative'' estimates that are no better than mere random guesses.

The rest of the paper is organized as follows. We start by presenting in \sref{model} the exact problem formulation for subspace estimation with missing data. This is followed by a brief review of the three algorithms to be analyzed in this work. The main results are presented in \sref{main_results}, where we show that the dynamic performance of Oja's method, GROUSE and PETRELS can be asymptotically characterized by the solutions of certain deterministic systems of ODEs. Numerical experiments are also provided to illustrate and verify our theoretical predictions. To place our asymptotic analysis in proper context, we discuss related work in the literature in \sref{related}. We consider various implications and insights drawn from our analysis in \sref{implications}. Due to space limitation, we only present informal derivations of the limiting ODEs and proof sketches in \sref{formal}. More technical details and the proofs of all the results presented in this paper can be found in the Supplementary Materials \cite{WangEL:18a}.

\emph{Notation}: Throughout the paper, we use $\mI_d$ to denote the $d \times d$ identity matrix. For any positive semidefinite matrix $\mM$, its principal squared root is written as $\mM^{-\frac{1}{2}}$. Depending on the context, $\norm{\cdot}$ denotes either the $\ell_2$ norm of a vector or the spectral norm of a matrix. For any $x \in \R$, the floor operation $\lfloor x \rfloor$ gives the largest integer that is smaller than or equal to $x$. Let $\set{X_n}$ be a sequence of random variables in a general probability space. $X_n \cip X$ means that $X_n$ converges in probability to a random variable $X$, whereas $X_n \cw X$ means that $X_n$ converges to $X$ weakly (\emph{i.e.} in law). Finally, $\charfn_\mathcal{A}$ denotes the indicator function for an event $\mathcal{A}$.


\section{Problem Formulation and Overview of Algorithms}
\label{sec:alg}


\subsection{Observation Model}
\label{sec:model}

We consider the problem of estimating a low-rank subspace using partial observations from a data stream. At any discrete-time $k$, suppose that a sample vector $\vs_k \in \R^n$ is generated according to 
\begin{equation} \label{eq:gen}
\vs_k = \mU \vc_k + \va_k.
\end{equation}
Here, $ \mU \in \R^{n \times d}$ is an unknown deterministic matrix whose columns 
form an {\em orthonormal} basis of a $d$-dimensional subspace, and $\vc_k \in \R^d$ is a random vector representing the expansion coefficients in that subspace. We further assume\footnote{The assumption that the covariance matrix is diagonal can be made without loss of generality, after a rotation of the coordinate system. To see that, suppose $\vc_k$ has a general covariance matrix $\mat{\Sigma}$, which is diagonalized as $\mat{\Sigma} = \mat{\Phi} \mLa \mat{\Phi}^\top$. Here, $\mat{\Phi}$ is an orthonormal matrix and $\mLa$ is a diagonal matrix as in \eref{def-Lambda}. The generating model \eref{gen} can then be rewritten as $\vs_k = (\mU \mat{\Phi}) (\mat{\Phi}^\top \vc_k) + \va_k$. Thus, our problem is equivalent to estimating a subspace spanned by $\mU \mat{\Phi}$, and $\mLa$ is the covariance matrix of the new expansion coefficient vector $\mat{\Phi}^\top \vc_k$.} that the covariance matrix of $\vc_k$ is diagonal:
\begin{equation}\label{eq:def-Lambda}
\mLa^2= \opdiag( \lambda_{1}^2, \lambda_{2}^2, \ldots, \lambda_{d}^2),
\end{equation}
where $\lambda_1 \ge \lambda_2 \ge \cdots \ge \lambda_d$ are some strictly positive numbers. The noise in the observations is modeled by a random vector $\va_k \in \R^n$ with zero mean and  covariance matrix equal to $\s^2\mI_n$. Furthermore, $\va_k$ is independent of $\vc_k$. Since $\set{\lambda_\ell}_{1 \le \ell \le d}$ in \eref{def-Lambda} indicate the ``strength'' of the subspace components relative to the noise $\va_k$ whose variance is $\s^2$, we refer to $ \lambda_\ell/\s$ as the signal-to-noise ratio (SNR) for the $\ell$th component of the subspace in our subsequent discussions.

We consider the missing data case, where only a subset of the entries of $\vs_k$ is available. This observation process can be modeled by a diagonal matrix
\begin{equation}
\mOm_{k} = \diag(v_{k,1},v_{k,2}, 
\ldots,v_{k,n}),\label{eq:def-Omega}
\end{equation}
where $v_{k,i}=1$ if the $i$th component of $ \vs_{k}$ is observed, and $v_{k,i}=0$ otherwise. Our actual observation, denoted by ${\vy}_k$, may then be written as 
\begin{equation}\label{eq:subsample}
{\vy}_k = \mOm_{k} \vs_k.
\end{equation}
Given a sequence of incomplete observations 
$\set{\vy_k, \mOm_k}_{k \ge 0}$ arriving in a stream, we aim to estimate 
the subspace spanned by the columns of $\mU$. 


\subsection{Oja's Method}

Oja's method \cite{Oja1982} is a classical algorithm for estimating low-rank subspaces from streaming samples. It was originally designed for the case where the full sample vectors $\vs_k$ in \eref{gen} are available. Given a collection of $K$ such sample vectors, it is natural to use the following optimization formulation to estimate the unknown subspace:
\begin{align}
\widehat{\mU} &=   \argmin_{\mX^{ \T} \mX= \mI_{d}} \ \sum_{k=1}^{K} \min_{\vw_k} \, \norm{ \vs_{k}- \mX \vw_k}^{2} \label{eq:pca1} \\
 &=  \argmax_{ \mX^{ \T} \mX= \mI_{d}}  \  \sum_{k=1}^{K} \optr\big( \mX^{ \T} \vs_{k} \vs_k^\T \mX \big) , \label{eq:pca-2}
\end{align}
where the equivalence between \eref{pca1} and \eref{pca-2} is established by solving the simple quadratic problem $\min_{\vw_k} \, \norm{ \vs_{k}- \mX \vw_k}^{2}$ and substituting the solution into \eref{pca1}.

%

Oja's method is a stochastic projected-gradient algorithm for solving
(\ref{eq:pca-2}). At each step $k$, let $\mX_k$ denote the current estimate of the subspace. Then, with the arrival of a new sample vector $\vs_{k}$, we first update $ \mX_{k}$ according to 
\begin{equation}
\t{ \mX}_{k}= \mX_{k}+ \frac{ \tau_{k}}{n} \vs_{k} \vw_{k}^{ \T}, \label{eq:Oja-1}
\end{equation}
where $\vw_{k}= \mX_{k}^{ \T} \vs_{k}$ and $\set{\tau_k}$ is a sequence of positive constants that control the step-size (or learning rate) of the algorithm. We note that, up to a scaling constant, $\vs_k \vw_k^T$ in \eref{Oja-1} is exactly equal to the gradient of the objective function $\mX^{ \T} \vs_{k} \vs_k^\T \mX$ in \eref{pca-2} due to the new sample $\vs_k$. Next, to enforce the orthogonality constraint, we compute
\begin{equation} \label{eq:norm}
\mX_{k+1}=  \t{\mX}_k ( \t{\mX}_k^\T  \t{\mX}_k)^{-\frac{1}{2}},
\end{equation}
where $(\cdot)^{-\frac{1}{2}}$ stands for the principal square root of a positive semidefinite matrix. 
In practice, \eqref{eq:norm} is implemented using the QR-decomposition of $\t{\mX}_k$.

To handle the case of partially-observed samples, we can modify Oja's method in two ways \cite{Balzano2018}. First,  we estimate the expansion coefficients $\vw_{k}$ in \eref{Oja-1} by solving a least squares problem that takes into account the missing data model:
\begin{equation}
\widehat{ \vw}_{k}= \argmin_{ \vw \in \R^d} \ \norm{ {\vy}_{k}- \mOm_{k} \mX_{k} \vw }^2, \label{eq:w-hat}
\end{equation}
where ${\vy}_k$ is the incomplete sample vector defined in \eref{subsample}, $\mOm_k$ is the corresponding subsampling matrix, and $\mX_k$ is the current estimate of the subspace. 
Next, we replace the missing elements in $ \vy_k$ by the corresponding entries in $ \mX_{k} \widehat{ \vw}_{k}$. This \emph{imputation} step leads to an estimate of the full vector:
\begin{equation}
\widehat{\vy}_k = {\vy}_k + (\mI_n - \mOm_k)\mX_k \widehat{\vw}_k.
\end{equation}
Replacing the original vectors $\vs_k$ and $\vw_k$ in \eref{Oja-1} by their estimated counterparts $\widehat{\vy}_k$ and $\widehat{\vw}_k$ leads to the modified Oja's method, a pseudocode of which is summarized in Algorithm \ref{alg:Oja}. 

To ensure that we have enough observed entries in $\vy_k$, we first check, with the arrival of a new partially observed vector $\vy_k$, whether
\begin{equation}\label{eq:eps_check}
\lambda_{\min}(\mX_k^\T \mOm_k \mX_k) > \epsilon ,
\end{equation}
where $\epsilon > 0$ is a small positive constant. If this is indeed the case, we then perform the standard update as described above; otherwise, we ignore the new sample vector and do not change the estimate in this step. Note that, under a suitable probabilistic model for the subsampling process (see assumption~\ref{ass:iid} in \sref{thms}), one can show that \eref{eps_check} is satisfied with high probability as long as $\epsilon < \alpha$, where $\alpha$ denotes the subsampling ratio defined in assumption~\ref{ass:iid}.



\begin{algorithm}[t] 
\caption{\label{alg:Oja} Oja's method with imputation \cite{Oja1982,Balzano2018}}
\begin{algorithmic}[1]
\REQUIRE An initial estimate $\mX_0$ such that $\mX_0^\T \mX_0=\mI_d$, a sequence of step-size parameters $\set{\tau_k}$ and a positive constant $\epsilon$.
\STATE $k \coloneqq 0$
\REPEAT
	\IF{$\lambda_{\min} (\mX_k^\T \mOm_k \mX_k) > \epsilon$}
	\STATE  
    $\widehat\vw_k \coloneqq \argmin_{\vw} \norm{{\vy}_k - \mOm_k \mX_k \vw }^2$
    \STATE   
    $\widehat{\vy}_k \coloneqq {\vy}_k + (\mI_n - \mOm_k)\mX_k \widehat\vw_k$
    \STATE
    $\t{\mX}_k \coloneqq \mX_k + \frac{\tau_k}{n}  
    \h{\vy}_k \widehat\vw_k^\T $
   	 \STATE
    	$\mX_{k+1}\coloneqq \t{\mX}_k ( \t{\mX}_k^\T  \t{\mX}_k)^{-\frac{1}{2}}$ 
    \ELSE
     \STATE $\mX_{k+1}\coloneqq\mX_k$
    \ENDIF
    \STATE $k\coloneqq k+1$
    
\UNTIL{termination}
\end{algorithmic}
\end{algorithm}

\subsection{GROUSE}
Similar to Oja's method, Grassmannian Rank-One Update Subspace Estimation (GROUSE) \cite{Balzano:2010} is a first-order stochastic
gradient descent algorithm for solving (\ref{eq:pca1}). The main difference is that GROUSE solves the optimization problem on the Grassmannian,
the manifold of all subspaces with a fixed rank. One advantage of this approach is that it avoids the explicit orthogonalization step in \eref{norm}, allowing the algorithm to achieve  lower computational complexity.

At each step, GROUSE first finds the coefficient
$ {\widehat \vw}_{k}$ according to (\ref{eq:w-hat}). It then computes the reconstruction
error vector 
\begin{equation}
\begin{aligned}
\vr_{k}=  \vy_{k}- \mOm_{k} \vp_{k}. \label{eq:def-r}
\end{aligned}
\end{equation}
Here,  $\vy_k$ and $\mOm_k$ are defined in \eqref{eq:gen} and \eqref{eq:def-Omega}, and
\begin{equation}
 \vp_{k} = \mX_{k} { \widehat\vw}_{k}, \label{eq:def-p}
 \end{equation} where $\widehat\vw_k$ is defined in  \eqref{eq:w-hat}. Next, it updates the
current estimate $ \mX_{k}$ on the Grassmannian as
\[
\mX_{k+1}=  \mX_{k}+\left[\frac{\big(\cos(\theta_{k})-1\big) \vp_{k}}{\norm{\vp_k}}+\frac{\sin(\theta_{k}) \vr_k}{\norm{\vr_k}}\right] \frac{\vw^\T_k}{\norm{\vw_k}},
\]
where
\begin{equation}
\theta_{k}= \frac{ \tau_{k}}{n} \norm{ \vr_{k}} \cdot \norm{ \vp_{k}}, \label{eq:def-h}
\end{equation}
and $\set{\tau_k}_k$ is a sequence of step-size parameters. The algorithm is summarized in Algorithm \ref{alg:grouse}. 


\begin{algorithm}[t]
\caption{\label{alg:grouse} GROUSE \cite{Balzano:2010}}
\begin{algorithmic}[1]
\REQUIRE An initial estimate $\mX_0$ such that $\mX_0^\T \mX_0=\mI_d$, a sequence of step-size parameters $\set{\tau_k}$ and a positive constant $\epsilon$.
 \STATE $k \coloneqq 0$
\REPEAT
\IF{$\lambda_{\min} (\mX_k^\T \mOm_k \mX_k) > \epsilon$} \label{ln:check-grouse}
	\STATE  
    $\widehat\vw_k \coloneqq \argmin_{\vw} \norm{{\vy}_k - \mOm_k \mX_k \vw }^2$
    \STATE
    $\vp_k \coloneqq \mX_k \widehat\vw_k$
    \STATE $\vr_k \coloneqq {\vy}_k -  \mOm_k \vp_k $
    \STATE $\theta_k \coloneqq \frac{\tau_k}{n} \norm{\vr_k} \cdot \norm{\vp_k}$
    \STATE  $
    \mX_{k+1} \coloneqq \mX_k + 
    \Big[
    \frac{(\cos(\theta_k)-1)  \vp_k}{\norm{\vp_k}} 
    + \frac{ \sin(\theta_k) \vr_k}{\norm{\vr_k}} \Big] \frac{\widehat\vw_k^\T}
    {\norm{\widehat\vw_k}}
    $
   \ELSE
     \STATE $\mX_{k+1}\coloneqq\mX_k$
    \ENDIF
    \STATE $k\coloneqq k+1$
    
\UNTIL{termination}
\end{algorithmic}
\end{algorithm}

\subsection{PETRELS}

When there is no missing data, an alternative to Oja's method is a classical algorithm called Projection Approximation Subspace Tracking (PAST) \cite{Yang1995}. This method estimates the underlying subspace $\mU$ by solving an exponentially-weighted least-squares problem
\begin{equation}
\mX_{k+1}= \argmin_{ \mX \in \R^{n \times d}} \ 
\sum_{ k^\prime=1}^{k} \gamma^{k- k^\prime} 
\norm{ \vs_{k^\prime}- \mX \vw_{k^\prime}}^{2}, \label{eq:past}
\end{equation}
where $\vw_{k^\prime} = \mX_{k^\prime}^T \vs_{k^\prime}$ and $\gamma \in(0,1]$ is a discount parameter. The solution of (\ref{eq:past}) has a simple recursive update rule
\begin{align} 
\mX_{k+1}
&= \mX_k +
\left( \vs_k - \mX_k \vw_k \right) 
\vw_k^\T \mR_k \label{eq:update-PAST}\\
\mR_{k+1} &= (\gamma \mR_k^{-1} + \vw_k \vw_k^\T)^{-1}. \label{eq:update-R}
\end{align}
Moreover, one can avoid the explicit calculation of the matrix inverse in \eref{update-R} by using the Woodbury identity and the fact that \eqref{eq:update-R} amounts to a rank-$1$ update.

Parallel Subspace Estimation and Tracking by Recursive Least Squares (PETRELS) \cite{Chi2013} extends PAST to the case of partially-observed data. The main change is that it estimates the coefficient $ \vw_{k}$ in
(\ref{eq:past}) using \eqref{eq:w-hat}. In addition, it provides a parallel sub-routine in its calculations so that updates to different coordinates can be computed in a fully parallel fashion. In its most general form, PETRELS needs to maintain and update a different $d \times d$  matrix $\mR_k^i$ for each of the $n$ coordinates. To reduce computational complexity, a simplified version of PETRELS is provided in \cite{Chi2013}, using a common $\mR_k$ for all the coordinates. 

In this paper, we focus on this simplified version of PETRELS, which is summarized in Algorithm \ref{alg:sim-petrels}. Note that we introduce an additional parameter $\a$ in lines \ref{ln:beta} and \ref{ln:update-R} of the pseudocode. The simplified algorithm given in \cite{Chi2013} corresponds to setting $\alpha = 1$. In our analysis, we set $\a$ to be equal to the subsampling ratio defined later in \eref{alpha}. Empirically, we find that, with this modification, the performance of the simplified algorithm matches that of the full PETRELS algorithm when the ambient dimension $n$ is large.

Finally, we note that the computational complexity per iteration of all three algorithms is $\mathcal{O}(nd^2)$. The most expensive step is  the estimation of the missing data. One main contribution of this work is to precisely predict the asymptotic ($n \to \infty$) performance of the algorithms after $k = \lfloor tn \rfloor$ iterations, for any $t > 0$. 

%
%
%
%
%
%

\begin{algorithm}[t]
\caption{\label{alg:sim-petrels} Simplified PETRELS \cite{Chi2013}}
\begin{algorithmic}[1]
\REQUIRE An initial estimate of the subspace $\mX_0$, $\mR_0= \frac{\delta}{n} \mI_d$ for some $\delta > 0$, and positive constants $\gamma$ and $\epsilon$.
\STATE $k \coloneqq 0$
\REPEAT
	 
	\IF {$\lambda_{\min} (\mX_k^\T \mOm_k \mX_k) > \epsilon$} \label{ln:check}
   	 	\STATE  $\widehat\vw_k \coloneqq \argmin_{\vw} \norm{{\vy}_k - \mOm_k \mX_k \vw }^2$
       		\STATE $\mX_{k+1} \coloneqq \mX_{k} + \mOm_{k} ({\vy}_k - \mX_k \widehat\vw_k ) \widehat\vw_k^\T \mR_{k}$

        		\STATE $\vv_{k} \coloneqq \gamma^{-1} \mR_{k} \widehat\vw_k$
        		\STATE $\beta_{k} \coloneqq 1+ \a\, \widehat\vw_k^\T \vv_{k}$  \label{ln:beta}
        		\STATE $\mR_{k+1} \coloneqq \gamma^{-1} \mR_{k} - \a\,
        			\vv_{k} \vv_{k}^\T /\beta_{k}$ \label{ln:update-R}
		 \label{ln:update-X}
        \ELSE
		\STATE $\mX_{k+1} \coloneqq\mX_k$
		\STATE $\mR_{k+1} \coloneqq \mR_k$

        \ENDIF

    \STATE $k\coloneqq k+1$
    
\UNTIL{termination}
\end{algorithmic}
\end{algorithm}


%
%


\section{Main Results: Scaling Limits}
\label{sec:main_results}

In this section, we present the main results of this work---a tractable and asymptotically exact analysis of the performance of the three algorithms reviewed in \sref{alg}.  

\subsection{Performance Metric: Principal Angles}

We start by defining the performance metric we will be using in our analysis. Recall the generative model defined in \eqref{eq:gen}. The ground truth subspace is represented by the matrix $\mU$, whose column vectors form 
an orthonormal basis of that subspace. For Algorithms \ref{alg:Oja}, \ref{alg:grouse}, and \ref{alg:sim-petrels}, the estimated subspace at the $k$th step is spanned by an orthogonal matrix
\begin{equation}\label{eq:Uhat}
\widehat{\mU}_k\bydef \mX_k (\mX_k^\T \mX_k)^{-1/2},
\end{equation}
where $\mX_k$ is the $k$th iterand generated by the algorithms. Note that, for Oja's method and GROUSE, $\widehat{\mU}_k= \mX_k$ as the matrix $\mX_k$ is already orthogonal, whereas for PETRELS, generally $\mX_k^\T \mX_k\neq\mI_d$ and thus the step in \eref{Uhat} is necessary.

In the special case of $d = 1$ (\emph{i.e.} rank-one subspaces), both $\mU$ and $\widehat{\mU}_k$ are unit-norm vectors. The degree to which these vectors are aligned can be measured by their \emph{cosine similarity}, defined as $\abs{\mU^\T \widehat{\mU}_k}$. This concept can be naturally extended to arbitrary $d \ge 1$. In general, the closeness of two $d$-dimensional subspaces may be quantified by their $d$ principal angles \cite{Ipsen1995, Deutsch1995}. In particular, the cosines of the principal angles are uniquely specified as the singular values of the $d \times d$ matrix
\begin{equation} \label{eq:def-Q}
\mQ_{k}^{(n)} \bydef \mU^{ \T} \widehat{\mU}_{k}=\mU^{ \T} \mX_k (\mX_k^\T \mX_k)^{-1/2}.
\end{equation}
In what follows, we shall refer to $\mQ_k^{(n)}$ as the \emph{cosine similarity matrix}. Since we will be studying the high-dimensional limit of $\mQ_k^{(n)}$ as the ambient dimension $n \to \infty$, we use the superscript $(n)$ to make the dependence of $\mQ_{k}^{(n)}$ on $n$ explicit. 

%

\subsection{The Scaling Limits of Stochastic Processes: Main Ideas}
\label{sec:toy_example}

To analyze the performance of Algorithms \ref{alg:Oja}, \ref{alg:grouse}, and \ref{alg:sim-petrels}, our goal boils down to tracking the evolution of the cosine similarity matrix $\mQ_k^{(n)}$ over time. Thanks to the streaming nature of all three methods, it is easy to see that the dynamics of their estimates $\mX_k$ can be modeled by homogeneous Markov chains with state space in $\R^{n \times d}$. Thus, being a function of $\mX_k$ [see \eref{def-Q}], the dynamics of $\Qkn$ forms a hidden Markov chain. We then show that, as $n \to \infty$ and with proper time scaling, the family of stochastic processes $\set{\Qkn}$ indexed by $n$ converges \emph{weakly} to a deterministic function of time that is characterized as the unique solution of some ODEs. Such convergence is known in the literature as the \emph{scaling limits} \cite{EthierK:85, Billingsley:1999, Jacod:2010, Wang2017c} of stochastic processes. To present our results, we first consider a simple one-dimensional example that illustrates the underlying ideas behind scaling limits. Our main convergence theorems are presented in \sref{thms}.

Consider a 1-D stochastic process defined by a recursion
\begin{equation}\label{eq:1D}
q_{k+1} = q_k + \tfrac{\tau}{n} f(q_k) + \tfrac{1}{n^{(1/2) + \delta}} v_k,
\end{equation}
where $f(\cdot)$ is a Lipschitz function, $\tau$ and $\delta$ are two positive constants, $v_k$ is a sequence of i.i.d. random variables with zero mean and unit variance, and $n > 0$ is a constant introduced to scale the step-size and the noise variance. (This particular scaling is chosen here because it mimics the actual scaling that appears in the high-dimensional dynamics of $\Qkn$ we shall study.)

When $n$ is large, the difference between $q_k$ and $q_{k+1}$ is small. In other words, we will not be able to see \emph{macroscopic} changes unless we observe the process over a large number of steps. To accelerate the time (by a factor of $n$), we embed $\set{q_k}$ in continuous-time by defining a piecewise-constant process
\begin{equation}\label{eq:1d_embedding}
q^{(n)}(t) \bydef q_{\nt},
\end{equation}
where $\lfloor \cdot \rfloor$ is the floor function. Here, $t$ is the rescaled (accelerated) time: within $t \in [0, 1]$, the original discrete-time process moves $n$ steps. Due to the scaling of the noise term in \eref{1D} (with the noise variance equal to $n^{-1-2\delta}$ for some $\delta > 0$), the rescaled stochastic process $q^{(n)}(t)$ converges to a deterministic limit function as $n \to \infty$. We illustrate this convergence behavior using the following example.

\begin{example} \label{ex:demo}
Let us consider the special case where $f(q) = -q$. We plot in \fref{demo-conv} simulation results of $q^{(n)}(t)$ for several different values of $n$. We see that, as $n$ increases, the rescaled stochastic processes $q^{(n)}(t)$ converge to a limit function (the black line in the figure), which will be denoted by $q(t)$. To prove this convergence, we first expand the recursion \eref{1D} [by using the fact that $f(q) = -q$] and get
\begin{equation}\label{eq:qk_expanded}
q_k = (1-\tfrac{\tau}{n})^k q_0 + \Delta_k,
\end{equation}
where $\Delta_k$ is a zero-mean random variable defined as
\[
\Delta_k = \frac{1}{n^{(1/2) + \delta}} \sum_{i=0}^{k-1} (1-\tfrac{\tau}{n})^{k-1-i} v_i.
\]
Since $\set{v_i}_i$ are independent random variables with unit variance, $\EE (\Delta_k)^2 \le \tfrac{1}{n^{1+2\delta}}(1-(1-\tfrac{\tau}{n})^2)^{-1} = \mathcal{O}(n^{-2\delta})$. It then follows from \eref{qk_expanded} that, for any $t > 0$, 
\begin{equation}\label{eq:1d_limit}
q^{(n)}(t) = q_{\nt} \cip \lim_{n\to \infty} (1-\tfrac{\tau}{n})^{\nt} q_0 = q_0 e^{-\tau t},
\end{equation}
where $\cip$ stands for convergence in probability.

From \eref{1d_limit} we can also see why $k = \nt$ as used in \eref{1d_embedding} is the ``correct'' scaling under which $q^{(n)}(t)$ has a non-trivial limit as $n \to \infty$. Specifically, if the time scaling is too large, \emph{e.g.} by choosing $k= n^b t$ with some $b>1$, the limit on the right-hand side of \eref{1d_limit} becomes
\[
\lim_{n\to \infty} (1-\tfrac{\tau}{n})^{\lfloor n^b t \rfloor} q_0 = \begin{cases}
q_0, &\text{if } t = 0\\
0, &\text{if } t > 0.
\end{cases}
\]
In this case, the limiting curve $q(t)$ will abruptly drop from $q_0$ to $0$ at $t = 0$, leaving out any details of the transition between these two extreme values. Similarly, if the time scaling is too small, \emph{e.g.}, $k= n^{b'}  t$, with some $b'<1$, the limiting curve $q(t) \equiv q_0$ for any finite $t$, again revealing no information.
\end{example}
\begin{figure}[t]
\centering{} \includegraphics[width=0.8\linewidth]{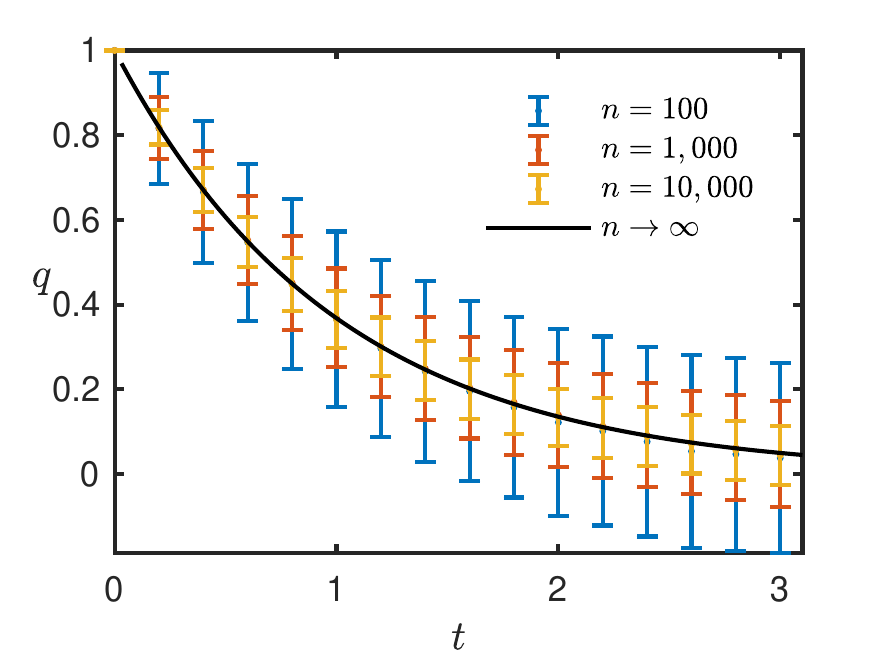}
\caption{\label{fig:demo-conv} Convergence of the 1-D  stochastic process $q_t^{(n)}$ described in Example \ref{ex:demo} to its deterministic scaling limit. Here, we use $\delta=0.25$. For each $n$, the associated error bars indicate $\pm1$  standard deviation over 1000 independent trials.}
\end{figure}

For general nonlinear functions $f(q)$, we can no longer directly simplify the recursion \eref{1D} as in \eref{qk_expanded}. However, similar convergence behavior of $q^{(n)}(t)$ still exists. Moreover, the limit function $q(t)$ can be characterized via an ODE. To see the origin of the ODE, we note that, for any $t > 0$ and $k = \nt$, we may rewrite \eref{1D} as
\begin{equation}\label{eq:prelimit}
\frac{q^{(n)}(t + 1/n)-q^{(n)}(t)}{1/n} = \tau f[q^{(n)}(t)] + n^{(1/2)-\delta} v_k.
\end{equation}
Taking the limit $n \to \infty$ on both sides of \eref{prelimit} and neglecting the noise term $n^{(1/2)-\delta} v_k$, we may then write---at least in a nonrigorous way---the following ODE
\begin{equation} \label{eq:toy-ode}
\frac{\dif }{\dif t} q(t) = \tau f[q(t)],
\end{equation}
which always has a unique solution due to the Lipschitz property of $f(\cdot)$. For instance, the ODE associated with the linear setting in Example~\ref{ex:demo} is $\frac{\dif}{\dif t} q(t) = -\tau q(t)$, whose unique solution $q(t) = q_0 e^{-\tau t}$ is indeed the limit established in \eref{1d_limit}. A rigorous justification of the above steps can be found in the theory of weak convergence of stochastic processes (see, for example, \cite{Billingsley:1999, Jacod:2010}).

\begin{figure}[t]
\centering{} \includegraphics[width=0.8\linewidth]{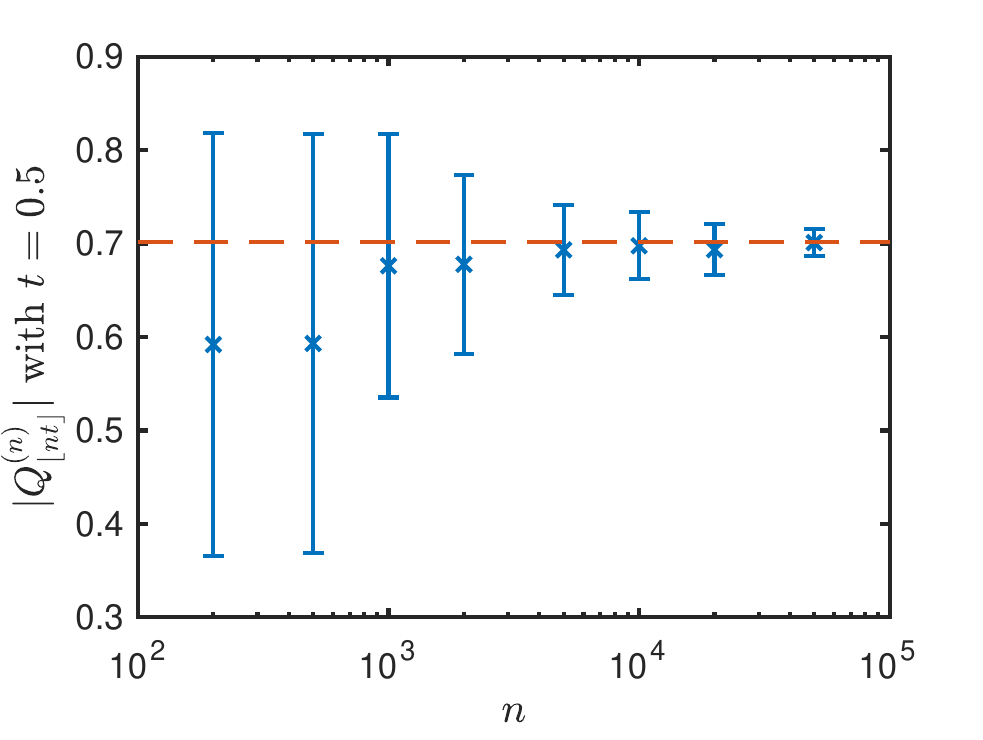}
\caption{\label{fig:scaling} Convergence of the cosine similarity $\mQ^{(n)}_{\nt}$ associated with GROUSE at a fixed rescaled time $t = 0.5$, as we increase $n$ from $200$ to $50, 000$. In this experiment, $d=1$ and thus $\mQ^{(n)}_k$ reduces to a scalar, denoted by $Q^{(n)}_k$. The error bars show the standard deviation of $Q^{(n)}_{\nt}$ over 100 independent trials. In each trial, we randomly generate a subspace $\mU$, the expansion coefficients $\set{\vc_k}$ and the noise vector $\set{\va_k}$ as in \eref{gen}. The red dashed lines is the limiting value predicted by our asymptotic characterization, to be given in Theorem~\ref{thm:ode-d1}.}
\end{figure}

Returning from the above detour, we recall that the central objects of our analysis are the cosine similarity matrices $\mQ_k^{(n)}$ defined in \eref{def-Q}. 
The dynamics of the cosine similarity matrix $\mQ_k$ is fully analogous to the situation demonstrated by the above toy example. In our work, we show that the increment $\mQ_{k+1} - \mQ_k$ has a mean of order $\mathcal{O}(1/n)$ and a variance of a smaller order. Thus, with time rescaling $k = \nt$ and $n \to \infty$, the process $\mQ(t)$ converges to the solutions of certain limiting ODEs given in the results shown in the next subsection.
This phenomenon is demonstrated in Figure \ref{fig:scaling}, where we plot the cosine similarity $\mQ_{\nt}^{(n)}$ of GROUSE at $t = 0.5$ for different values of $n$. In this experiment, we set $d = 1$ and thus $\mQ_{\nt}^{(n)}$ reduces to a scalar. The standard deviations of $\mQ_{\nt}^{(n)}$ over 100 independent trials, shown as error bars in Figure \ref{fig:scaling}, decrease as $n$ increases. This indicates that the performance of these 
stochastic algorithms can indeed be characterized by certain deterministic limits when the dimension is high.

\subsection{The Scaling Limits of Oja's, GROUSE and PETRELS}
\label{sec:thms}

To study the scaling limits of the cosine similarity matrices $\mQ^{(n)}_{\nt}$, we embed the discrete-time process $ \mQ_{k}^{(n)}$ into a continuous time process $ \mQ^{(n)}(t)$ via a simple piecewise-constant interpolation:
\begin{equation} \label{eq:embed-Q}
\mQ^{(n)}(t) \bydef \mQ_{ \nt}^{(n)}.
\end{equation}
The main objective of this work is to establish the high-dimensional limit of $\mQ^{(n)}(t)$ as $n \to \infty$. Our asymptotic analysis is carried out under the following technical assumptions on the generative model \eref{gen} and the observation model \eref{def-Omega}.

\begin{enumerate}[label={(A.\arabic*)}]
\item \label{ass:a} The elements of the noise vector $ \va_{k}$  are i.i.d. random variables
with zero mean, unit variance, and finite higher-order moments;

\item $\vc_{k}$ in \eref{gen} is a $d$-D random vector with zero-mean and  covariance matrix $\mLa$ as given in \eref{def-Lambda}. Moreover, all the higher-order moments of $\vc_k$ exist and are finite, and $\{\vc_k\}$ is independent of $\{\va_k\}$;

\item \label{ass:iid} We assume that $\set{v_{k, i}}$ in the observation model \eref{def-Omega} is a collection of independent and identically distributed binary random variables such that
\begin{equation}\label{eq:alpha}
\PP(v_{k, i} = 1) = \alpha,
\end{equation}
for some constant $\alpha \in (0, 1)$. Throughout the paper, we refer to $\alpha$ as the \emph{subsampling ratio}. We shall also assume that the algorithmic parameter $\e$ used in Algorithms \ref{alg:Oja}--\ref{alg:sim-petrels} satisfies the condition $\e< \a$.

\item \label{ass:u} The target subspace matrix  $\mU$ and the initial guess $\mX_0$ are both \emph{incoherent} in the sense that
\begin{equation}\label{eq:incoherent}
\sum_{i=1}^{n} \sum_{j=1}^d  U_{i,j}^4 \leq \frac{C}{n} \mbox{ and }
\sum_{i=1}^{n} \sum_{j=1}^d  X_{0,i,j}^4 \leq \frac{C}{n},
\end{equation}
where $U_{i,j}$ and $X_{0,i,j}$ denote the $(i,j)$th entries of $\mU$ and $\mX_0$, respectively, and $C$ is a generic constant that does not depend on $n$.

\item \label{ass:init} The initial cosine similarity $\mQ_0^{(n)}$ converges entrywise and in probability to a deterministic matrix $\mQ(0)$.

\item \label{ass:step-size}  For Oja's method and GROUSE, the step-size parameters $ \tau_{k}= \tau(k/n)$, where $ \tau( \cdot)$ is a deterministic function such that $\sup_{t \ge 0} \abs{\tau(t)} \le C$ for a generic constant $C$ that does not depend on $n$. For PETRELS, the discount factor
\begin{equation}\label{eq:discount}
\gamma = 1 - \tfrac{\mu}{n},
\end{equation}
for some constant $\mu > 0$. 
\end{enumerate}

Assumption~\ref{ass:u} requires some further explanation. Condition \eref{incoherent} essentially requires the basis matrix $\mU$ and the initial guess $\mX_0$ to be generic. To see this, consider a $\mU$ that is drawn uniformly at random from the Grassmannian for rank-$d$ subspaces. Such a $\mU$ can be generated as
\begin{equation}\label{eq:random_U}
\mU = \mX (\mX^\top \mX)^{-1/2},
\end{equation}
where $\mX$ is an $n \times d$ random matrix whose entries are i.i.d. standard normal random variables. For such a generic choice of $\mU$, one can show that its entries $U_{i, j} \sim \mathcal{O}(1/\sqrt{n})$ and that \eref{incoherent} holds with high probability when $n$ is large.

 \begin{theorem}[Oja's method and GROUSE] 
\label{thm:ode-d1} 
Fix $T > 0$, and let $\big\{ \mQ^{(n)}(t)\big\}_{t\in [0,T]}$ be the time-varying cosine similarity matrices associated with either Oja's method or GROUSE over the finite interval $t \in [0, T]$. Under assumptions \ref{ass:a}--\ref{ass:step-size}, we have 
\[
\big\{ \mQ^{(n)}(t)\big\}_{t\in [0,T]} \cw \mQ(t),
\]
where $\cw$ stands for weak convergence and $\mQ(t)$ is a deterministic matrix-valued process. Moreover, $\mQ(t)$ is the unique solution of the ODE 
\begin{equation} \label{eq:ODE-Oja}
\frac{\dif}{\dif t} \mQ(t)= F(\mQ(t), \tau(t)\mI_d ),
\end{equation}
where $F:\mR^{d\times d} \times \mR^{d \times d} \mapsto \mR^{d\times d}$ is a matrix-valued function defined as
\begin{equation} \label{eq:def-FQG}
 F(\mQ,\mG)\bydef \big[ \a \mLa^{2} \mQ - \tfrac{ \s^4}{2}\mQ\mG  
  - \mQ \big(\mI_d+ \tfrac{ \s^2}{2}\mG  \big)  \mQ^{ \T} \a\mLa^{2} \mQ \big] \mG.
\end{equation}
Here $\a$ is the subsampling ratio, and $\mLa$ is the diagonal covariance matrix defined in \eref{def-Lambda}.
\end{theorem}

In \sref{formal}, we present a (nonrigorous) derivation of the limiting ODE \eref{ODE-Oja}. Full technical details and a complete proof can be found in the Supplementary Materials \cite{WangEL:18a}. An interesting conclusion of this theorem is that the cosine similarity matrices $\mQ^{(n)}(t)$ associated with Oja's method and GROUSE converge to the same asymptotic trajectory. We will elaborate on this point in \sref{insights}.

To establish the scaling limits of PETRELS, we need to introduce three auxiliary 
$d\times d$ matrices
\begin{equation} \label{eq:def-AKW}
\begin{aligned}
\mA_k^{(n)} &\bydef \tfrac{1}{n} \mR_k^{-1} \\
\mK_k^{(n)} &\bydef \mU_k^\T \mX_k \\
\mW_k^{(n)} &\bydef \mX_k^\T \mX_k,
\end{aligned}
\end{equation}
where the matrices $\mR_k$ and $\mX_k$ are those used in Algorithm~\ref{alg:sim-petrels}.
Then, the cosine similarity matrix can be expressed by $\mQ_k^{(n)}=\mK_k^{(n)} \big(\mW_k^{(n)}\big)^{-\frac{1}{2}} $. Similar to \eqref{eq:embed-Q}, we embed the discrete-time processes $\mA_k^{(n)}$, $\mW_k^{(n)}$ and $\mW_k^{(n)}$ into continuous-time processes: 
\begin{equation} \label{eq:def-AKW-t}
\begin{aligned}
\mA^{(n)}(t) &\bydef \mA_{\nt}^{(n)} \\
\mK^{(n)}(t)&\bydef \mK_{\nt}^{(n)}  \\
\mW^{(n)}(t) &\bydef \mW_{\nt}^{(n)}.
\end{aligned}
\end{equation}
The following theorem, whose proof can be found in the Supplementary Materials \cite{WangEL:18a}, characterizes the asymptotic dynamics of PETRELS.
\begin{theorem}[PETRELS]\label{thm:limit}
For any fixed $T > 0$, let $\big\{ \big( \mA^{(n)}(t), \mK^{(n)}(t), \mW^{(n)}(t) \big) \big\}_{t\in [0,T]}$ be the process on $\R^{3d^2}$ defined in \eref{def-AKW-t}, and let $\big\{ \mQ^{(n)}(t)\big\}_{t\in [0,T]}$ be the time-varying cosine similarity matrices associated with PETRELS on the interval $t \in [0, T]$. 
Under assumptions~\ref{ass:a}--\ref{ass:step-size}, we have
\[ \begin{aligned}
&\big\{ \big( \mA^{(n)}(t), \mK^{(n)}(t), \mW^{(n)}(t) \big) \big\}_{t\in [0,T]}
 \\
 & \quad   \quad \;
  \cw \big( \mA(t), \mK(t), \mW(t) \big), \text{ and } \\
&\big\{ \mQ^{(n)}(t)\big\}_{t\in [0,T]} \cw \mQ(t),
 \end{aligned}
\]
 as $n \rightarrow \infty$, where 
\begin{equation}\label{eq:pet-Q}
\mQ(t)= \mK(t) \mW(t)^{-\frac{1}{2}} ,
\end{equation}
and $\big( \mA(t), \mK(t), \mW(t) \big)$ is the unique solution of the following system of coupled ODEs:
\begin{equation} \label{eq:ODE-pet}
\begin{aligned}
\tfrac{\dif}{\dif t} \mA(t) &= J_1(\mA(t), \mK(t), \mW(t)),  \\
\tfrac{\dif}{\dif t} \mK(t) &= J_2(\mA(t), \mK(t), \mW(t)),  \\
\tfrac{\dif}{\dif t} \mW(t) &= J_3(\mA(t), \mK(t), \mW(t)).
\end{aligned}
\end{equation}
Here,  $J_1$, $J_2$ and $J_3$ are functions defined by
\begin{equation}\label{eq:def-pet-J}
\begin{aligned}
J_1 (\mA, \mK, \mW) &\bydef   \mW^{-1}
\big( 
 \mK^\T \a\mat \Lambda^2 \mK + \s^2 \mW \big)
 \mW ^{-1}
-\mu\mA
\\
J_2(\mA, \mK, \mW) &\bydef (\a \mat \Lambda^2 + \s^2 \mI_d)\mK \mW^{-1} \mA^{-1}
\\
& \quad \;
- \mK \mW^{-1} \big(
 \mK^\T \a\mat \Lambda^2 \mK + \s^2 \mW
\big)
\mW^{-1}
\mA^{-1}
\\
J_3 (\mA, \mK, \mW)&\bydef \s^2 \mA^{-1} \mW^{-1} 
\\
&\quad \; ( \mK^\T \a \mat \Lambda^2 \mK + \s^2 \mW) 
  \mW^{-1}
\mA^{-1},
\end{aligned}
\end{equation}
where $\mu > 0$ is the constant given in \eref{discount}.  
\end{theorem}

\begin{figure*}
\centering
\subfigure[GROUSE (crosses) and Oja (circles)]{\label{fig:comparison:1}
\includegraphics[width=0.45\linewidth]{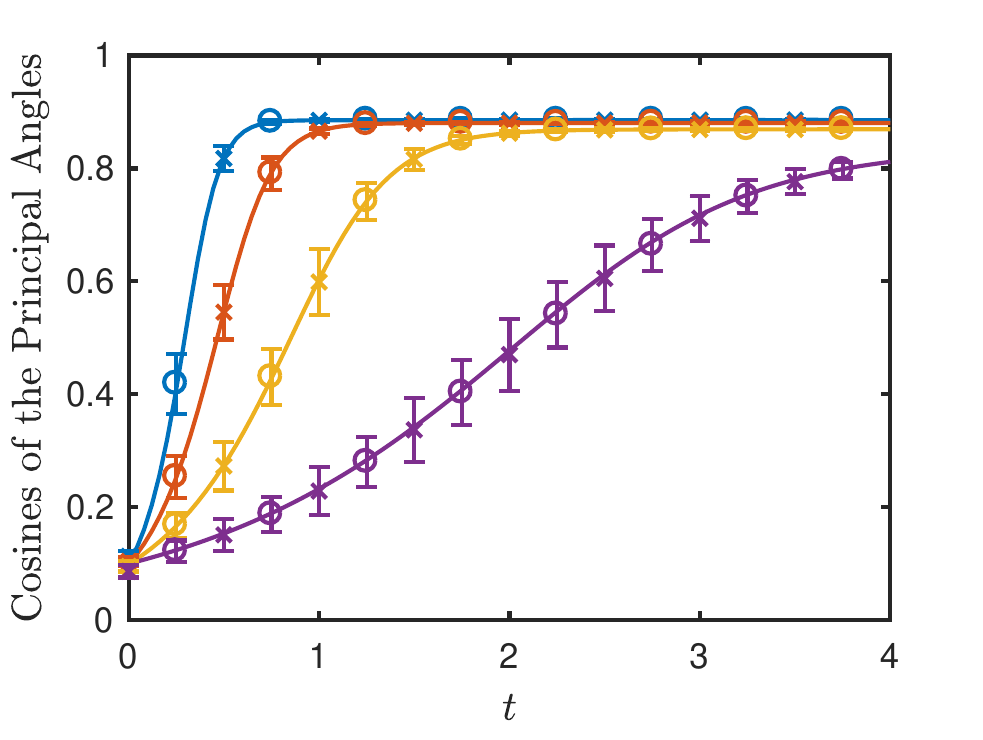}
}
\hspace{1ex}
\subfigure[PETRELS]{\label{fig:comparison:2}
\includegraphics[width=0.45\linewidth]{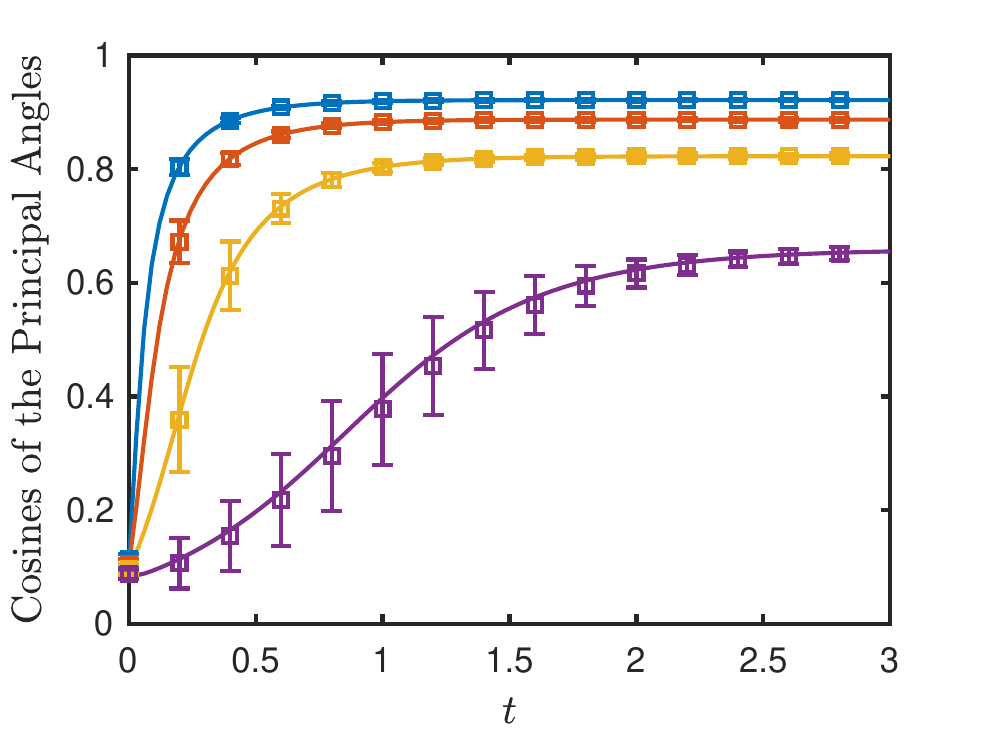}
}
\caption{\label{fig:comparison} Numerical simulations vs. asymptotic characterizations. (a) Results for Oja's method and GROUSE, where the solid lines are the theoretical predictions of the cosines of $4$ principal angles by the solution of the ODE \eqref{eq:ODE-Oja}. The crosses (for Oja's method) and circles (for GROUSE) show the simulation results averaged over 100 independent trials. In each trial, we randomly generate a subspace $\mU$ as in \eref{random_U}, the expansion coefficients $\set{\vc_k}$ and the noise vector $\set{\va_k}$. The error bars indicate $\pm 2$ standard deviations. (b) Similar comparisons of numerical simulations and theoretical predictions for PETRELS.}
\end{figure*}

Theorem~\ref{thm:ode-d1} and Theorem~\ref{thm:limit} establish the scaling limits of Oja's method, GROUSE and PETELS, respectively, as $n \to \infty$. In practice, the dimension $n$ is always finite, and thus the actual trajectories of the performance curves will fluctuate around their asymptotic limits. To bound such fluctuations via a finite-sample analysis, we first need to slightly strengthen assumption~\ref{ass:init} as follows:
\begin{enumerate}[label={(A.\arabic*)},resume]
\item \label{ass:init_strong} Let $\mQ_0^{(n)}$ be the initial cosine similarity matrices. There exists a fixed matrix $\mQ(0)$ such that
\[
\EE \norm{\mQ_0^{(n)} - \mQ(0)}_2 \le {C}{n}^{-1/2},
\]
where $\norm{\cdot}_2$ denotes the spectral norm of a matrix and $C > 0$ is a constant that does not depend on $n$.
\end{enumerate}

\begin{theorem}[Finite Sample Analysis] \label{thm:finite}
Let $\mQ^{(n)}(t)$ be the time-varying cosine similarity matrices associated with Oja's method, GROUSE, or PETELS, respectively. Let $\mQ(t)$ denote the corresponding scaling limit given in \eref{ODE-Oja} and \eref{pet-Q}, respectively. Fix any $T > 0$. Under assumptions \ref{ass:a}--\ref{ass:u}, \ref{ass:step-size}--\ref{ass:init_strong}, for any $t \in [0, T]$, we have
\begin{equation}
\sup_{n \ge 1} \, \EE \norm{\mQ^{(n)}(t) - \mQ(t)}_2 \leq \frac{C(T)}{\sqrt{n}},
\end{equation}
where $C(T)$ is a constant that can depend on the terminal time $T$ but not on $n$.
\end{theorem}

The above theorem, whose proof can be found in the Supplementary Materials \cite{WangEL:18a}, shows that the rate of convergence towards the scaling limits is $\mathcal{O}(1/\sqrt{n})$.

\begin{example}
To demonstrate the accuracy of the asymptotic characterizations given in Theorem~\ref{thm:ode-d1} and Theorem~\ref{thm:limit}, we compare the actual performance of the algorithms against their theoretical predictions in \fref{comparison}. In our experiments, we generate a random orthogonal matrix $\mU$ according to \eref{random_U} with $n = 20,000$ and $d = 4$. For Oja's method and GROUSE, we use a constant step size $\tau = 0.5$. For PETRELS, the discount factor is $\gamma = 1 - \mu/n$ with $\mu = 5$, and $\mR_0=\frac{\delta}{n} \mI_d$ with $\delta=10$. The covariance matrix is set to
\[
\mLa = \diag\set{5,4, 3, 2 },
\]
 the subsampling ratio is $\alpha = 0.5$ and the variance of the background noise $\s^2=1$. \fref{comparison:1} shows the evolutions of the cosines of the 4 principal angles between $\mU$ and the estimates given by Oja's method (shown as crosses) and GROUSE (shown as circles). We compute the theoretical predictions of the principal angles by performing a SVD of the limiting matrices $\mQ(t)$ as specified by the ODE \eref{ODE-Oja}. (In fact, this ODE has a simple analytical solution; see \sref{closed_form_Oja} for details.) \fref{comparison:2} shows similar comparisons between PETRELS and its corresponding theoretical predictions. In this case, we solve the limiting ODEs \eref{ODE-Q-P} and \eref{ODE-G-P} numerically. 
\end{example}

\subsection{Related Work}
\label{sec:related}

The problem of estimating and tracking low-rank subspaces has received a lot of attention recently in the signal processing and learning communities. Under the setting of fully observed data, an earlier work \cite{Mitliagkas2013} studied a block-version of Oja's method and provided a sample complexity estimate for the case of $d = 1$. Similar analysis is available for general $d$-dimensional subspaces \cite{hardt2014noisy,balcan2016improved}. The streaming version of Oja's method and its sample complexities have also been extensively studied, \emph{e.g.}, \cite{li2016near, jain2016streaming,balsubramani2013fast,shamir2016convergence,allen2017first}.

For the case of incomplete observations, the sample complexity of a block version of Oja's method with missing data is analyzed in \cite{Mitliagkas2014} under the same generative model as in \eqref{eq:gen}. In \cite{gonen2016subspace}, the authors provide the sample complexity for learning a low-rank subspace from subsampled data under a nonparametric model: the complete data vectors are assumed to be i.i.d. samples from a general probability distribution on a subset of the Euclidean unit ball in $\R^n$. 
Different from that work, we study the dynamics of performance of the three algorithms rather than sample complexity. 
In the streaming setting, Oja's method, GROUSE, PETRELS are three popular algorithms for tackling the challenge of subspace learning with partial information. Other interesting approaches include online matrix completion methods \cite{krishnamurthy2013low, jin2016provable,lois2015online}. See \cite{Balzano2018} for a recent review of relevant literature in this area. Local convergence of GROUSE is given in \cite{Balzano:2015, ZhangB:2016}. Global convergence of GROUSE is established in \cite{Zhang2015} under the noiseless setting. In general, establishing finite sample global performance guarantees for GROUSE and other algorithms such as Oja's and PETRELS in the missing data case is still an open problem.

Unlike most work in the literature that seeks to establish finite-sample performance guarantees for various subspace estimation algorithms, our results in this paper provide an asymptotically exact characterization of three popular methods in the high-dimensional limit. The main technical tool behind our analysis is the weak convergence of stochastic processes towards their scaling limits that are characterized by deterministic ODEs or stochastic differential equations (see, \emph{e.g.}, \cite{Meleard:1987, Sznitman:1991, EthierK:85,Wang2017c}).

Using ODEs to analyze stochastic recursive algorithms has a long history 
\cite{Kurtz:1970lr,Ljung:1977}. An ODE analysis of an early subspace tracking 
algorithm was given in Yang \cite{Yang:1996}, and this result was adapted to 
analyze PETRELS for the nonsubsampled case
\cite{Chi2013}. Our results in this paper differ from 
previous analysis in several important aspects. First, it handles the more challenging 
case of incomplete observations. Second, 
previous ODE analysis in Yang \cite{Yang:1996} and Chi \emph{et al}. \cite{Chi2013} keeps the ambient 
dimension $n$ \emph{fixed} and studies the asymptotic limit as the step 
size tends to $0$. The resulting ODEs involve $\mathcal{O}(n)$ variables. 
In contrast, our analysis studies the limit as the dimension 
$n \rightarrow \infty$, and the resulting ODEs only involve at most $3d^2$ variables, where $d$ is the dimension of the subspace which, in many practical situations, is a small constant. 
This low-dimensional characterization makes our 
limiting results more practical to use, especially when the ambient dimension 
$n$ is large.

Our results build upon a previous work \cite{WangL:16} that obtains the ODE limits of Oja's method under the fully observed model, for the special case of $d = 1$. This paper extends that line of work to the more challenging case with missing data, and for arbitrary $d$. The partially observed setting presents several technical difficulties: First, unlike the fully observed case with Gaussian noise where the cosine similarity $\mQ_k$ forms a low-dimensional closed Markov chain, the situation is more complicated under partial observations. Due to subsampling (which breaks rotational symmetry), the cosine similarity $\mQ_k$ no longer forms a closed Markov chain at any finite dimension $n$. Only when we take the asymptotic limit $n \to \infty$ do we obtain a closed dynamics given by the ODE limits. Second, in the partially observed case, an extra estimation step of the missing data should be made. This presents some additional technical challenges in the performance analysis. 

It is important to point out a limitation of our asymptotic analysis: we require the initial estimate $\mX_0$ to be asymptotically correlated with the true subspace $\mU$. To see why this is an issue, we note that if the initial cosine similarity matrix $\mQ(0) = \mat{0}$ [\emph{i.e.}, a fully uncorrelated initial estimate, which can be obtained by setting $\mX$ to be an $n \times d$ i.i.d. Gaussian matrix, and assigning $\mX_0 = \mX (\mX^T  \mX)^{-1/2}$], then the ODEs in Theorems~\ref{thm:ode-d1} and \ref{thm:limit} only provide a trivial solution $\mQ(t) \equiv 0$, yielding no useful information.
Strictly speaking, our asymptotic predictions are still valid in this random initialization case, since Assumptions
\ref{ass:u} and \ref{ass:init} are satisfied, but  the predictions are not useful as $\mQ(t) \equiv 0$ for any finite $t$. It means that, starting from a random initialization, the algorithms cannot escape from the initial region in $\mathcal{O}(n)$ iterations. 
The utility of our results is to predict and trace the performance of the algorithms in the second phase of the dynamics, after the algorithms already have some estimates that are correlated with the true subspace.
In practice, a correlated initial estimate can be obtained by performing PCA on a small batch of samples \cite{candes2009exact,gonen2016subspace}; it may  also be available from additional side information about the true subspace $\mU$. Therefore, the requirement that $\mQ(0)$ be invertible is not overly restrictive. In practice, we observe in numerical simulations that, under sufficiently high SNRs, Oja's method, GROUSE and PETRELS can successfully estimate the subspace by starting from random initial guesses that are uncorrelated with $\mU$. Extending our analysis framework to handle the case of random initial estimates is an important line of future work.


\section{Implications of High-Dimensional Analysis}
\label{sec:implications}

The scaling limits presented in \sref{main_results} provide asymptotically exact characterizations of the dynamic performance of Oja's method, GROUSE, and PETRELS. In this section, we discuss implications of these results. Analyzing the limiting ODEs also reveals the fundamental limits and phase transition phenomena associated with the steady-state performance of these algorithms.

\subsection{Algorithmic Insights}
\label{sec:insights}

By examining Theorem~\ref{thm:ode-d1} and Theorem~\ref{thm:limit}, we draw the following conclusions regarding the three subspace estimation algorithms.

\emph{1. Connections and differences between the algorithms.} Theorem \ref{thm:ode-d1} implies that, as $n \to \infty$, Oja's method and GROUSE converge to the same deterministic limit process characterized as the solution of the ODE \eref{ODE-Oja}. This result is somewhat surprising, as the update rules of the two methods (see Algorithm~\ref{alg:Oja} and Algorithm~\ref{alg:grouse}) appear to be sufficiently different.  

Theorem \ref{thm:limit} shows that PETRELS is also intricately connected to the other two algorithms. 
We here consider a special case that $\mK(0)=\mU^\T \mX_0$ is a diagonal matrix. In this case, $\mA(t)$, $\mK(t)$ and $\mW(t)$ will also be diagonal matrices for any $t>0$. 
Define 
\begin{equation} \label{eq:def-G}
\mG(t)\bydef \mW^{-\frac{1}{2}} \mA^{-1}\mW^{-\frac{1}{2}}.
\end{equation}
One can show that the evolution of $\mG(t)$ and $\mQ(t)$ as defined in \eqref{eq:pet-Q} is also governed by a first-order ODE, which can be deduced from \eqref{eq:ODE-pet} as
\begin{align}
\frac{\dif}{\dif t} \mQ(t) &= F(\mQ(t), \mG(t)), \label{eq:ODE-Q-P}\\
\frac{\dif}{\dif t} \mG(t) &= H(\mQ(t), \mG(t)). \label{eq:ODE-G-P}
\end{align}
Here, $F$ is the function defined in \eqref{eq:def-FQG} and $H$ is a function defined by
\begin{equation}\label{eq:def-HQG}
H(\mQ, \mG) \bydef \mG \left[ \mu - \mG (\s^2\mG+\mI_d) (\mQ^\T \a \mLa^2 \mQ + \s^2 \mI_d ) \right].
\end{equation}
The ODE \eqref{eq:ODE-Q-P} of the cosine similarity matrix $\mQ(t)$ for PETRELS has exactly the same form as the one for GROUSE and Oja's method shown in \eqref{eq:ODE-Oja}, except for the fact that the nonadaptive step-size $\tau(t)\mI_d$ in \eqref{eq:ODE-Oja} is  replaced by a $d \times d$ matrix $\mG(t)$, itself governed by an ODE \eqref{eq:ODE-Q-P}. Thus, $\mG(t)$ in PETRELS can be viewed as an adaptive scheme for adjusting the step-size. 

\begin{figure}[t]
\centering
\includegraphics[width=\linewidth]{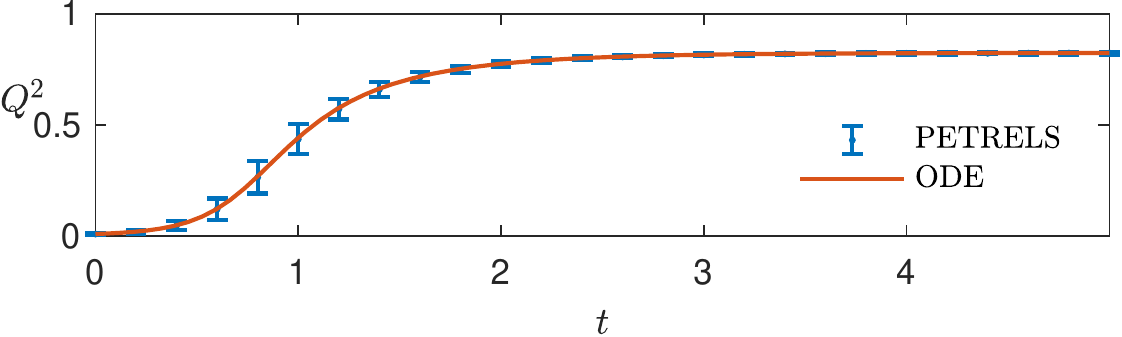} \\
\includegraphics[width=\linewidth]{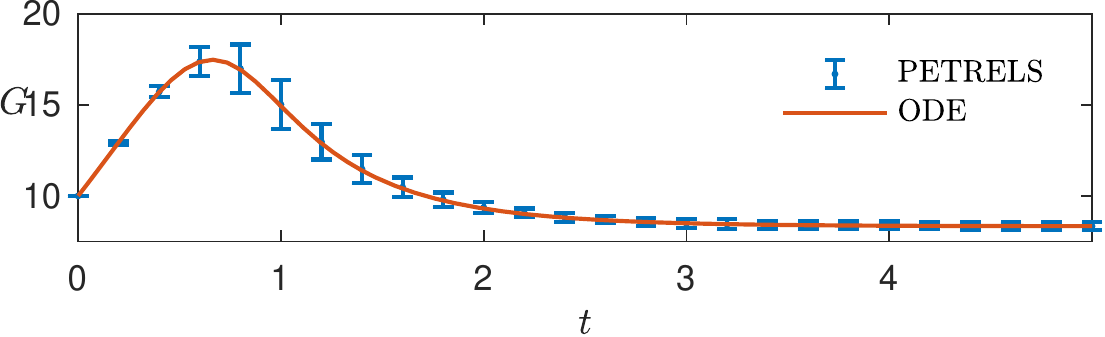}
\caption{\label{fig:f-t}Monte Carlo simulations of the PETRELS algorithm v.s. asymptotic predictions obtained by the limiting ODEs given in Theorem~\ref{thm:limit} for $d=1$. In this case, the two matrices $\mQ(t)$ and $\mG(t)$ reduce to two scalars $Q(t)$ and $G(t)$. The variable $G(t)$ acts as an effective step-size, which adaptively adjusts its value according to the change in $Q(t)$. The error bars shown in the figures represent one standard deviation over 50 independent trials.
The signal dimension is $n = 10^4$.}
\end{figure}

To investigate how $\mG(t)$ evolves, we run an experiment for $d= 1$. In this case, the quantities $\mQ(t)$, $\mG(t)$ and $\mat{\Lambda}$ reduce to three scalars, denoted by $Q(t)$, $G(t)$, and $\lambda$, respectively. Figure \ref{fig:f-t} shows the dynamics of PETRELS to recover this 1-D subspace. It shows that $G(t)$ increases initially, which helps boost 
the convergence speed. As $Q(t)$ increases (meaning the estimates becoming more accurate), however, the effective step-size $G(t)$ gradually decreases, in order to help $Q(t)$ reach a higher steady-state value. 


\emph{2. Subsampling vs. the SNR.} The ODEs in Theorems \ref{thm:ode-d1} and \ref{thm:limit} also reveal an interesting (asymptotic) equivalence between the subsampling ratio $\alpha$ and the SNR as specified by the matrix $\mLa$. To see this, we observe from the definition of the functions $F$ in \eref{def-FQG} and $J_1, J_2, J_3$ in \eref{def-pet-J} that $\a$ always appears together with $\mat{\Lambda}$ in the form of a product $\a \mat{\Lambda}^2$. This implies that an observation model with subsampling ratio $\alpha$ and signal strength $\mLa$ will have the same asymptotic performance as a different model with subsampling ratio $\widehat{\alpha}$ and signal strength $\sqrt{\alpha/\widehat{\alpha}}\, \mLa$. In simpler terms, having missing data is asymptotically equivalent to lowering the signal strength in the fully-observable setting. We note that  \cite{gonen2016subspace} pointed out a connection between the subsampling ratio and the sample complexity. Here, we reveal  a  connection between the subsampling ratio and the SNR.


\subsection{Oja's Method and GROUSE: Analytical Solutions and Phase Transitions}
\label{sec:closed_form_Oja}

Next, we investigate the dynamics of Oja's method and GROUSE by studying the solution of the ODE given in Theorem \ref{thm:ode-d1}. To that end, we consider a change of variables by defining
\begin{equation} \label{eq:def-P}
\mP(t)\bydef [\mQ(t) \mQ^\T(t)]^{-1}.
\end{equation}
One may deduce from \eref{ODE-Oja} that the evolution of $\mP(t)$ is also governed by a first-order ODE:
\begin{equation}\label{eq:P_ode}
\frac{\dif}{\dif t} \mP(t)=  
\mA(t)
 -\mP(t) \mB(t) - \mB(t) \mP(t),
\end{equation}
where
\begin{align}
\mA(t) &= \tau(t) [2+\tau(t)\s^2]\a\mat{\Lambda}^2 \label{eq:A_mtx}\\
\mB(t) &=\tau(t)\big( \a \mat{\Lambda}^2 - \tfrac{\tau(t)}{2} \s^4 \mI_d \big)\label{eq:B_mtx}
\end{align}
are two diagonal matrices. Thanks to the linearity of \eref{P_ode}, it admits an analytical solution
\begin{equation} \label{eq:P}
\begin{aligned}
\mP(t)&= e^{- \int_{0}^{t}\mB(r)\dif r} \mP(0) e^{- \int_{0}^{t}\mB(r)\dif r} \\
&\qquad\qquad+\int_{0}^t\mA(s)e^{ -2\int_{s}^{t}\mB(r)\dif r}\dif  s.
\end{aligned}
\end{equation}
Note that the first two terms on the right-hand side of \eqref{eq:P} represent the influence of the initial estimate $\mP(0)=[\mQ(0)\mQ^\T(0)]^{-1}$ on the current state at $t$. In the special case of the algorithms using a constant step size, \emph{i.e.}, $\tau(t) \equiv \tau > 0$, the solution \eref{P} may be further simplified as
\begin{equation}\label{eq:P-const}
\mP(t)= e^{-t \mB} \mP(0) e^{-t\mB } + \mZ(t),
\end{equation}
where $\mZ(t) = \diag\set{z_1(t), \ldots, z_d(t)}$ with
\begin{equation}\label{eq:z}
z_\ell(t) =  \frac{(2+\tau \s^2 )\alpha \lambda_\ell^2}{2\alpha \lambda_\ell^2 - \tau \s^4} \left(1 - e^{\tau(2\alpha \lambda_\ell^2 - \tau \s^4) t}\right)
\end{equation}
for $1 \le \ell \le d$. Note that if $2\alpha \lambda_\ell^2 - \tau \s^4= 0$ for some $\ell$, then the above expression for $z_\ell$ is understood via the convention that $(1-e^{-\tau t0})/0 = \tau t$.

The formula \eref{P-const} reveals a phase transition phenomenon for the steady-state performance of the two algorithms as we change the step-size parameter $\tau$. To see that, we first recall that the eigenvalues of $\mQ^{(n)}(t) (\mQ^{(n)}(t))^\top$ are exactly equal to the squared cosines of the $d$ principal angles $\set{\theta_\ell^{n}(t)}$ between the true subspace $\mU$ and the estimate given by the algorithms. We say an algorithm generates an asymptotically \emph{informative solution} if
\begin{equation}\label{eq:informative}
\lim_{t \to \infty} \lim_{n \to \infty} \cos^2\big[ \theta^{n}_\ell(t) \big] > 0 \quad \text{for all } 1 \le \ell \le d,
\end{equation}
\emph{i.e.}, the steady-state estimates of the algorithms achieve nontrivial correlations with all the directions of $\mU$. In contrast, a \emph{noninformative solution} corresponds to 
\begin{equation}\label{eq:noninformative}
\lim_{t \to \infty} \lim_{n \to \infty} \cos^2 \big[ \theta^{n}_\ell(t) \big] = 0 \quad \text{for all } 1 \le \ell \le d,
\end{equation}
in which case the steady-state estimates carry no information about $\mU$. For $d > 1$, one may also have a third situation where only a subset of the directions of $\mU$ can be recovered (with nontrivial correlations) by the algorithm.

\begin{proposition}\label{prop:phase_Oja_GROUSE}
Let $\theta^{(n)}_\ell(t)$ denotes the $\ell$th principal angle between the true subspace and the estimate obtained by Oja's method or GROUSE with a constant step size $\tau$. Under the same assumptions as in Theorem~\ref{thm:ode-d1}, we have
\begin{equation}\label{eq:Oja_GROUSE_steady}
\lim_{t \to \infty} \lim_{n \to \infty} \cos^2(\theta_\ell^{(n)}(t)) =  \max \Big\{ 0, \tfrac{2{\a \lambda_\ell^2}-{\tau}\s^4 }{{\a \lambda_\ell^2}(2+ \tau \s^2) } \Big\},
\end{equation}
where $\set{\lambda_\ell}$ are the SNR parameters defined in \eref{def-Lambda}. It follows that the two algorithms provide asymptotically informative solutions \emph{if and only if}
\begin{equation}\label{eq:tau_steady}
 \tau <  \frac{2\a}{\s^4} \min_{1 \le \ell \le d}  \lambda_\ell^{2}.
\end{equation}
\end{proposition}

\begin{IEEEproof}
Suppose the diagonal matrix $\mB$ in \eref{B_mtx} has $d_1$ positive diagonal entries (with $0 \le d_1 \le d$), and $d_2=d-d_1$ negative or zero entries. Without loss of generality, we may assume that $\mB$ can be split into a block form
 $\begin{bmatrix} \mB_1 & \mat{0}_{d_1\times d_2} \\ \mat{0}_{d_2 \times d_1}& -\mB_2 \end{bmatrix}$
  such that 
$\mB_1$ only contains the positive diagonal entries, and $-\mB_2$ only contains the nonpositive entires. 
Accordingly, we split the other two matrices  in \eqref{eq:P-const} as 
$\mP(0)=\begin{bmatrix} \mP_{1,1} &\mP_{1,2} \\ \mP_{2,1}& \mP_{2,2} \end{bmatrix}$ 
and 
$\mZ(t)=\begin{bmatrix} \mZ_1(t) & \mat{0}_{d_1\times d_2} \\ \mat{0}_{d_2 \times d_1}& \mZ_2(t) \end{bmatrix}$. Applying the block matrix inverse formula to  \eqref{eq:P-const}, we get
\begin{equation}\label{eq:P_inv}
\mP^{-1}(t)= 
 \begin{bmatrix} \mW_{1,1}(t) & 
\mW_{1,2}(t)\\
\mW_{2,1}(t) & \mW_{2,2}(t)
\end{bmatrix},
\end{equation}
where
\[
\begin{aligned}
&\mW^{-1}_{1,1}(t)= e^{-t\mB_1} \mP_{1,1}e^{-t\mB_1} + \mZ_1(t)\\
&\ - e^{-t\mB_1} \mP_{1,2} e^{t \mB_2}\left(e^{t\mB_2} \mP_{2,2} e^{t\mB_2} + \mZ_2\right)^{-1}  e^{t\mB_2} \mP_{2,1} e^{-t \mB_1}.
\end{aligned}
\]
It is easy to verify from the definitions of $\mB$ and $\mZ$ that 
\begin{equation}\label{eq:W11}
\lim_{t \to \infty} \mW_{1,1}(t) = \diag\set{\tfrac{2\alpha \lambda_1^2-\tau \s^4 }{\a \lambda_1^2(2+ \tau \s^2)}, \ldots,  \tfrac{2\alpha \lambda_{d_1}^2-\tau \s^4}{\a \lambda_{d_1}^2(2+ \tau\s^2)}}.
\end{equation}
Similarly, we may verify that 
\begin{equation}\label{eq:W_zero}
\begin{aligned}
\lim_{t \to \infty} \mW_{1,2}(t) &= \mat{0}_{d_1 \times d_2}\\
\lim_{t \to \infty} \mW_{2,2}(t) &= \mat{0}_{d2 \times d_2}.
\end{aligned}
\end{equation}
Substituting \eref{W11} and \eref{W_zero} into \eref{P_inv} and recalling that the eigenvalues of $\mP^{-1}(t)$ are exactly equal to the squared cosines of the principal angles, we reach \eref{Oja_GROUSE_steady}. Applying the conditions given in \eref{informative} and \eref{noninformative} to \eref{Oja_GROUSE_steady} yields \eref{tau_steady}.
\end{IEEEproof}

\subsection{Steady-State Analysis of PETRELS}

The steady-state property of PETRELS can also be obtained by studying the limiting ODEs as given in Theorem~\ref{thm:limit}. The coupling of $\mQ(t)$ and $\mG(t)$ in \eref{ODE-Q-P} and \eref{ODE-G-P}, however, makes the analysis much more challenging. Unlike the case of Oja's method and GROUSE, we are not able to obtain closed-form analytical solutions of the ODEs for PETRELS. In what follows, we restrict our discussions to the special case of $d = 1$. This simplifies the task, as the matrix-valued ODEs \eqref{eq:ODE-Q-P} and \eqref{eq:ODE-G-P} reduce to scalar-valued ones. 

\begin{figure}[t]
\centering
\subfigure[informative solution]{\label{fig:portraits:1}
\includegraphics[width=0.47\linewidth]{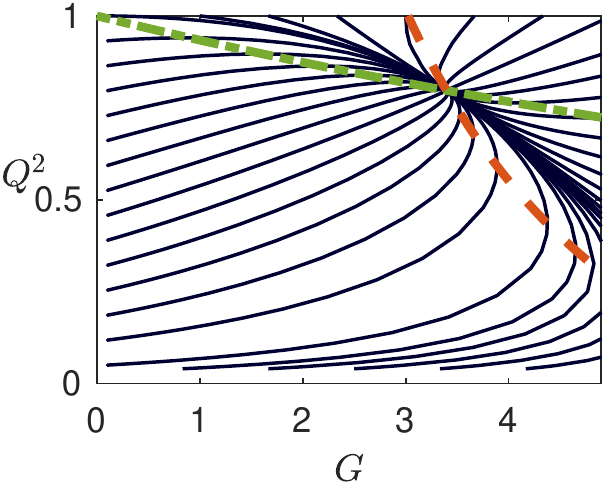}}
\subfigure[informative solution]{\label{fig:portraits:2}
\includegraphics[width=0.47\linewidth]{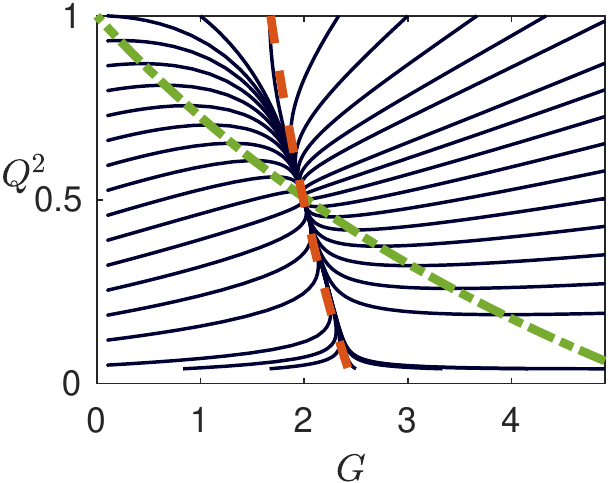}}
\subfigure[noninformative solution]{\label{fig:portraits:3}
\includegraphics[width=0.47\linewidth]{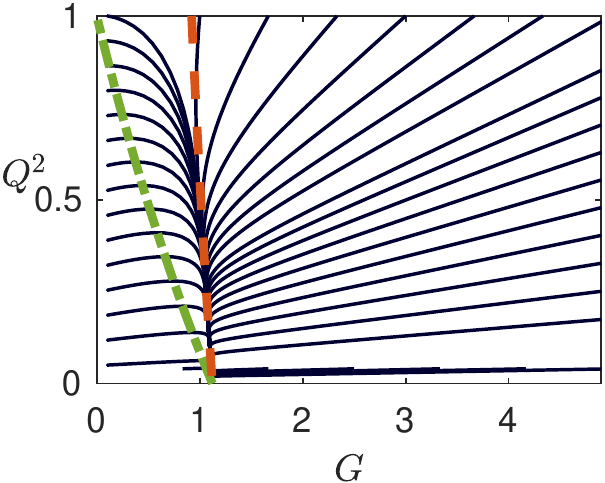}}
\subfigure[noninformative solution]{\label{fig:portraits:4}
\includegraphics[width=0.47\linewidth]{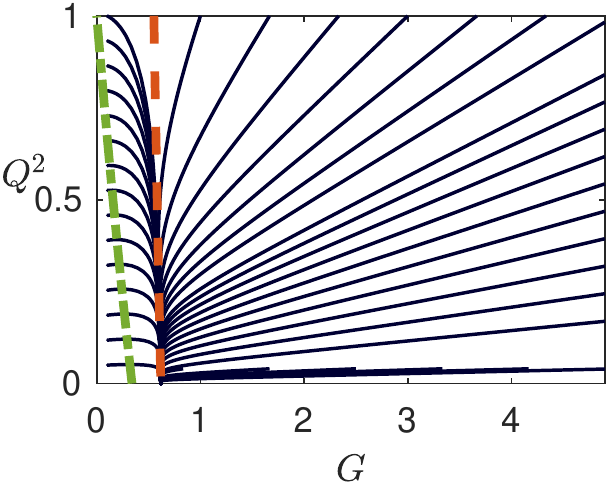}}
\caption{\label{fig:protraits}
Phase portraits of the nonlinear ODEs in
Theorem~\ref{thm:limit}: The black curves are trajectories of the 
solutions $(Q^2(t),G(t))$ of the ODES starting from different
initial values. The green  and red  curves represent nontrivial 
solutions of the two stationary equations
$\frac{\dif Q^2(t)}{\dif t}=0$ and $ \frac{\dif G(t)}{\dif t}=0$.
Their intersection point, if it exists, is a stable fixed point of the dynamical system. 
The fixed-points of the top two figures correspond to $Q^2(\infty)>0$, 
and thus the steady-state solutions in these two cases are informative. 
In contrast, the fixed-points of the bottom two figures are associated 
with noninformative steady-state solutions with $Q^2( \infty)=0$. }
\end{figure}

It is not hard to verify that, for any solution $\set{Q(t),R(t)}$ with an initial condition 
$\set{Q(0),R(0)}$, there is a symmetric solution $\set{-Q(t), G(t)}$ for the initial condition $\set{-Q(0), G(0)}$. To remove this redundancy, it is convenient to investigate the dynamics of $Q^2(t)$ and $G(t)$, which 
satisfy the following ODEs
\begin{align} \label{eq:ODE-1d-Q}
\frac{\dif }{\dif t}[Q^2(t)]&= GQ^2[2\a \lambda^2 - \s^4G - 2Q^2 (\s^2+\tfrac{1}{2}G)\a\lambda^2]
\\
\frac{\dif }{\dif t}G(t)&=G[\mu - G(G\s^2+1) (Q^2 \a \lambda^2 +\s^2)]. \label{eq:ODE-1d-G}
\end{align}
Figure \ref{fig:protraits} visualizes several different solution trajectories of these ODEs as the black curves in the $Q$--$G$ plane. 
These solutions start from different initial conditions at the borders of the figures, and they converge to certain stationary points. The locations of these stationary points depend on the SNR $\set{\lambda_\ell}$, the subsampling ratio $\alpha$ and the discount parameter $\mu$ used by the algorithm. In Figures~\ref{fig:portraits:1} and \ref{fig:portraits:2}, the stationary points correspond to $Q^2 > 0$, and thus the algorithm generates asymptotically informative solutions according to the definition in \eref{informative}. In contrast, \fref{portraits:3} and \fref{portraits:4} show the situations where the steady-state solutions are noninformative.

\begin{proposition}
\label{prop:phase_PETRELS}
Let $d=1$. Under the same assumptions as in Theorem~\ref{thm:limit}, PETRELS generates an asymptotically informative solution \emph{if and only if}
\begin{equation} \label{eq:mu-c}
\mu<   \left(2 \alpha \lambda^2/\s^2+\tfrac{1}{2} \right)^{2}-\tfrac{1}{4},
\end{equation}
where $\mu$ is the parameter defined in \eref{discount}, $\lambda$ denotes the SNR in \eref{def-Lambda}, and $\alpha$ is the subsampling ratio.
\end{proposition}
\begin{IEEEproof}
It follows from Theorem~\ref{thm:limit} that verifying conditions \eref{informative} and \eref{noninformative} boils down to studying the fixed point of a dynamical system governed by the limiting ODEs \eqref{eq:ODE-1d-Q} and \eqref{eq:ODE-1d-G}. This task is in turn equivalent to setting the left-hand sides of the ODEs to zero and solving the resulting equations.

Let $\set{Q^\ast, G^\ast}$ be any solution to the equations $\frac{\dif G^2}{\dif t} = 0$ and $\frac{\dif G}{\dif t} = 0$. From the forms of the right-hand sides of \eqref{eq:ODE-1d-Q} and \eqref{eq:ODE-1d-G}, we see that $\set{Q^\ast, G^\ast}$ must fall into one of the following three cases:

Case I: $G^\ast = 0$ and $Q^\ast$ can take arbitrary values;

Case II: $Q^\ast = 0$ and $G^\ast$ is the unique positive solution to
\begin{equation}\label{eq:G-II}
G^\ast (G^\ast + 1) = \mu;
\end{equation}

Case III: $Q^\ast \neq 0$ and $G^\ast \neq 0$.

A local stability analysis, deferred to the end of the proof, shows that the fixed points in Case I are always unstable, in the sense that any small perturbation will make the dynamics move away from these fixed points. Thus, we just need to focus on Case II and Case III, with the former corresponding to an uninformative solution and the latter to an informative one. We will show that, under \eref{mu-c}, a fixed point in Case III exists and it is the unique stable fixed point. That solution disappears when \eref{mu-c} ceases to hold, in which case the solution in Case II becomes the unique stable fixed point.

To see why \eref{mu-c} provides the phase transition boundary, we note that a solution in Case III, if it exists, must satisfy $(Q^\ast)^2 =f(G^\ast)$ and $(Q^\ast)^2=h(G^\ast)$, where
\begin{align}
f(G)&\bydef \frac{\a \lambda^2 - \tfrac{1}{2}\s^4 G }{\big(\s^2+\tfrac{1}{2}G \big)\a\lambda^2} \\
h(G)&\bydef \bigg ( \frac{\mu}{G(\s^2G+1)} -\s^2 \bigg)\frac{1}{\a\lambda^2}.
\end{align}
The above two equations are derived from $\frac{\dif Q^2(t)}{\dif t}=0$ and $\frac{\dif G(t)}{\dif t}=0$.
In \fref{protraits}, the functions $f(G)$ and $h(G)$ are plotted as the green and red dashed lines, respectively.

It is easy to verify from their definitions that $f(G)$ and $h(G)$ are both monotonically decreasing in the feasible region ($0\leq Q^2 \leq 1$ and $G>0$). Moreover,   $0=f^{-1}(1)<h^{-1}(1)$, where $f^{-1}$ and $h^{-1}$ denote the inverse function of $f$ and $h$, respectively. Thus, a solution in Case III exists if $f^{-1}(0) > h^{-1}(0)$, which then leads to \eqref{eq:mu-c} after some algebraic manipulations. 

Finally, we examine the local stability of the fixed points in  Case I and Case II.
In both cases, a fixed point $(Q^\ast, G^\ast)$ of the 2-dimensional ODE \eqref{eq:ODE-1d-Q} and \eqref{eq:ODE-1d-G}
is stable if and only if
\[
\frac{\partial}{\partial [Q^2]}\left[\frac{\dif }{\dif t}Q^2(t)\right] \Bigg \vert_{Q=Q^\ast, G=G^\ast}<0
\]
and 
\[
\frac{\partial}{\partial G}\left[\frac{\dif }{\dif t}G(t)\right] \Bigg \vert _{Q=Q^\ast, G=G^\ast}<0,
\]
where $\frac{\dif }{\dif t}Q^2(t)$ and $\frac{\dif }{\dif t}G(t)$ are the functions on  the right-hand side of \eqref{eq:ODE-1d-Q} and \eqref{eq:ODE-1d-G}, respectively.
It follows that all the Case I fixed points are always unstable, because
$\frac{\partial}{\partial G}\big[\frac{\dif }{\dif t}G(t)\big]\big\vert _{G=0}=\mu>0.$
Furthermore, the Case II fixed point is also unstable if \eqref{eq:mu-c} holds,
because 
\[
\frac{\partial}{\partial [Q^2]}\left[\frac{\dif }{\dif t}Q^2(t)\right] \Bigg \vert_{Q=0, G=G^\ast}
=2\a \lambda^2 - G^\ast>0,
\]
where $G^\ast$ is the value specified in \eqref{eq:G-II}. On the other hand, when \eqref{eq:mu-c}  does not hold, the Case II fixed point becomes stable.
\end{IEEEproof}



\begin{figure}[t]
\centering{}\quad\includegraphics[width=\linewidth]{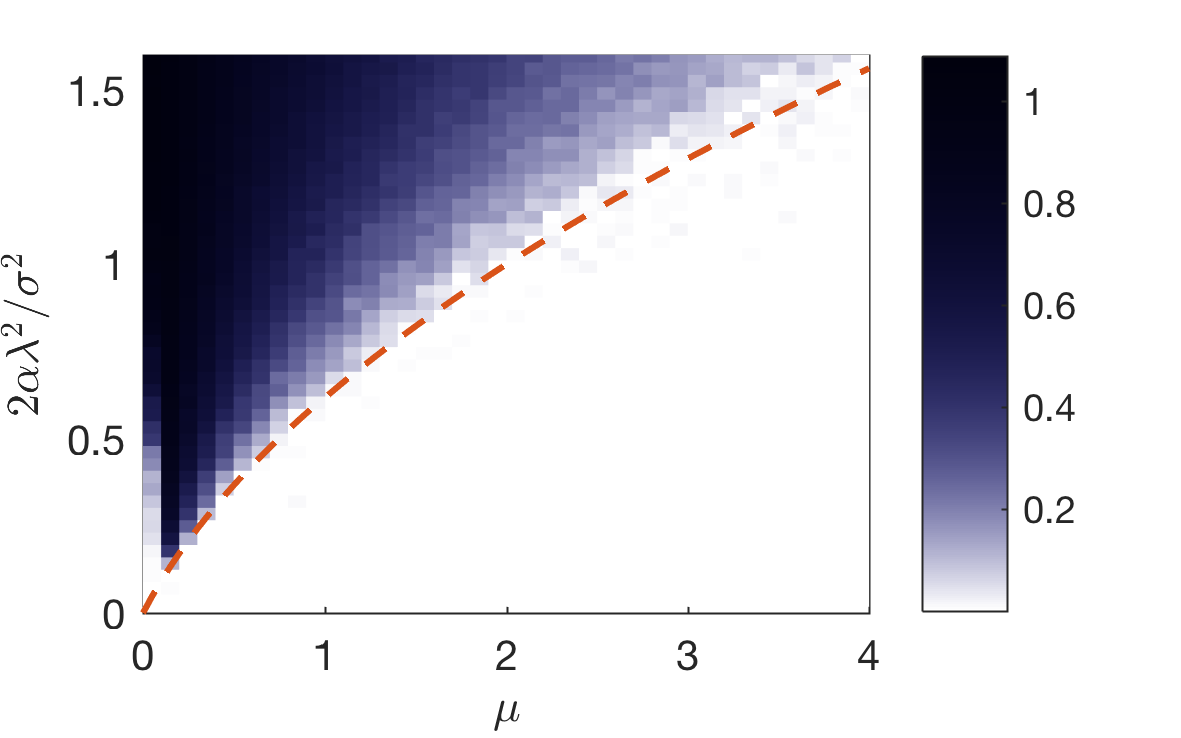}\caption{\label{fig:phase}The grayscale in the figure visualizes the steady-state squared cosine similarities of  PETRELS corresponding to different values of the SNR $\lambda^2$, the subsampling ratio $\alpha$, and the step-size parameter $\mu$. The red curve is the theoretical prediction given in Proposition~\ref{prop:phase_PETRELS} of a phase transition boundary, below which no informative solution can be achieved by the algorithm. The theoretical prediction matches numerical results.}
\end{figure}

\begin{example}
Proposition~\ref{prop:phase_PETRELS} predicts a critical choice of  
$\mu$ (as a function of the SNR $\lambda$ and the subsampling ratio $\alpha$) that separates informative solutions from noninformative ones. This prediction is confirmed numerically in Figure \ref{fig:phase}. In our experiments, we set $d = 1$, $n = 10,000$. We then scan the parameter space of $\mu$ and $\alpha \lambda^2/\s^2$. For each choice of these two parameters on our search grid, we perform 100 independent trials, with each trial using a different realizations of $\vc_k$ and $\va_k$ in \eref{gen} and a different $\mU$ drawn uniformly at random from the $n$-D sphere. The grayscale in \fref{phase} shows the average value of the squared cosine similarity $Q(t)$ at $t = 10^3$.
\end{example}


\section{Derivations of the ODEs and Proof Sketches} \label{sec:formal}

In this section, we present a nonrigorous derivation of the limiting ODEs and sketch the main ingredients of our proofs of Theorems~\ref{thm:ode-d1} and \ref{thm:limit}. More technical details and the complete proofs can be found in the Supplementary Materials \cite{WangEL:18a}.

\subsection{Derivations of the ODE} \label{sec:ode}

In what follows, we show how one may derive the limiting ODE in Theorem~\ref{thm:ode-d1}. We focus on GROUSE, but the other two algorithms can be treated similarly. 

Our first observation is that the dynamic of GROUSE can be modeled by a 
Markov chain $( \mX_{k}, \mU_{k})$ on $ \R^{2dn}$, where $ \mU_{k}\equiv \mU$
for all $k$. The update rule of this Markov chain is 
\begin{equation}
 \mX_{k+1}- \mX_{k}= \Big[  \tfrac{\left( \cos( \theta_{k})-1 \right) \vp_{k}}{ \norm{ \vp_{k}}}+  \tfrac{\sin( \theta_{k}) \vr_{k}}{ \norm{ \vr_{k}}} \Big]  \tfrac{\h \vw_k^\T}{\norm{\h \vw_k} }\charfn_{\mathcal{A}_k},
\label{eq:mc-1d}
\end{equation}
where the vectors $ \vr_{k}$ and $\vp_k$ are defined in \eqref{eq:def-r} and \eqref{eq:def-p}, 
and 
\begin{align}
\mathcal{A}_k &=  \{\,\lambda_{\min} (\mX_k^\T \mOm_k \mX_k) > \epsilon\} \nonumber
\\
\h \vw_k &= (\mX_k^\T \mOm_k \mX_k)^{-1}  \mX_k^\T \mOm_k \vs_k. \label{eq:hw}
\end{align} 
The indicator function $\charfn_{\mathcal{A}_k}$ in \eqref{eq:mc-1d} encodes the test in line \ref{ln:check-grouse} of Algorithm \ref{alg:grouse}. This test  guarantees that the matrix inverse in \eqref{eq:hw} exists.
Multiplying both sides of \eqref{eq:mc-1d} from the left by $\mU^\T$, we get
\begin{equation}\label{eq:Delta}
\mQ_{k+1} - \mQ_k = \tfrac{1}{n} \boldsymbol{ \mathcal{ G}} _k,
\end{equation}
where
\begin{equation}
 \boldsymbol{ \mathcal{ G}} _k \bydef n \Big[ \tfrac{ \left( \cos( \theta_{k})-1 \right) \mU^\T \vp_{k}}{ \norm{ \vp_{k}}}+  \tfrac{ \sin( \theta_{k}) \mU^\T  \vr_{k}}{ \norm{ \vr_{k}}} \Big] 
 \tfrac{\h \vw_k^\T}{\norm{\h \vw_k} }
 \charfn_{\mathcal{A}_k}
\label{eq:grouse-update-Q}
\end{equation}
specifies the increment of the cosine similarity from $\mQ_k$ to $\mQ_{k+1}$. 

To derive the limiting ODE, we first rewrite \eref{Delta} as
\begin{equation}\label{eq:decomposition}
\frac{\mQ_{k+1} - \mQ_k}{1/n} = \EEk \boldsymbol{ \mathcal{ G}} _k + ( \boldsymbol{ \mathcal{ G}}_k - \EEk \boldsymbol{ \mathcal{ G}} _k),
\end{equation}
where $\EEk$ denotes conditional expectation with respect to all the random elements encountered up to step $k-1$, \emph{i.e.}, $\set{\vc_j, \va_j, \mOm_j}_{0 \le j \le k-1}$ in the generative model \eref{gen} and the initial state $(\mX_0, \mU)$. One can show that
\begin{equation}\label{eq:martingale}
\EE \norm{ \boldsymbol{ \mathcal{ G}} _k - \EEk  \boldsymbol{ \mathcal{ G}} _k}^2 = \mathcal{O}(1)
\end{equation}
and
\begin{equation}\label{eq:drift}
\EEk  \boldsymbol{ \mathcal{ G}} _k= F(\mQ_k, \tau_k \mI_d) + \mathcal{O}(1/\sqrt{n}),
\end{equation}
where $F(\cdot, \cdot)$ is the function defined in \eref{def-FQG}, and $\norm{\cdot}$ denotes the spectral norm.  
Substituting \eref{drift} into \eref{decomposition} and omitting the zero-mean difference term $( \boldsymbol{ \mathcal{ G}} _k - \EEk  \boldsymbol{ \mathcal{ G}} _k)$, we get
\begin{equation} \label{eq:leading-Q}
\frac{  \mQ_{k+1}-\mQ_{k}}{1/n} = F(\mQ_{k}, \tau_{k}\mI_d)+ \mathcal{O}(1/\sqrt{n}).
\end{equation}
Let $\mQ(t)$ be a continuous-time process defined as in \eref{embed-Q}, with $t = k/n$ being the rescaled time. In an intuitive but nonrigorous way, we have $ \frac{  \mQ_{k+1}-\mQ_{k}}{1/n} \to \frac{\dif }{\dif t}\mQ(t)$ as $n \to \infty$. This then gives us the ODE in \eref{ODE-Oja}.

In what follows, we provide some additional details behind the estimate in \eref{drift}. To simplify our presentation, we first introduce a few variables.  Let 
\begin{equation} \label{eq:def-k}
\begin{aligned}
\mZ_{k} & \bydef {\mX_k^\T \mOm_{k} \mX_{k}}, \qquad 
\t z_{k} \bydef \tfrac{1}{n} \norm{ \mOm_{k} \vs_{k}}^{2},
\\
\t \vq_{k} & \bydef \mU^{ \T} \mOm_{k} \vs_{k}, \qquad \vq_{k} \bydef \mX_{k}^{ \T} \mOm_{k} \vs_{k}, 
\\
\t \mQ_{k} & \bydef \mU^{ \T} \mOm_{k} \mX_{k}, 
\end{aligned} 
\end{equation}
where $\vs_k$ and $\mOm_k$ are defined in \eqref{eq:gen} and \eqref{eq:def-Omega}.
Since the columns of $\mU$ and $\mX_k$ are unit vectors, all these variables in \eqref{eq:def-k} are $ \mathcal{O}(1)$
quantities when $n \to \infty$. (See Lemma \ref{lem:bound-vars} in Supplementary Materials.)

Given its definition in \eref{def-h}, we rewrite $\theta_k$ used in \eqref{eq:mc-1d} as
\begin{equation} \label{eq:theta-k}
\theta_{k}^{2}= \frac{\tau_{k}^{2}}{n} \vq_k^\T \mZ_k^{-2} \vq_k 
\Big[\t z_{k}- \frac{1}{n} \vq_k^\T \mZ_k^{-1} \vq_k  \Big],
\end{equation}
which is an $\mathcal{O}(1/n)$ quantity.
Thus, it is natural to expand the two terms $ \cos( \theta_{k})$ and $ \sin( \theta_{k})$ that appear in (\ref{eq:mc-1d}) via a Taylor series expansion, which yields
\begin{equation}
\label{eq:expand-theta}
\begin{aligned}
\cos( \theta_{k}) & =1- \tfrac{1}{2}\theta_k^2 + \tfrac{1}{4!}\cos(\varphi) \theta_k^4
\\
\sin( \theta_{k}) & = \theta_k - \tfrac{1}{3!} \cos(\t \varphi) \theta_k^3. 
\end{aligned}
\end{equation}
Here, the last terms in the above expansions are the remainders of the Taylor expansions, in which $\varphi$ and $\t \varphi$ are some numbers between 0 and $\theta_k$.
Substituting \eqref{eq:theta-k} and \eqref{eq:expand-theta} into \eqref{eq:grouse-update-Q} gives 
\begin{equation}\label{eq:update-Q2}
\begin{aligned}
 \boldsymbol{ \mathcal{ G}} _k =& \tau_{k}
 \big[\t \vq_{k} \vq_{k}^\T - \big( \t \mQ_{k}+ \tfrac{ \tau_{k}}{2} \t z_{k}\mQ_{k} \big) 
 \mZ_k^{-1} \vq_k \vq_k^\T 
 \big] 
 \mZ_k^{-1}\charfn_{\mathcal{A}_k}
 \\
 &+ \mathcal{O}(n^{-1/2}).
 \end{aligned}
\end{equation}
A rigorous justification of this step is presented as Lemma \ref{lem:update-Q} in the Supplementary Materials.

One can show that, as $n \to \infty$,  $\mZ_k$ and $\t z_k$ defined in \eqref{eq:def-k} converge to $\a\mI_d$ and $\a \s^2$ respectively, the quantity $\t \mQ_k \to \a \mQ_k$, and $\charfn_{\set{z_{k}> \epsilon}} \to 1$ (for any $\epsilon < \a$). Furthermore, 
\begin{align*}
\EE \norm{ \EEk \vq_{k}\vq_k^\T- \a \big(\s^2 \mI_d+  \a \mQ_k^\T \mat{\Lambda}^2 \mQ_k \big)} \le C/n, \\
\EE \norm{ \EEk \t \vq_{k} \vq_{k}^\T - \a (\s^2\mI_d+  \a\mat{\Lambda}^2)\mQ_{k}} \le C/n,
\end{align*}
for some constant $C$, where $\norm{\mM}$ is the spectral norm for a matrix of $\mM$. (The convergence of these variables is established in Lemma~\ref{lem:conv-var} in the Supplementary Materials.) Finally, by substituting the limiting values of the variables $\mZ_k, \t z_k, \t \mQ_k, \vq_k\vq_k^\T, \t \vq_k \vq_k^\T$  into  \eqref{eq:update-Q2}, we reach the estimate in \eref{drift}.

\subsection{Main Steps of Our Proofs} \label{sec:sketch}
The proofs of Theorems \ref{thm:ode-d1} and \ref{thm:limit} follow a standard argument for proving the weak convergence of stochastic processes \cite{Meleard:1987, Sznitman:1991,billingsley2013convergence}. For example, to establish the scaling limit of Oja's method and GROUSE as stated in Theorem \ref{thm:ode-d1}, our proof consists of three main steps.

First we show that the sequence of stochastic processes $\{\{ \mQ^{(n)}(t)\}_{0\leq t \leq T} \}_{n=1,2,\ldots}$ 
indexed by $n$ is tight. 
The tightness property then guarantees that any such sequence must have a 
converging sub-sequence. Second, we prove that the limit of any converging (sub)-sequence must be a solution of the ODE \eqref{eq:ODE-Oja}. Third, we show that the ODE  
\eqref{eq:ODE-Oja} admits a \emph{unique} solution. This last property can be easily established from the fact that the function 
$F(\cdot, \cdot)$ on the right-hand side of  \eqref{eq:ODE-Oja} is a Lipschitz function 
(noting that $\abs{Q_{i,j}(t)}\leq 1$ given the initial condition $\abs{Q_{i,j}(0)}\leq 1$, where $Q_{i,j}(t)$ is the entry of $\mQ(t)$ at the $i$th row and $j$th column). 
Combining the above three steps, we may then conclude that the entire sequence of stochastic processes
$\{\{ \mQ^{(n)}(t)\}_{0\leq t \leq T} \}_{n=1,2,\ldots}$ must converge weakly to the unique solution of the ODE.


\section{Conclusion}
\label{sec:conclusion}

In this paper, we present a high-dimensional analysis of three popular algorithms, namely, Oja's method, GROUSE, and PETRELS, for estimating and tracking a low-rank subspace from streaming and incomplete observations. We show that, with proper time scaling, the time-varying trajectories of estimation errors of these methods converge weakly to deterministic functions of time. Such scaling limits are characterized as the unique solutions of certain ordinary differential equations (ODEs). Numerical simulations verify the accuracy of our asymptotic results.  In addition to providing asymptotically exact performance predictions, our high-dimensional analysis yields several insights regarding the connections (and differences) between the three methods. Analyzing the limiting ODEs also reveals and characterizes phase transition phenomena associated with the steady-state performance of these techniques.

\bibliographystyle{IEEEtran}

\begin{thebibliography}{10}
\providecommand{\url}[1]{#1}
\csname url@samestyle\endcsname
\providecommand{\newblock}{\relax}
\providecommand{\bibinfo}[2]{#2}
\providecommand{\BIBentrySTDinterwordspacing}{\spaceskip=0pt\relax}
\providecommand{\BIBentryALTinterwordstretchfactor}{4}
\providecommand{\BIBentryALTinterwordspacing}{\spaceskip=\fontdimen2\font plus
\BIBentryALTinterwordstretchfactor\fontdimen3\font minus
  \fontdimen4\font\relax}
\providecommand{\BIBforeignlanguage}[2]{{%
\expandafter\ifx\csname l@#1\endcsname\relax
\typeout{** WARNING: IEEEtran.bst: No hyphenation pattern has been}%
\typeout{** loaded for the language `#1'. Using the pattern for}%
\typeout{** the default language instead.}%
\else
\language=\csname l@#1\endcsname
\fi
#2}}
\providecommand{\BIBdecl}{\relax}
\BIBdecl

\bibitem{Balzano:2010}
L.~Balzano, R.~Nowak, and B.~Recht, ``Online {identification} and {tracking} of
  {subspaces} from {highly} {incomplete} information,'' in \emph{Proc. Allerton
  Conference on Communication, Control and Computing.}, 2010.

\bibitem{Chi2013}
Y.~Chi, R.~Calderbank, and Y.~C. Eldar, ``{PETRELS: Parallel subspace
  estimation and tracking by recursive least squares from partial
  observations},'' \emph{IEEE Transactions on Signal Processing}, vol.~61,
  no.~23, pp. 5947--5959, 2013.

\bibitem{Oja1982}
E.~Oja, ``{Simplified neuron model as a principal component analyzer},''
  \emph{Journal of Mathematical Biology}, vol.~15, no.~3, pp. 267--273, 1982.

\bibitem{Balzano:2015}
L.~Balzano and S.~J. Wright, ``Local {convergence} of an {algorithm} for
  {subspace} {identification} from {partial} {data},'' \emph{Foundations of
  Computational Mathematics}, vol.~15, no.~5, pp. 1279--1314, Oct. 2015.

\bibitem{ZhangB:2016}
D.~Zhang and L.~Balzano, ``Convergence of a {Grassmannian} {gradient} {descent}
  {algorithm} for {subspace} {estimation} {from} {undersampled} {data},''
  \emph{arXiv:1610.00199}, 2016.

\bibitem{Zhang2015}
------, ``{Global Convergence of a Grassmannian Gradient Descent Algorithm for
  Subspace Estimation},'' in \emph{Proceedings of the 19th International
  Conference on Artificial Intelligence and Statistics (AISTATS)}, vol.~51,
  2016, pp. 1460--1468.

\bibitem{gonen2016subspace}
A.~Gonen, D.~Rosenbaum, Y.~C. Eldar, and S.~Shalev-Shwartz, ``{Subspace
  Learning with Partial Information},'' \emph{Journal of Machine Learning
  Research}, vol.~17, pp. 1--21, 2016.

\bibitem{Meleard:1987}
S.~Meleard and S.~Roelly-Coppoletta, ``A propagation of chaos result for a
  system of particles with moderate interaction,'' \emph{Stochastic Processes
  and their Applications}, vol.~26, pp. 317--332, Jan. 1987.

\bibitem{Sznitman:1991}
A.-S. Sznitman, ``\BIBforeignlanguage{en}{Topics in propagation of chaos},'' in
  \emph{\BIBforeignlanguage{en}{Ecole d'{Et{\'e}} de {Probabilit{\'e}s} de
  {Saint}-{Flour} {XIX} --- 1989}}, ser. Lecture {Notes} in {Mathematics},
  P.-L. Hennequin, Ed.\hskip 1em plus 0.5em minus 0.4em\relax Springer Berlin
  Heidelberg, 1991, no. 1464, pp. 165--251.

\bibitem{EthierK:85}
S.~N. Ethier and T.~G. Kurtz, \emph{Markov Processes: Characterization and
  Convergence}.\hskip 1em plus 0.5em minus 0.4em\relax Wiley, 1985.

\bibitem{billingsley2013convergence}
P.~Billingsley, \emph{{Convergence of probability measures}}.\hskip 1em plus
  0.5em minus 0.4em\relax John Wiley {\&} Sons, 2013.

\bibitem{Jacod:2010}
J.~Jacod and A.~Shiryaev, \emph{Limit theorems for stochastic processes}.\hskip
  1em plus 0.5em minus 0.4em\relax Springer, 2003, vol. 288.

\bibitem{WangL:16}
C.~Wang and Y.~M. Lu, ``Online learning for sparse pca in high dimensions:
  Exact dynamics and phase transitions,'' in \emph{Proc. IEEE Information
  Theory Workshop (ITW)}, Cambridge, UK, Sep. 2016.

\bibitem{Wang2017}
------, ``{The Scaling Limit of High-Dimensional Online Independent Component
  Analysis},'' in \emph{Advances in Neural Information Processing Systems},
  2017.

\bibitem{Wang2017c}
C.~Wang, J.~Mattingly, and Y.~M. Lu, ``{Scaling Limit: Exact and Tractable
  Analysis of Online Learning Algorithms with Applications to Regularized
  Regression and PCA},'' \emph{arXiv:1712.04332}, 2017.

\bibitem{WangEL:18a}
\BIBentryALTinterwordspacing
C.~Wang, Y.~C. Eldar, and {Y. M. Lu}, ``Subspace estimation from incomplete
  observations: A high-dimensional analysis ({Main Text and Supplementary
  Materials}),'' \emph{{arXiv:1805.06834}}, 2018. [Online]. Available:
  \url{https://arxiv.org/abs/1805.06834}
\BIBentrySTDinterwordspacing

\bibitem{Balzano2018}
L.~Balzano, Y.~Chi, and Y.~M. Lu, ``{A Modern Perspective on Streaming PCA and
  Subspace Tracking: The Missing Data Case},'' \emph{Proceedings of the IEEE},
  no. to appear, 2018.

\bibitem{Yang1995}
B.~Yang, ``{Projection Approximation Subspace Tracking},'' \emph{IEEE
  Transactions on Signal Processing}, vol.~43, no.~1, pp. 95--107, 1995.

\bibitem{Ipsen1995}
I.~C.~F. Ipsen and C.~D. Meyer, ``{The Angle Between Complementary
  Subspaces},'' \emph{The American Mathematical Monthly}, vol. 102, no.~10, pp.
  904--911, 1995.

\bibitem{Deutsch1995}
F.~Deutsch, ``{The angle between subspaces of a Hilbert space},'' in
  \emph{Approximation theory, wavelets and applications}.\hskip 1em plus 0.5em
  minus 0.4em\relax Springer, 1995, pp. 107--130.

\bibitem{Billingsley:1999}
P.~Billingsley, \emph{Convergence of probability measures}, 2nd~ed., ser. Wiley
  series in probability and statistics. {Probability} and statistics
  section.\hskip 1em plus 0.5em minus 0.4em\relax New York: Wiley, 1999.

\bibitem{Mitliagkas2013}
I.~Mitliagkas, C.~Caramanis, and P.~Jain, ``{Memory limited, streaming PCA},''
  in \emph{Advances in Neural Information Processing Systems}, 2013, pp.
  2886--2894.

\bibitem{hardt2014noisy}
M.~Hardt and E.~Price, ``The noisy power method: A meta algorithm with
  applications,'' in \emph{Advances in Neural Information Processing Systems},
  2014, pp. 2861--2869.

\bibitem{balcan2016improved}
M.-F. Balcan, S.~S. Du, Y.~Wang, and A.~W. Yu, ``An improved gap-dependency
  analysis of the noisy power method,'' in \emph{Conference on Learning
  Theory}, 2016, pp. 284--309.

\bibitem{li2016near}
C.~J. Li, M.~Wang, H.~Liu, and T.~Zhang, ``Near-optimal stochastic
  approximation for online principal component estimation,'' \emph{arXiv
  preprint arXiv:1603.05305}, 2016.

\bibitem{jain2016streaming}
P.~Jain, C.~Jin, S.~M. Kakade, P.~Netrapalli, and A.~Sidford, ``Streaming pca:
  Matching matrix bernstein and near-optimal finite sample guarantees for oja's
  algorithm,'' in \emph{Conference on Learning Theory}, 2016, pp. 1147--1164.

\bibitem{balsubramani2013fast}
A.~Balsubramani, S.~Dasgupta, and Y.~Freund, ``The fast convergence of
  incremental pca,'' in \emph{Advances in Neural Information Processing
  Systems}, 2013, pp. 3174--3182.

\bibitem{shamir2016convergence}
O.~Shamir, ``Convergence of stochastic gradient descent for pca,'' in
  \emph{International Conference on Machine Learning}, 2016, pp. 257--265.

\bibitem{allen2017first}
Z.~Allen-Zhu and Y.~Li, ``First efficient convergence for streaming k-pca: a
  global, gap-free, and near-optimal rate,'' in \emph{Foundations of Computer
  Science (FOCS), 2017 IEEE 58th Annual Symposium on}.\hskip 1em plus 0.5em
  minus 0.4em\relax IEEE, 2017, pp. 487--492.

\bibitem{Mitliagkas2014}
I.~Mitliagkas, C.~Caramanis, and P.~Jain, ``{Streaming PCA with many missing
  entries},'' \emph{Preprint}, 2014.

\bibitem{krishnamurthy2013low}
A.~Krishnamurthy and A.~Singh, ``Low-rank matrix and tensor completion via
  adaptive sampling,'' in \emph{Advances in Neural Information Processing
  Systems}, 2013, pp. 836--844.

\bibitem{jin2016provable}
C.~Jin, S.~M. Kakade, and P.~Netrapalli, ``Provable efficient online matrix
  completion via non-convex stochastic gradient descent,'' in \emph{Advances in
  Neural Information Processing Systems}, 2016, pp. 4520--4528.

\bibitem{lois2015online}
B.~Lois and N.~Vaswani, ``Online matrix completion and online robust pca,'' in
  \emph{Information Theory (ISIT), 2015 IEEE International Symposium on}.\hskip
  1em plus 0.5em minus 0.4em\relax IEEE, 2015, pp. 1826--1830.

\bibitem{Kurtz:1970lr}
T.~G. Kurtz, ``Solutions of {ordinary} {differential} {equations} as {limits}
  of {pure} {jump} {Markov} {processes},'' \emph{Journal of Applied
  Probability}, vol.~7, no.~1, p.~49, Apr. 1970.

\bibitem{Ljung:1977}
L.~Ljung, ``Analysis of recursive stochastic algorithms,'' \emph{IEEE
  Transactions on Automatic Control}, vol.~22, no.~4, pp. 551--575, 1977.

\bibitem{Yang:1996}
B.~Yang, ``Asymptotic convergence analysis of the projection approximation
  subspace tracking algorithms,'' \emph{Signal Processing}, vol.~50, no. 1--2,
  pp. 123--136, Apr. 1996.

\bibitem{candes2009exact}
E.~J. Cand{\`e}s and B.~Recht, ``Exact matrix completion via convex
  optimization,'' \emph{Foundations of Computational mathematics}, vol.~9,
  no.~6, p. 717, 2009.

\bibitem{Bossy1997}
M.~Bossy and D.~Talay, ``{A stochastic particle method for the McKean-Vlasov
  and the Burgers equation},'' \emph{Mathematics of Computation of the American
  Mathematical Society}, vol.~66, no. 217, pp. 157--192, 1997.

\end{thebibliography}


\newpage

\renewcommand{\theequation}{S-\arabic{equation}}
\renewcommand \thesection {S-\Roman{section}}
\setcounter{section}{0}
\setcounter{equation}{0}

{
\center{{ \bf Supplementary Materials } }

$ $
}

\section{Outline of the Proofs of Theorems \ref{thm:ode-d1}--\ref{thm:finite}} \label{sec:gen-conv}

In the supplementary materials, we prove Theorems \ref{thm:ode-d1}--\ref{thm:finite} stated in the main text.

Section \ref{sec:limit} dicusses the convergence of a general sequence of stochastic processes.
In particular, we prove two lemmas: Lemma \ref{thm:weak} is a generalized version of Theorems \ref{thm:ode-d1} and \ref{thm:limit}, and
Lemma \ref{thm:gen-finite} is a generalized version of Theorem \ref{thm:finite}.
We provide a set of sufficient conditions \ref{c:m}--\ref{c:init} for the two lemmas to hold. Once we have proved Lemma~\ref{thm:weak} and Lemma~\ref{thm:gen-finite}, the remaining task is to show that the specific stochastic processes associated with Oja's method, GROUSE,  and PETRELS satisfy this set of sufficient conditions.
 
In Section  \ref{sec:useful-ineq}, we  provide some useful inequalities that will be used later, and  
 in Section \ref{sec:gen-ineq} we prove some common inequalities that  only depend on the generative model \eqref{eq:gen} .
 
 In Section \ref{sec:grouse}, we prove two lemmas that are specialized for GROUSE.
In Section \ref{sec:oja}, we prove their counterparts specialized for Oja's method. Then, in Section \ref{sec:grouse-oja}, we prove that Condition \ref{c:m}--\ref{c:init} are satisfied
 for both GROUSE and Oja's method using the lemmas proved in Section \ref{sec:grouse} (for GROUSE)
 and Section \ref{sec:oja} (for Oja's method). This section completes the proof of Theorems 1 and 3 for these two algorithms.  

 In Section \ref{sec:petrels}, we prove that Condition \ref{c:m}--\ref{c:init} are also satisfied for PETRELS.
This then completes the proof of Theorems 2 and 3 for PETRELS.
 
 Finally,  the two  lemmas claimed in Section \ref{sec:limit} are proved in Section \ref{sec:gen-proof}.

\section{Deterministic Scaling Limit of Stochastic Processes} \label{sec:limit}
In this section, we  provide a set of sufficient conditions
on when a general stochastic process converges to a deterministic process.
Theorems \ref{thm:ode-d1}--\ref{thm:finite} in the main text will be
proved on top of this result in subsequent sections. 


Consider a sequence of $d$-dimensional discrete-time stochastic processes
  $\{\vq^{(n)}_k\in\R^d$, $k=0,1,2,\ldots, \nT\}_{n=1,2,\ldots}$, with  some constant $T>0$.
 We assume that  the increment $\vq^{(n)}_{k+1} - \vq^{(n)}_k$ can be decomposed into three parts 
  \begin{equation} \label{eq:q-decomp}
   \vq^{(n)}_{k+1} - \vq^{(n)}_k=\tfrac{1}{n} L(\vq_k^{(n)}) + \vm^{(n)}_k + \vr^{(n)}_k
   \end{equation}
     such that 
\begin{enumerate}[label={(C.\arabic*)}]
  \item \label{c:m} The process  $\vu_k^{(n)} \bydef \sum_{k^\prime=0}^k\vm_{k^\prime}^{(n)}$ is a martingale,  and $\EE \norm{\vmn_k}^2\leq C(T)/n^{1+\epsilon_1}$ for some positive $\e_1$.
  \item \label{c:r} $\EE \norm{\vr_k^{(n)}}\leq C(T)/n^{1+\e_2}$ for some positive $\e_2$.
  \item \label{c:l}$L(\vq)$ is a Lipschitz function, i.e.,  $\norm{L(\vq)- L(\t \vq)} \leq C \norm{\vq - \t \vq}$.
  \item \label{c:bound} $\lim_{b \to \infty} \limsup_n \mathbb{P}.
  \Big(\max_{k=0,1,2,\nT} \norm{\vqn_k } \geq b\Big)=0$.
  \item \label{c:init} $\EE\norm{\vqn_k}^2\leq C$ for all $k\leq \nT$.
  
\end{enumerate}
Then, we have the following two lemmas.

\begin{lemma} \label{thm:weak}
  
  Let $\vq^{(n)}(t)$, $0\leq t\leq T$ be a continuous stochastic process such that $\vq^{(n)}(t)=\vq^{(n)}_k $ with $k=\nt$.
  Assuming Conditions \ref{c:m}--\ref{c:init} hold, and  assuming that $\vq^{(n)}_0$ converges weakly to a deterministic $\vq_0$, 
  then  the sequence of $\{ \{\vq^{(n)}(t)\}_{t\in[0,T]} \}_{n=1,2,\ldots}$ converges weakly to a deterministic process $\{\vq(t)\}_{t\in[0,T]}$, which is the unique solution of the ODE
  \begin{equation} \label{eq:ode-q}
    \frac{\dif }{\dif t}\vq(t)  = L(\vq(t)), \mbox{ with } \vq(0)=\vq_0.
    \end{equation}
 \end{lemma}
The proof can be found in Section \ref{sec:gen-proof}. 
 
 \begin{lemma} \label{thm:gen-finite}
If  Conditions \ref{c:m}--\ref{c:l} hold, and
 $\EE \norm{\vq^{(n)}_0 -\vq_0} \leq C/n^{\e_3}$ 
 with a positive $\e_3$, then  for any finite $n$, we have
  \[
    \norm{\vq^{(n)}_k - \vq(\tfrac{k}{n})} \leq C(T)n^{- \min\{ \frac{1}{2}\e_1, \e_2, \e_3 \}},
    \]
    where $\vq(\cdot)$ is the solution of the ODE \eqref{eq:ode-q}.
\end{lemma}
 The proof of this lemma can be found in Section \ref{sec:gen-proof}.

\section{Some useful inequalities} \label{sec:useful-ineq}

In the proofs of Lemmas~\ref{thm:weak} and \ref{thm:gen-finite}, we repeatedly use the following inequality: for any positive
integer $ \ell$ and some variable $a_{1}$, $ \ldots$, $a_{m}$, we
have
\begin{equation}
\left( \textstyle \sum_{i=1}^{m}a_{i} \right)^{ \ell} \leq m^{ \ell-1} \textstyle \sum_{i=1}^{m} \abs{a_{i}}^{ \ell}. \label{eq:ineq-1}
\end{equation}
This is a consequence of the convexity of the function $f(x)=x^\ell$
on the interval $x \geq0$ and for $\ell\geq 1$.

The following lemmas  are also useful
in our proofs.
\begin{lemma}
Let $a_{1}$, $a_{2}$, $ \ldots$, $a_{n}$ be $n$ i.i.d. random
variables with zero mean, unit variance and bounded higher-order moments
$ \EE \abs{a_{i}}^{ \ell} \leq \infty$ for some $ \ell>2$, then

\begin{equation}
\EE \left( \tfrac{1}{n} \textstyle \sum_{i=1}^{n}a_{i}^{2} \right)^{ \ell/2} \leq \EE a_1^{ \ell}. \label{eq:ineq-3}
\end{equation}

Moreover, for any fixed vector $ \vx \in \R^{n}$,
\begin{equation}
\EE \abs{ \textstyle \sum_{i=1}^{n}x_{i}a_{i}}^{ \ell} \leq C \norm{ \vx}^{ \ell} \EE \abs{a_{1}}^{ \ell}, 
\label{eq:ineq-4}
\end{equation}
where $x_{i}$ is $i$th element of $ \vx$, and $C$ is a constant.
\end{lemma}
\begin{IEEEproof}
See the reference \cite[Lemma~6]{Wang2017c}.
\end{IEEEproof}

\begin{lemma} \label{lem:itr}
Let $n$ be a positive integer and $x_0,x_1,x_2,\ldots$ be a sequence of variables satisfying 
\begin{equation}\label{eq:ineq-xk}
\abs{x_{\t k}} \leq \left(1+\tfrac{C}{n} \right) \abs{x_{\t k-1}} + \tfrac{B}{n^{1+\ell}},
\end{equation}
with some positive constant $C$, $B$ and $\ell$. Then for  all non-negative integer  $ k\leq nT$ with some $T>0$,
\begin{equation*}
\abs{x_{ k}} \leq e^{C \cdot T}  \left(  x_0  + \tfrac{B}{C n^\ell} \right) .
\end{equation*}
\end{lemma}
\begin{IEEEproof}
For any $ k\leq nT$, we apply the inequality \eqref{eq:ineq-xk} iteratively, and get
\begin{align*}
\abs{x_{k}} 
&\leq \left(1+\tfrac{C}{n} \right) \abs{x_{k-1}} + \tfrac{B}{n^{1+\ell}}\\
&\leq \left(1+\tfrac{C}{n} \right)^2 \abs{x_{k-2}} 
+ \left(1+\tfrac{C}{n} \right)\tfrac{B}{n^{1+\ell}} 
+ \tfrac{B}{n^{1+\ell}} \\
&\leq \left(1+\tfrac{C}{n} \right)^k \abs{x_{0}} 
+ \tfrac{B}{n^{1+\ell}} 
[ (1+\tfrac{C}{n})^{k-1} + (1+\tfrac{C}{n})^{k-2} \\
&\quad + \ldots. + (1+\tfrac{C}{n}) +1] \\
&= \left(1+\tfrac{C}{n} \right)^k \abs{x_{0}} + \tfrac{B}{Cn^\ell}
\left[ {(1+\tfrac{C}{n})^{k} - 1} \right]\\
&\leq \left(1+\tfrac{C}{n} \right)^{\frac{n}{C}\cdot \frac{C}{n} k} \left( \abs{x_{0}} + \tfrac{B}{Cn^\ell} \right)
\\
&\leq e^{C\cdot T} \left( \abs{x_{0}} + \tfrac{B}{Cn^\ell} \right),
\end{align*}
where in reaching the last line we used an inequality that $(1+\tfrac{1}{a})^a < e$ for any $a>0$.
\end{IEEEproof}
\section{Inequalities on the samples $\vs_k$, and observation mask matrix $\mOm_k$}
\label{sec:gen-ineq}
We provide two lemmas showing some inequalities on the observation data $\vs_k$, and $\mOm_k$, where $\vs_k$ and $\mOm_k$ are defined in \eqref{eq:gen} and \eqref{eq:def-Omega} respectively.
We emphasis that $\mX_k$ stated in the following two lemmas can be any matrix that 
are not necessarily associated with any algorithm.

Let $\EEk$ denote the expectation w.r.t. $\vs_k$ and $\mOm_k$.
\begin{lemma}
\label{lem:bound-vars} Let $\mX_k$ be any $d\times n$ matrix such that $\mX_k^\T \mX_k=\mI_d$. The variables $\mZ_k$, $\t z_k$, $\vq_k$ and $\t \vq_k$  defined in \eqref{eq:def-k}, in which $\vs_k$ and $\mOm_k$ are defined in \eqref{eq:gen} and \eqref{eq:def-Omega} satisify
\begin{align}
\norm{\mZ_{k}} &\leq1 \label{eq:bound-Z}\\
\norm{\t \mQ_k} &\leq1 \label{eq:bound-Q} \\
\EEk \t z_{k}^{ \ell} & \leq C( \ell) \label{eq:bound-w} \\
\EEk \norm{\vq_{k}}^{ \ell} & \leq C( \ell) \label{eq:bound-q} \\
\EEk \norm{\t \vq_{k}}^{ \ell} & \leq C( \ell), \label{eq:bound-p}
\end{align}
\end{lemma}
where $\ell$ is any positive integer, and $C( \ell)$ is a constant that can depend
on $ \ell$ but not on the ambient dimension $n$ 


\begin{IEEEproof}[Proof of Lemma \ref{lem:bound-vars}]
The inequalities \eqref{eq:bound-Z} and \eqref{eq:bound-Q} are straightforward due to the submultiplicativity of the induced matrix norm, and  
$\norm{\mOm_k}\leq 1$:
\begin{align*}
\norm{\mZ_k} \leq \norm{\mX^\T_k} \cdot \norm{\mOm_k} \cdot \norm{\mX_k} \leq 1 \\
\norm{\t \mQ_k} \leq \norm{\mU^\T_k} \cdot \norm{\mOm_k} \cdot \norm{\mX_k} \leq 1.
\end{align*}

For \eqref{eq:bound-w}, we have
\begin{align*}
\EEk \t z_{k}^{\ell} & =
n^{-\ell} \; \EEk \norm{\vs_{k}}^{2\ell}
=n^{-\ell} \; \EEk \norm{\mU \vc_k + \va_k}^{2\ell}
\\
 & \leq  n^{-\ell}\; 
 \EEk \left( \normF{\mU} \cdot \norm{\vc_{k}} +\norm{\va_{k}}\right)^{2\ell}\\
 & \leq n^{-\ell} \; 2^{2\ell-1} 
 \left(\normF{\mU} ^{2\ell}\EEk \norm{\vc_{k}}^{2\ell}+\EEk\norm{\va_{k}}^{2\ell}\right)\\
 & \leq 2^{2\ell-1} \left(\tfrac{d}{n^{\ell}}\EEk c_{k}^{2\ell}+\EEk a_{k,1}^{2\ell}\right)\leq C(\ell),
\end{align*}
where  the second line is due to
$\norm{\mU\vc_k}\leq \norm{\mU}_2 \cdot \norm{\vc_k} \leq  \normF{\mU} \cdot \norm{\vc_k}$.
Furthermore, 
 in reaching the second last line, we used \eqref{eq:ineq-1}
and in reaching the last line we used \eqref{eq:ineq-3} that $\EE\norm{\va_{k}}^{2\ell}\leq n^\ell \EE a_{k,1}^{2\ell}$.

Let $\vx_{k,i}$ denote the $i$th column of $\mX_k$. For (\ref{eq:bound-q}), we have 
\begin{align*}
&\EEk\abs{[\vq_{k}]_{i}}^{\ell} 
= \EEk \abs{\vx_{k,i}^\T \mOm_k \vs_k }
 =\EEk\abs{ [\t \mQ_{k}\vc_{k}]_{i}+\vx_{k,i}^{\T}\mOm_{k}\va_{k}}^{\ell}
\\
 & \leq 2^{\ell-1}
 \left( \EE \abs{ [\t \mQ_{k}\vc_{k}]_{i} }^\ell
 +\EEk\abs{\vx_{k,i}^{\T}\mOm_{k}\va_{k}}^{\ell}\right)
 \\
 & \leq C(\ell) \left(1+\norm{\vx_{k,i}}^{\ell}\EEk\abs{v_{k,i} a_{k,i}}^{\ell}\right)\leq C(\ell),
\end{align*}
where   $v_{k,i}$ is the $i$th diagonal element in $\mOm_k$.
Here, the second last line is due to (\ref{eq:ineq-1}),
and in reaching the last line we used (\ref{eq:ineq-4}). 
Since $\vq_k$ is a $d$-dimensional vector, where $d$ is a finite number, the above inequality implies \eqref{eq:bound-q}. (Note that $C(\ell)$ depends on the fixed number $d$, but we omit $d$ for the sake of simplifying notations.)

Finally, \eqref{eq:bound-p} can be proved in the same way as we prove
\eqref{eq:bound-q}.
\end{IEEEproof}

\begin{lemma}
\label{lem:conv-var} 
Let $s_k\bydef\sum_{i=1}^n \sum_{j=1}^d [\mX_{k}]_{i,j}^4$.
Under the same setting as stated in Lemma~\ref{lem:bound-vars}, the variables defined in \eqref{eq:def-k} satisfies
\begin{align}
\EEk& \norm{\mZ_{k}- \alpha \mI_d }^{2} 
 \leq Cs_k \label{eq:conv-z} 
 \\
\EEk& \norm{ \t \mQ_{k}- \alpha \mQ_{k} }^{2} 
 \leq C(s_k +\tfrac{1}{n}) \label{eq:conv-tQ} 
 \\
\EEk& \left(\t z_{k}- \a \s^2 \right)^{2} 
 \leq C\tfrac{1}{n}  \label{eq:conv-tz}
 \\
\EE& \norm{ \EEk \vq_{k}\vq_k^\T- \a \big(\s^2 \mI_d+  \a \mQ_k^\T \mat{\Lambda}^2 \mQ_k \big)} 
 \leq C(s_k +\tfrac{1}{n}) \label{eq:conv-q2} 
 \\
\EE& \norm{ \EEk \t \vq_{k} \vq_{k}^\T - \a (\s^2\mI_d+  \a\mat{\Lambda}^2)\mQ_{k}} 
 \leq C(s_k +\tfrac{1}{n}). \label{eq:conv-pq}
\end{align}
\end{lemma}

\begin{IEEEproof} 
The proof utilizes a fact that  $v_{k,i}$, $i=1,2, \ldots, n$, the diagonal terms of $\mOm_k$, are
i.i.d. Bernoulli random variables with  mean $ \a$. 
Therefore, 
\begin{equation}
\begin{aligned}
\EEk \mOm_k&=\a \mI_n \\ 
\EEk[ \mOm_k \mA \mB \mOm_k]&=\a^2 \mA \mB + \a(1-\a) \diag( \textstyle \sum_{\ell} [\mA]_{i,\ell} [\mB]_{\ell,i})
\end{aligned} \label{eq:ineq-O}
\end{equation} for any matrices $\mA$ and $\mB$ with the consistent dimensions.


The first three inequalities are straightforward. For \eqref{eq:conv-z},
\begin{align*}
&\EEk \norm{\mZ_{k}- \alpha \mI_d }^{2} \\
&\leq\EEk \normF{\mZ_{k}- \alpha \mI_d }^{2} \\
&= \EEk \optr\big[(\mZ_{k}- \alpha \mI_d)^\T (\mZ_{k}- \alpha \mI_d)\big] \\
&= \optr \big( \mX_k^\T \EEk \big[ \mOm_k \mX_k \mX_k^\T \mOm_k \big] \mX_k - \a^2 \mI_d  \big)\\
&=\a (1-\a)\optr \big( \mX_k^\T \diag(\textstyle \sum_{\ell=1}^d [\mX_k]^2_{i,\ell}, i=1,2,\ldots,n ) \mX_k^\T \big) \\
&=\a(1-\a) \textstyle \sum_{\ell,\ell^\prime=1}^d \textstyle \sum_{i=1}^n [\mX_k]^2_{i,\ell} [\mX_k]^2_{i,\ell^\prime}\\
&\leq \tfrac{1}{2} \a(1-\a) \textstyle \sum_{\ell,\ell^\prime=1}^d \textstyle \sum_{i=1}^n \big( [\mX_k]^4_{i,\ell} + [\mX_k]^4_{i,\ell^\prime} \big)\\
&= d \a (1-\a) s_k.
\end{align*}
We can prove \eqref{eq:conv-tQ} in the same way.
For \eqref{eq:conv-tz}, 
\begin{align*}
& \EEk\left(\t z_{k}-\a\right)^{2}
 = \EEk\big[ \tfrac{1}{n}(\mU\vc_k + \va)^\T \mOm_k (\mU\vc_k + \va) -\a\big]^{2}\\
 &=\EEk\big[ (\tfrac{1}{n} \va^\T \mOm_k \va - \a) 
 + \tfrac{2}{n}\va_k^\T \mOm_k \mU \vc_k 
 + \tfrac{1}{n} \vc_k^\T \mU^\T\mOm_k \mU \vc_k \big]^2 \\
 & \leq 3 \EEk(\tfrac{1}{n} \va^\T \mOm_k \va - \a) ^2
 + \tfrac{6}{n^2} \EEk (\va_k^\T \mOm_k \mU \vc_k)^2 \\
 &\quad
 + \tfrac{3}{n^2} \EEk(\vc_k^\T \mU^\T\mOm_k \mU \vc_k)^2.
\end{align*}
Then, we can bound the threes terms in the last inequality using \eqref{eq:ineq-O} and  Assumption \ref{ass:u} in the same way as we prove \eqref{eq:conv-z}, which yields \eqref{eq:conv-tz}.

Next, we   prove (\ref{eq:conv-q2}) and (\ref{eq:conv-pq}).
By definitions \eqref{eq:def-k}, we know
\begin{align}
\EEk \vq_{k}\vq_k^\T & = \mX_{k}^{ \T} \left( \EEk \mOm_{k} \vs_{k} \vs_{k}^{ \T} \mOm_{k} \right) \mX_{k} \label{eq:Eq2} \\
\EEk \t \vq_{k}\vq_{k}^\T & = \mU^{ \T} \left( \EEk \mOm_{k} \vs_{k} \vs_{k}^{ \T} \mOm_{k} \right) \mX_{k}, \label{eq:Epq}
\end{align}
where $\vs_k$ is defined in \eqref{eq:gen}.
From the definition of the generative model \eqref{eq:gen} and \eqref{eq:def-Omega}, we can explicitly compute the average  
\begin{equation}
\begin{aligned} & \EEk \mOm_{k} \vs_{k} \vs_{k}^{ \T} \mOm_{k} \\
 & = \a \big( \s^2 \mI_d + \a  \mU \mat{\Lambda}^2 \mU^{ \T} \big) \\
 &\quad 
 +  \a(1- \a) \diag \big( \textstyle \sum_{\ell=1}^d \lambda^{2}_\ell [\mU]_{i,\ell}^{2}, i=1,2,\ldots,n \big).
\end{aligned}
\label{eq:Eyy}
\end{equation}

Substituting \eqref{eq:Eyy} into \eqref{eq:Eq2}, for $\ell,\ell^\prime = 1,2,\ldots,d$, we have
\begin{align*}
&\abs{\EEk 
\big[ \vq_{k}\vq_k^\T-\a\big(\s^2 \mI_d+\a \mQ_k^\T \mat{\Lambda}^2\mQ_{k}\big)\big]_{\ell,\ell^\prime} } \\
& =\a(1-\a)\textstyle \sum_{i} \lambda^{2}_\ell [\mX_k]_{i,\ell}^{2} [\mU]_{i,\ell}^{2}\\
& \leq \tfrac{1}{2} \a(1-\a)\textstyle \sum_{i=1}^n \textstyle \sum_{\ell=1}^d \lambda^{2}_\ell \big( [\mX_k]_{i,\ell}^{4} +  [\mU]_{i,\ell}^{4} \big)\\
 & \leq\tfrac{1}{2}\a(1-\a)\left(s_{k}+\tfrac{C}{n} \right) \max_{\ell=1,2,\ldots,d}(\lambda^{2}_\ell)
 \\
 & \leq C(s_k + \tfrac{1}{n})
  ,
\end{align*}
where in reaching the second last line, we used Assumption \ref{ass:u}. 
This inequality implies \eqref{eq:conv-q2}. 
In the same way, we can prove \eqref{eq:conv-pq}.
\end{IEEEproof}

\section{Dynamics of GROUSE} \label{sec:grouse}
We first show some extra details of the formal derivation that has been briefly introduced in 
Section \ref{sec:ode} in the main text. Then, we present two lemmas:
 Lemma \ref{lem:update-Q} bounds the remainder term in the increment of $\mQ_{k+1}-\mQ_k$,
 and Lemma \ref{lem:bound-s} shows that 
  the 4th moment  $s_k$ in Lemma \ref{lem:conv-var} is bounded by $C(T)n^{-1}$. 

We first recall that  $( \mX_{k}, \mU)$ forms a  Markov chain whose update rule is given in \eqref{eq:mc-1d}. It is driven by 
$\mOm_k$,   $\vc_{k}$,
$\va_k$ and the initial states
$( \mX_{0}, \mU)$. We denote the expectation over all these random
variables by $ \E[ \cdot]$, and denote by $ \E_{k}[ \cdot]= \EE[ \cdot| \mX_{k}, \mU]$
the conditional expectation w.r.t. $\vc_k$ and $\va_k$ given $(\mX_k, \mU)$ at step $k$.
(The definition $\EE[ \cdot| \vx_{k}, \vu]$ is consistent with  the notation $\EEk$ defined in the previous section.)

\subsection{Formal derivation of the ODE for GROUSE}
We here provide additional details of the formal derivation, which has been introduced in 
Section \ref{sec:ode} in the main text.

From \eqref{eq:hw} and \eqref{eq:def-k}, we have 
\[
\h \vw_k=\mZ_k^{-1}\vq_k.
\]
Then, with \eqref{eq:def-r} and \eqref{eq:def-p}, we have  
\[
\begin{aligned}
\norm{\vr_k}&=n\t z_{k} - \vq_k^\T \mZ_k^{-1} \vq_k \\
\vp_k &= \mX_k \mZ_k^{-1} \vq_k.
\end{aligned}
\] 
Substituting the above equations into \eqref{eq:def-h}, we get  the expression of 
$\theta_k$ shown in \eqref{eq:theta-k}.
Next, substituting Taylor's expansions  \eqref{eq:expand-theta} 
into \eqref{eq:mc-1d}, we can rewrite (\ref{eq:mc-1d})
as
\begin{equation}
\begin{aligned}
\mX_{k+1}-\mX_{k} =& \tfrac{\tau_{k}}{n}
 \big[\mOm_k \vs_k \vq_{k}^\T - \big( \mOm_k+ \tfrac{ \tau_{k}}{2} \t z_{k}\mI_d \big) \mX_k
 \mZ_k^{-1} \vq_k \vq_k^\T 
 \big] 
 \\
 & \mZ_k^{-1}\charfn_{\mathcal{A}_k}+ \mD_k,
 \end{aligned} \label{eq:mc-simplified}
 \end{equation}
where $\mZ_k,\t z_k, \t \vq_k, \vq_k, \t \mQ_k$ are defined in \eqref{eq:def-k}, and $ \mD_k$ collects higher-order terms whose explicit
form is 
\begin{equation} \label{eq:delta}
\begin{aligned}
 \mD_k=  &\Big( \frac{\cos(\varphi) \theta_k^4}{24}  
 \frac{\vp_k\h\vw_k^\T}{\norm{\vp_k}\norm{\h\vw_k}}+
 \frac{\cos(\t \varphi) \theta_k^3}{6} \frac{\vr_k \h \vw_k^\T}{\norm{\vr_k} \norm{\h \vw_k}}+
 \\
& \tfrac{\tau_k^2}{2n^2} \mX_k 
\mZ_k^{-1}\vq_k \vq_k^\T
\mZ_k^{-1} \vq_k^\T \mZ_k^{-1}\vq_k \Big) \charfn_{\mathcal{A}_k}.
\end{aligned}
\end{equation}
Here, $\varphi$ and $\t \varphi$ are some numbers between $0$ and $\theta_k$.
Then, we can prove that 
\begin{align}
&\EEk \abs{[\mD_k]_{i,\ell}} \leq \EEk \Big(\theta_k^4 + \abs{\theta_k}^3  
\nonumber
\\
&\;\quad+ \tfrac{\tau_k^2}{n^2}
\abs{\textstyle \sum_{\ell^\prime=1}^d [\mX_k]_{i,\ell} \mZ_k^{-1}\vq_k \vq_k^\T
\mZ_k^{-1} \vq_k^\T \mZ_k^{-1}\vq_k } 
\Big) \charfn_{\mathcal{A}_k}
\nonumber
\\
&\leq Cn^{-3/2}, \label{eq:bound-Dij}
\end{align}
where in reaching the last line, we used  \eqref{eq:theta-k}, Lemma \ref{lem:bound-vars},
and $\normF{\mZ_k^{-1}}\charfn_{\mathcal{A}_k} \leq (d/\epsilon)^{-\frac{1}{2}}$. 

Multiplying from the right by $\mU^\T$ on both sides of \eqref{eq:mc-simplified},
we get 
\begin{equation}\label{eq:update-Q}
\begin{aligned}
 \mQ_{k+1}-\mQ_{k}
=& \tfrac{ \tau_{k}}{n}
 \big[\t \vq_{k} \vq_{k}^\T - \big( \t \mQ_{k}
 + \tfrac{ \tau_{k}}{2} \t z_{k}\mQ_{k} \big)
  \\
& 
 \mZ_k^{-1} \vq_k \vq_k^\T 
 \big] 
 \mZ_k^{-1}\charfn_{\mathcal{A}_k}
 + \t \mR_k,
 \end{aligned}
\end{equation}
where $\t \mR_k = \mU^\T \mD_k$ is the remainder term  collecting all high-order contributions.
This is \eqref{eq:update-Q2} in the main text.

\subsection{The remainder terms and higher-order moments}
We prove the following two lemmas that will be used in the proof of  Theorems \ref{thm:ode-d1} and  \ref{thm:finite}  in Section \ref{sec:grouse-oja}.

\begin{lemma} \label{lem:update-Q}
For all $k\leq nT$, the higher-order term in the update rule \eqref{eq:update-Q} of $\mQ_k$
is bounded by
\begin{equation} \label{eq:ineq-tR}
\EE \norm{\t \mR_k} \leq C(T)n^{-3/2}.
\end{equation}
\end{lemma}
\begin{IEEEproof}
From \eqref{eq:delta}, the explicit expression of the remainder term of $\t \mR_k=\mU^\T \mD_k$ can be written as
\begin{align*}
\t \mR_{k}=  &\Big( \frac{\cos(\varphi) \theta_k^4}{24}  
 \frac{\mU^\T \vp_k\h\vw_k^\T}{\norm{ \vp_k}\norm{\h\vw_k}}+
 \frac{\cos(\t \varphi) \theta_k^3}{6} \frac{\mU^\T \vr_k \h \vw_k^\T}{\norm{\vr_k} \norm{\h \vw_k}}+
 \\
& \tfrac{\tau_k^2}{2n^2} \mQ_k \mZ_k^{-1}\vq_k \vq_k^\T
 \mZ_k^{-1} \vq_k^\T \mZ_k^{-1}\vq_k \Big) \charfn_{\mathcal{A}_k}.
\end{align*}

Then, 
based on Lemma \ref{lem:bound-vars} and the facts that $\normF{\mU}=d$, $\norm{\mQ_k}\leq d^2$ and 
 $\normF{\mZ_k^{-1}}\charfn_{\mathcal{A}_k} \leq (d/\epsilon)^{-\frac{1}{2}}$,
 it is straightforward to prove \eqref{eq:ineq-tR}.
\end{IEEEproof}
\begin{lemma}
\label{lem:bound-s} Let  $\mX_k$ be the iterand of the GROUSE algorithm, whose  update rule  is given in 
\eqref{eq:mc-1d}. Then, for
all $k \leq nT$, 
\begin{equation}
\textstyle \sum_{i=1}^n \textstyle \sum_{\ell=1}^d \EE [\mX_{k}]_{i,\ell}^4 \leq C(T)/n, \label{eq:bound-s-grouse}
\end{equation}
where $C(T)$ is a constant can depend on $T$ but not on $n$.
\end{lemma}

\begin{IEEEproof} 
Let $\vb_{k,i}$ be the $i$th row of the matrix $\mX_k$.
It is sufficient to prove that
\begin{equation} \label{eq:vb4}
\EE\norm{\vb_{k,i}}^4 \leq C(T)/n^2
\end{equation}   holds for all  $i=1,2,\ldots,n$.

We first  prove an iterative inequality between $\EE \norm{\vb_{k+1,i}}^4$ and $\EE\norm{\vb_{k,i}}^4$. Knowing that 
$\abs{[\mX_k]_{i,\ell}}\leq 1$, $\abs{[\mU]_{i,\ell}}\leq 1$, and  Lemma  \ref{lem:bound-vars}, 
  from \eqref{eq:mc-simplified}, we get  
\begin{equation}
\EE \norm{  \vb_{k+1,i} - \vb_{k,i} }^{4} \leq Cn^{-4}. \label{eq:ineq-2}
\end{equation}
In addition,  from Lemma \ref{lem:bound-vars} and \eqref{eq:mc-simplified}, we can also show that
\begin{align} 
&\norm{ \EE_{k}(  \vb_{k+1,i} - \vb_{k,i} )} \nonumber \\
&\leq \tfrac{\tau_k}{n} 
\norm{\EEk v_{k,i} [\vs_{k}]_i \vq_k^\T]} 
+  \tfrac{C}{n} \big ( \norm{ \vb_{k,i} } + n^{-\frac{1}{2}} \big) \nonumber
\\
& =\tfrac{\tau_k}{n} \norm{   \EEk(v_{k,i} [\vs_{k}]_i \vs_k^\T \mOm_k) \mX_k }
+  \tfrac{C}{n} \big ( \norm{ \vb_{k,i} }+ n^{-\frac{1}{2}} \big) \nonumber
\\
&\leq \tfrac{C}{n}  \big(  \norm{ \vb_{k,i} } + \norm{ \t \vu_{i} }+ n^{-\frac{1}{2}} \big), \label{eq:bound-EE1}
\end{align}
where in reaching the last line, we used \eqref{eq:Eyy}, and $\t \vu_i$ is the 
$i$th row of the matrix $\mU$.
Then, we have
\begin{align} \nonumber 
 \EE & \norm{\vb_{k+1,i}}^4  \leq  \EE \big( \norm{\vb_{k,i}} + \norm{\vb_{k+1,i} - \vb_{k,i}} \big)^4
 \\ \nonumber
=   &\EE \Big[ \norm{ \vb_{k,i} }^4 
   + 4 \norm{ \vb_{k,i}}^3 \norm{\vb_{k+1,i} - \vb_{k,i}}
   \\ \nonumber
   &+ 6 \norm{ \vb_{k,i}}^2 \norm{\vb_{k+1,i} - \vb_{k,i}}^2
   + 4 \norm{ \vb_{k,i}} \norm{\vb_{k+1,i} - \vb_{k,i}}^3
   \\ \nonumber
   &+  \norm{\vb_{k+1,i} - \vb_{k,i}}^4 \Big]
   \\ \nonumber
   \leq &\EE \Big[ \norm{ \vb_{k,i} }^4 
   + 4 \norm{ \vb_{k,i}}^3 \EEk \norm{\vb_{k+1,i} - \vb_{k,i}}
   \\ \nonumber
   &+ \tfrac{C}{n}\norm{ \vb_{k,i} }^4  + Cn\norm{\vb_{k+1,i} - \vb_{k,i}}^4
\\
\leq & \left(1+\tfrac{C}{n}\right)\EE\norm{ \vb_{k,i} }^4+\tfrac{C}{n^3}. 
 \label{eq:vb-itr}
\end{align}
In reaching the last  line, we used  \eqref{eq:ineq-2}, \eqref{eq:bound-EE1},
Young's  inequality and a fact that
$
\frac{\abs{x}^3}{ \sqrt{n}} = n^{-2} (\, \abs{x}{ \sqrt{n}})^3 \leq C  n^{-2} 
[ ( {x}{ \sqrt{n}})^4 + 1] = C( x^4 + n^{-2}),
$ with some constant $C$.

Finally, using Lemma \ref{lem:itr} we can prove \eqref{eq:vb4} from   \eqref{eq:vb-itr}, and therefore
\eqref{eq:bound-s-grouse} holds.
\end{IEEEproof}

\section{Dynamics of Oja's method}\label{sec:oja}
In this section, we first provide a formal derivation of the ODE for Oja's method and then prove two lammas that are  counterparts  of Lemmas  \ref{lem:update-Q}  and  \ref{lem:bound-s} for Oja's method. The proofs follows the same procedure as we did in the previous section except that  we now consider $\mX_k$ as the iterand of the Algorithm \ref{alg:Oja}.

In Algorithm \ref{alg:Oja}, Line 7  involves an inverse of the matrix
$\t \mX_k^\T \t \mX_k$. 
In order to ensure this inverse is well-defined, we add 
a check before this line. Specifically,  
if
$\lambda_{\min}(\t \mX_k^\T \t \mX_k)\leq \e^\prime$ , then the update of this step is skipped by assigning 
$\mX_{k+1}=\mX_k$. We set $\e^\prime$   a small number strictly less than 1 ({\em e.g.,} $\e^\prime=0.1$). This check is used in both theoretical analysis and practical implementation in order to avoid the error of divided-by-zero. We note that the probability that this check is violated tends to zero as $n\to\infty$, since $\t \mX_k^\T \t \mX_k =\mI_d+\mathcal{O}(\frac{1}{n}).$

\subsection{Formal derivation of the ODE for Oja's method}
The update rule of Algorithm  \ref{alg:Oja} is formulated as
\begin{equation} \label{eq:oja-X}
\mX_{k+1} = \t\mX_k
\big( \t \mX_k^\T \t\mX_k \big) ^{-\frac{1}{2}} 
\charfn_{\mathcal{A}_k}\charfn_{\mathcal{A^\prime}_k}
+ \mX_k (1-\charfn_{\mathcal{A}_k}\charfn_{\mathcal{A^\prime}_k}),
\end{equation}
where $\t\mX_k=\mX_k + \tfrac{\tau}{n}\mB_k$. Here,  $\charfn_{\mathcal{A}_k}$ is an indicator function which is 1 if $\lambda_{\min}(\mX_k^\T \mOm_k \mX_k)> \epsilon$, and 0 otherwise. Furthermore,  $\charfn_{\mathcal{A^\prime}_k}$  is another indicator function which is 1 if $\lambda_{\min}(\t \mX_k^\T \t \mX_k)> \e^\prime$, and 0 otherwise.
The matrix $\mB_k$ is defined as 
\begin{equation}
\mB_k \bydef \h \vy_k \h \vw_k ^\T
= \mOm_k s_k \vq_k^\T \mZ_k^{-1} + (\mI_n -\mOm_k) \mX_k \mZ_k^{-1} \vq_k \vq_k^\T \mZ_k^{-1},
\end{equation}
where $\vq_k, \mZ_k$ are defined in \eqref{eq:def-k}. Then, we know that
\begin{align}
\mX_k^\T \mB_k &=\mZ_k^{-1} \vq_k \vq_k^\T \mZ_k^{-1} \label{eq:XB} \\
\mU^\T \mB_k &= \t \vq_k \vq_k^\T \mZ_k^{-1} + (\mQ_k - \t \mQ_k)\mZ_k^{-1} \vq_k \vq_k^\T \mZ_k^{-1} \label{eq:UB}\\
\mB_k^\T \mB_k &= \big[ n\t z_k + \mZ_k^{-1} \vq_k \vq_k^{\T} (\mZ_k^{-1} - \mI_d) \big]
\mZ_k^{-1} \vq_k \vq_k^{\T} \mZ_k^{-1}, \label{eq:BB}
\end{align}
where $\t \vq_k$ and $\t z_k$ are defined in  \eqref{eq:def-k}.

Note that when $\charfn_{\mathcal{A}_k}\charfn_{\mathcal{A^\prime}_k}=1$, the matrix $\t \mX_k^\T \t\mX_k$ is  positive-definite since
\begin{equation*}
\t \mX_k^\T \t\mX_k 
= \mI_d + 2\tfrac{\tau}{n} \mX_k^\T \mB_k + \tfrac{\tau^2}{n^2}\mB_k^\T \mB_k.
\end{equation*}
Then, we  use Taylor's 
expansion\footnote{Taylor's expansion of the positive-definite matrix $\big(\t \mX_k^\T \t\mX_k )^{-\frac{1}{2}}$
is well-defined: Let
$\big(\t \mX_k^\T \t\mX_k )^{-\frac{1}{2}}=\mV \opdiag\big((1+\mu_i)^{-\frac{1}{2}},\ell=1,2,\ldots,d  \big)\mV^\T$ be the eigenvalue decomposition of $\big(\t \mX_k^\T \t\mX_k )^{-\frac{1}{2}}$. Taylor's expansion is done by expanding its eigenvalues as $(1+\mu_i)^{-\frac{1}{2}}= 1 - \tfrac{1}{2}\mu_i + \ldots$.}
 to expand $\big(\t \mX_k^\T \t\mX_k )^{-\frac{1}{2}}$ up to the first order term 
\begin{equation} \label{eq:XX-oja}
\big(\t \mX_k^\T \t\mX_k \big)^{-\frac{1}{2}}
= \mI_d - \tfrac{\tau}{n} \mX_k^\T \mB_k - \tfrac{\tau^2}{2n^2}\mB_k^\T \mB_k + \mD_k.
\end{equation}
Here, $\mD_k$ represents high-order terms, which is bounded by
\begin{align} \label{eq:Oja-bound-D}
&\EEk \abs{[\mD_k]_{\ell,\ell^\prime}}\charfn_{\mathcal{A }_k} \charfn_{\mathcal{A^\prime}_k} 
\nonumber
\\
&\leq C \EEk \abs{
\big[\big(\tfrac{\tau}{n} \mX_k^\T \mB_k - \tfrac{\tau^2}{2n^2}\mB_k^\T \mB_k\big)^2 \big]_{\ell,\ell^\prime}} \charfn_{\mathcal{A }_k} \charfn_{\mathcal{A^\prime}_k} 
\nonumber
\\
&\leq C n^{-3/2},
\end{align}
where in reaching the last line, we used Lemma \ref{lem:bound-vars}.

Combining \eqref{eq:oja-X} with \eqref{eq:XB}, \eqref{eq:BB} and \eqref{eq:XX-oja}, we have
\begin{equation} \label{eq:Oja-update-X}
\begin{aligned}
\mX_{k+1} = &\mX_k - \tfrac{\tau}{n} 
\big[ \mOm_k \vs_k - \big(  \mOm_k + \tfrac{\tau}{2} \t z_k \mI_n\big)\mX_k \mZ_k^{-1} \vq_k \big]
\\
&  \vq_k^\T \mZ_k^{-1} \charfn_{\mathcal{A }_k} \charfn_{\mathcal{A^\prime}_k} 
+ \mS_k \charfn_{\mathcal{A }_k} \charfn_{\mathcal{A^\prime}_k},
\end{aligned}
\end{equation}
where $\mS_k$ collects all the high-order terms defined as
\begin{equation} \label{eq:Oja-E}
\begin{aligned}
\mS_k \bydef& \tfrac{\tau^2}{2 n^2} \mX_k \vq_k \vq_k^\T (\mZ_k^{-1} - \mI_d) \mZ_k^{-1} \vq_k \vq_k^\T \mZ_k^{-1} 
\\
&- \tfrac{\tau^2}{2n^2} \mB_k \big[ \mB^T_k \mX_k + \mX^\T \mB_k + \tfrac{\tau}{n} \mB^\T_k\mB_k \big] 
\\
&+ (\mX_k + \mB_k)\mD_k. 
\end{aligned}
\end{equation}

Multiplying $\mU^\T$ from the left on both sides of \eqref{eq:Oja-update-X}, we get
\begin{equation}
\begin{aligned}
\mQ_{k+1} - \mQ_k 
= &\tfrac{\tau}{n} 
\big[
\t \vq_k \vq_k^\T - \big(  \t \mQ_k + \tfrac{\tau}{n} \t z_k \mQ_k  \big) \mZ_k^{-1} \vq_k \vq_k^{\T}
\big]
\\
& \mZ_k^{-1} \charfn_{\mathcal{A }_k} \charfn_{\mathcal{A^\prime}_k} 
+ \t \mR_k,
\end{aligned}\label{eq:oja-Q}
\end{equation}
where 
\begin{equation} \label{eq:oja-R}
\t \mR_k \bydef\mU^\T \mE_k \charfn_{\mathcal{A }_k} \charfn_{\mathcal{A^\prime}_k}.
\end{equation}
Now, we get the update equation of $\mQ_k$, which is the same as \eqref{eq:update-Q} 
except that the definition of the high-order term $\t \mR_k$ is different. Thus, we know the dynamics of $\mQ_k$ of Oja's method and GROUSE are asymptotically the same. They only differ in the high-order term, which is vanishing when the ambient dimension $n\to\infty$.

\subsection{The remainder terms and higher-order moments}
Next, we  introduce the counterpart  of Lemmas \ref{lem:update-Q} for Oja's method.
\begin{lemma} \label{lem:update-Q-oja}
For $k\leq \lfloor nT \rfloor$,
the increment, $\mQ_{k+1} - \mQ_k$, is \eqref{eq:oja-Q} with the high-order term $\t \mR_k$
bounded by
\begin{equation} \label{eq:oja-tR-bound}
\EE \norm{\t \mR_k} \leq C(T)n^{-3/2}.
\end{equation}
\end{lemma}

\begin{IEEEproof}
From \eqref{eq:Oja-E} and \eqref{eq:oja-R}, we have
\begin{align*}
\t \mR_k=& \Big( \tfrac{\tau^2}{2 n^2} \mQ_k \vq_k \vq_k^\T (\mZ_k^{-1} - \mI_d) \mZ_k^{-1} \vq_k \vq_k^\T \mZ_k^{-1} 
\\
&- \tfrac{\tau^2}{2n^2} \mU^\T\mB_k \big[ \mB^T_k \mX_k + \mX^\T \mB_k + \tfrac{\tau}{n} \mB^\T_k\mB_k \big] 
\\
&+ (\mI_d + \mU^\T\mB_k)\mD_k \Big)
\charfn_{\mathcal{A }_k} \charfn_{\mathcal{A^\prime}_k}. 
\end{align*} 
Substituting \eqref{eq:XB}--\eqref{eq:BB} into the above equation, and with 
Lemma \ref{lem:bound-vars}, we can prove \eqref{eq:oja-tR-bound}.
 \end{IEEEproof}

The following lemma for Oja's method is the counterpart  of Lemmas \ref{lem:bound-s} for GROUSE.
\begin{lemma} \label{lem:bound-s-oja}
Let $\mX_k$ being the iterand of the  Oja's algorithm, whose update rule is given in
\eqref{eq:oja-X}. For
all $k \leq nT$, 
\begin{equation}
 \sum_{i=1}^n \sum_{\ell=1}^d \EE [\mX_{k}]_{i,\ell}^4 \leq C(T)/n, \label{eq:bound-s}
\end{equation}
where $C(T)$ is a constant can depend on $T$ but not on $n$.
\end{lemma}
 \begin{IEEEproof}
We prove this lemma in  the same way as that of Lemma \ref{lem:bound-s}.
In particular, let $\vb_{k,i}$ be the $i$th row of the matrix $\mX_k$. We can show that
 \begin{align*}
 \EEk \norm{\vb_{k+1,i} - \vb_{k,i}}^4 &\leq C n^{-4},
 \\
  \norm{\EEk\vb_{k+1,i} - \vb_{k,i} }
&\leq \tfrac{C}{n}  \big(  \norm{ \vb_{k,i} } + \norm{ \t \vu_{i} }+ n^{-\frac{1}{2}} \big).
 \end{align*}
 Then, we bound $\EE \norm{\vb_{k,i}} ^4$ iteratively  as we did in \eqref{eq:vb-itr}.  Since the proof is the same as what we did in proving Lemma \ref{lem:bound-s}, we omit the details here.
 \end{IEEEproof}

\section{Proof of Theorems \ref{thm:ode-d1} and  \ref{thm:finite} for GROUSE and Oja's Method}
\label{sec:grouse-oja}

Next, we prove Theorems \ref{thm:ode-d1} and  \ref{thm:finite} for both GROUSE and Oja's method.
The proof is based on the general convergence results 
as stated in Lemmas \ref{thm:weak} and  \ref{thm:gen-finite}. 
Specifically, the main task in this section is to check that Conditions \ref{c:m}--\ref{c:init} presented in Section \ref{sec:gen-conv} hold for the stochastic process $\mQ_k\bydef \mU^\T \mX_k$, where $\mX_k$ is the iterand in Algorithms \ref{alg:Oja} and \ref{alg:grouse}. The following proofs are valid for both GOURSE and Oja's methods. 

Note that $\mQ_k$ is a $d$-by-$d$ matrix. In order to use Lemmas \ref{thm:weak} and  \ref{thm:gen-finite}, we need to reshape the matrix $\mQ_k$ to an $n^2$-dimensional vector. This is equivalent to replace the vector norm in Conditions \ref{c:m}--\ref{c:init} by the Frobenius norm $\normF{\mQ_k}$. For any finite $d$, the Frobenius norm
is equivalent to the spectral norm $\norm{\mQ_k}$ in the sense that 
there are two constants $C_1$ and $C_2$ such that
\[
C_1\norm{\mQ_k} \leq \normF{\mQ_k} \leq C_2 \norm{\mQ_k}.
\]
Therefore, we directly work in the matrix representation using the spectral norm. 

We first decompose the increment of $\mQ_{k}$ into three parts
\begin{equation} \label{eq:decomp}
\mQ_{k+1} - \mQ_k= \tfrac{1}{n} \mL_k + \mM_k + \mR_k.
\end{equation}
Here, the first term 
\begin{equation}
\mL_{k} \bydef F(\mQ_k, \tau_k \mI_d) \label{eq:def-l}
\end{equation}
contains the leading-order contribution to $\mQ_{k}-\mQ_{0}$, in 
which $F(\cdot, \cdot)$ is defined in \eqref{eq:def-FQG}.
The second term is a martingale term defined as
\begin{align}
\mM_{k}  &\bydef \mat{\Delta}_{k}-\EEk \mat{\Delta}_{k} \label{eq:def-m} \\
\mat{\Delta}_{k} &\bydef \tfrac{ \tau_{k}}{n \a} 
\left[ \t \vq_{k} \vq_{k}- \left(1+ \tfrac{ \tau_{k}}{2}\s^2 \right)
\mQ_{k}\vq_{k} \vq_k^\T \right], \label{eq:def-Delta}
\end{align}
where $\t \vq_k$, $\vq_k$ are defined in \eqref{eq:def-k}.
The last term   in \eqref{eq:decomp} is
\begin{equation} \label{eq:def-r2}
\mR_{k}  \bydef \mQ_{k+1}-\mQ_{k}- \mat{\Delta}_{k}
+\EEk \mat{\Delta}_{k}-\tfrac{1}{n}\mL_{k},
\end{equation}
which collects all higher-order 
contributions.

Since $\norm{\mQ_k} \leq 1$, Conditions \ref{c:l}, \ref{c:bound} and \ref{c:init} trivially hold.

Furthermore, Condition \ref{c:m} also holds, because from \eqref{eq:def-m} and \eqref{eq:def-Delta}, we get 
\begin{align*}
&\EEk [\mM_{k}]_{\ell,\ell^\prime}^{2}  \leq\EEk [\mat{\Delta_{k} }]_{\ell,\ell^\prime}^{2}\\
 & \leq2\tfrac{\tau_{k}^{2}}{n^{2}\a^{2}}
 \left[\EEk [\t \vq_{k}]_{\ell}^{2} [\vq_{k}]_{\ell^\prime}^{2}
 + d \left(
 	1+\tfrac{\tau_{k}}{2} \s^2\right)^2 \EEk \norm{\vq_k}^2 [\vq_{k}]_{\ell^\prime}^{2}
\right]\\
 & \leq\tfrac{C}{n^{2}},
\end{align*}
where the last line is due to Young's inequality, (\ref{eq:bound-q})
and (\ref{eq:bound-p}).

Finally, Condition \ref{c:r} is ensured by the following lemma.

\begin{lemma} \label{lem:bound-r}
For all $k\leq nT$, we have 
\begin{equation*}
\EE \norm{\mR_k} \leq C(T) n^{-3/2}.
\end{equation*}

\end{lemma}
\begin{IEEEproof}
From \eqref{eq:def-r2}, we have 
\begin{equation} \label{eq:ineq-r}
 \norm{ \mR_k } \leq  \norm{ \EEk \mat{\Delta}_{k}- \tfrac{1}{n}\mL_{k}}
+  \norm{\mQ_{k+1}-\mQ_{k}- \mat{\Delta}_{k}}.
\end{equation}
Therefore, the proof is divided into two steps, which bound  the expectations of the two terms on the right-hand side separately.

The first term can be bounded straightforwardly.
From \eqref{eq:conv-q2}, \eqref{eq:conv-pq}, \eqref{eq:def-Delta}, 
and Lemma \ref{lem:bound-s} for GROUSE (Lemma \ref{lem:bound-s-oja} for Oja's method),
we  immediately get  
\begin{equation} \label{eq:Dlk}
\EE \norm{ \EEk \mat{\Delta}_{k}- \tfrac{1}{n}\mL_{k}}
\leq C(T)n^{-2}.
\end{equation}

Next, we are going to bound the expectation of the second term on the right hand side of \eqref{eq:ineq-r}.
From  Lemma \ref{lem:update-Q} for GROUSE (and Lemma \ref{lem:update-Q-oja} for Oja's method),  we have
\[
\begin{aligned}
&\mQ_{k+1}-\mQ_{k}- \mat{\Delta}_{k}\\
&= \tfrac{\tau_k}{n} 
\big[
\t \vq_k \vq_k^\T - (\t \mQ_k - \tfrac{\tau_k}{2} \t z_k \mQ_k)\mZ_k^{-1} \vq_k \vq_k^\T
\big]\mZ_k^{-1}\charfn_{\mathcal{A}_k} \\
&\quad + \t \mR_k - \mat{\Delta}_k \charfn_{\mathcal{A}_k}
-  \mat{\Delta}_k ( 1- \charfn_{\mathcal{A}_k} ).
\end{aligned}
\]
Its expectation can be bounded by   
\begin{equation}
\begin{aligned} & \EE 
\norm{\mQ_{k+1}-\mQ_{k}- \mat{\Delta}_{k}}
 \\& \leq \EE \norm{ \t \mR_{k}}+ \EE \norm{ \mat{\Delta}_{k}} ( 1- \charfn_{\mathcal{A}_k} )
\\
&\quad +  \tfrac{ \tau_{k}}{n}   \EE 
\Big[
	\norm{\t \vq_k \vq_k^\T (\mZ_k^{-1} - \a^{-1}\mI_d)}  \\
	&\quad+ 
	\norm{\t \mQ_k \mZ_k^{-1} \vq_k \vq_k^\T \mZ_k^{-1} - \a^{-1} \mQ_k \vq_k \vq_k^\T } \\
	&\quad+ 
	\tfrac{\tau_k}{2} \norm{\t z_k \mQ_k \mZ_k^{-1} \vq_k \vq_k^\T \mZ_k^{-1} - \s^2 \a^{-1} \mQ_k \vq_k \vq_k^\T}
\Big] \charfn_{\mathcal{A}_k}.
\end{aligned}
\label{eq:bound-Q-1}
\end{equation}
Here, we used \eqref{eq:def-Delta}.

The first term on the right hand side of \eqref{eq:bound-Q-1} is bounded by 
\begin{equation}
\EE\norm{ \t \mR_{k}}\leq C(T) n^{-3/2}
, \label{eq:bdR}
\end{equation}
 due to Lemma \ref{lem:update-Q} for GROUSE (and Lemma \ref{lem:update-Q-oja} for Oja's method).

The second term  on the right hand side of \eqref{eq:bound-Q-1} is bounded by 
\begin{align} \nonumber
&\EE \norm{ \mat{\Delta}_{k}} \cdot \charfn_{\mathcal{A}_k}
=\EE \norm{ \mat{\Delta}_{k}} \cdot \charfn[\lambda_{\min}(\mZ_{k}) \leq \epsilon] \\
& \leq 
\left[ \EE\norm{ \mat{\Delta}_{k}}^{2} \right]^{ \frac{1}{2}} 
\left[ \mathbb{P}[\lambda_{\min}(\mZ_{k}) \leq \epsilon] \right]^{ \frac{1}{2}} 
\nonumber \\
 & \leq \tfrac{C}{n} 
 \big( \mathbb{P} \big[ \lambda_{\min}(\mZ_k - \a \mI_d)^{2} \geq \left( \epsilon- \a \right)^{2} \big] \big)^{ \frac{1}{2}} \nonumber \\
  & \leq \tfrac{C}{n} 
 \big( \mathbb{P} \big[ \norm{\mZ_k - \a \mI_d}^{2} \geq \left( \epsilon- \a \right)^{2} \big] \big)^{ \frac{1}{2}} \nonumber \\
 & \leq \tfrac{C}{n \cdot  \abs{ \epsilon- \a}} 
 \cdot  \big[ \EE \norm{\mZ_{k}- \a\mI}^{2} \big]^{ \frac{1}{2}} \nonumber \\
 & \leq Cn^{- \frac{3}{2}}, \label{eq:bound-d0}
\end{align}
where in reaching the last line we used (\ref{eq:conv-z}), and  Lemma \ref{lem:bound-s} for GROUSE (and Lemma \ref{lem:bound-s-oja} for Oja's method).

The remaining task is to bound the terms in the last three lines
of \eqref{eq:bound-Q-1}. Specifically, the third last line of  \eqref{eq:bound-Q-1}
is bounded by
\begin{align} \nonumber
&\EE \norm{\t \vq_k \vq_k^\T (\mZ_k^{-1} - \a^{-1}\mI_d)} 
=\a^{-1} \EE \norm{\t \vq_k \vq_k^\T \mZ_k^{-1}( \a \mI- \mZ_k)}
\\ \nonumber
&\leq  \a^{-1} \Big(\EE
\norm{\vq_k \vq_k^\T \mZ_k^{-1} }^2
 \Big)^{\frac{1}{2}}
 \Big(\EE
\norm{ \a - \mZ_k }^2
 \Big)^{\frac{1}{2}}
 \\ \nonumber
 & \leq Cn^{- \frac{1}{2}}s_{k}^{ \frac{1}{2}} \\ \nonumber
 & \leq Cn^{- \frac{1}{2}} \left(1+s_{k} \right)
 \\
 &\leq C(T)n^{- \frac{1}{2}},
\end{align}
where in reaching the third line, we used $\norm{\mZ_k^{-1}}\leq C/\sqrt{\epsilon}$, and Lemma \ref{lem:bound-vars},
and the last line is due to Lemma \ref{lem:bound-s} for GROUSE (Lemma \ref{lem:bound-s-oja} for Oja's method).
Similarly, the second last line of  \eqref{eq:bound-Q-1}
is bounded by
\begin{align}
\nonumber
&\EE\norm{\t \mQ_k \mZ_k^{-1} \vq_k \vq_k^\T \mZ_k^{-1} - \a^{-1} \mQ_k \vq_k \vq_k^\T }\charfn_{\mathcal{A}_k}
\\ 
\nonumber
&\leq\EE
\norm{(\t \mQ_k - \a \mQ_k ) \mZ_k^{-1} \vq_k \vq_k^\T \mZ_k^{-1} } \charfn_{\mathcal{A}_k}
\\ \nonumber & \quad
+\EE\norm{ \mQ_k (\a \mI_d -\mZ_k) \mZ_k^{-1}\vq_k \vq_k^\T \mZ_k^{-1} }
\charfn_{\mathcal{A}_k}
\\ 
\nonumber & \quad
+\a^{-1}\EE\norm{\mQ_k \vq_k \vq_k^\T \mZ_k^{-1}(\a  \mI_d -  \mZ_k)} \charfn_{\mathcal{A}_k}
\\ 
\nonumber
&\leq \Big( \EE
\norm{(\t \mQ_k - \a \mQ_k )  }^2 \Big)^\frac{1}{2}
\Big( \EE
\mZ_k^{-1} \vq_k \vq_k^\T \mZ_k^{-1}
\charfn_{\mathcal{A}_k}
\Big)^\frac{1}{2} 
\\ 
\nonumber & 
+\Big(
\EE\norm{\a \mI_d -  \mZ_k}^2
\Big)^\frac{1}{2}
\Big(
\EE \norm{ \mQ_k}
\cdot
\norm{  \mZ_k^{-1}\vq_k \vq_k^\T \mZ_k}^2
\charfn_{\mathcal{A}_k}
\Big)^\frac{1}{2} 
\\ \nonumber & 
+
\a^{-1} 
\Big(
 \EE\norm{\a\mI - \mZ_k}
\Big)^\frac{1}{2}
\Big(
\EE\norm{\mQ_k \vq_k \vq_k^\T \mZ_k^{-1}}^2
\charfn_{\mathcal{A}_k}
\Big)^\frac{1}{2}
\\ 
& \leq C(T)n^{-\frac{1}{2}}, \label{eq:bound-d2}
\end{align}
where in reaching the last inequality, we used $\norm{\mZ_k^{-1}}\leq C/\sqrt{\epsilon}$, and Lemma \ref{lem:bound-vars}
and Lemma \ref{lem:bound-s} for GROUSE (Lemma \ref{lem:bound-s-oja} for Oja's method).
The term in the last line of \eqref{eq:bound-Q-1} can be bounded in the same way, which yields
\begin{equation}
\EE 	\tfrac{\tau_k}{2} \norm{\t z_k \mQ_k \mZ_k^{-1} \vq_k \vq_k^\T \mZ_k^{-1} - \s^2\a^{-1} \mQ_k \vq_k \vq_k^\T}
\leq C(T)n^{-\frac{1}{2}}
\label{eq:bound-d3}
\end{equation}

Finally, substituting  \eqref{eq:bdR}--\eqref{eq:bound-d3}  into \eqref{eq:bound-Q-1}, we reach 
\begin{equation*}
\EE \norm{\mQ_{k+1}-\mQ_{k}- \mat{\Delta}_{k}} \leq C(T) n^{-3/2}.
\end{equation*}
Combining the above inequality with \eqref{eq:Dlk} and  \eqref{eq:ineq-r}, we finish the proof of Lemma~\ref{lem:bound-r}. 
\end{IEEEproof}

\section{Convergence of Simplified PETRELS}
 \label{sec:petrels}
 We first provide the formal derivation of the limiting ODE for the simplified PETRELS. The rigorous proof will be presented in subsequent subsections.
\subsection{Formal derivation of the limiting ODE}
Recall that  pseudo-code of this algorithm as presented in Algorithm~\ref{alg:sim-petrels}.
Note that $(\mX_{ k}, \mR_{k})_{ k=0,1,2,\ldots}$ forms a Markov chain on $\R^{d\times n + d^2}$. It is driven by the initial states $(\mX_{0}, \mA_{0})$, and the randomness $\mOm_k$, $\vc_k$ and $\va_k$. Those random variables  generate the observation sample $\vs_k=\mOm_k(\mU \vc_k + \va_k)$ according to \eqref{eq:gen} and \eqref{eq:subsample}. The update rule, from Algorithm~\ref{alg:sim-petrels}, is
\begin{align*}
\mX_{k+1}&=\mX_k + \mOm_k( \vs_k - \mX_k \h \vw_k)\h \vw_k^\T \mR_k,\charfn_{\mathcal{A}_k}
\\
\mR_{k+1}&=(\gamma \mR_k^{-1} + \a \widehat\vw_k \widehat\vw^\T_k)^{-1}
\charfn_{\mathcal{A}_k} + \mR_k (1-\charfn_{\mathcal{A}_k}),
\end{align*}
where 
\begin{equation} \label{eq:pet-hw}
\h \vw_k=(\mX_k^\T \mOm_k \mX_k)^{-1}  \mX_k^\T \mOm_k \vs_k,
\end{equation}
and  $\charfn_{\mathcal{A}_k}$ is an indicator function which is 1 if $\lambda_{\min}(\mX_k^\T \mOm_k \mX_k)> \epsilon$, and 0 otherwise.

We denote by $\EEk$  the conditional expectation of the Markov chain  given the states $(\mX_{0}, \mA_{0})$, $(\mX_{1}, \mA_{1})$,\ldots,$(\mX_{k}, \mA_{k})$, and denote by $\EE$  the expectation of all the randomness.

It is convenient to change the variable $\mR_k$ to $\mA_k  \bydef \frac{1}{n}\mR_k^{-1}$  as defined in \eqref{eq:def-AKW}.  The update rule of the Markov chain can be written as 
\begin{align}
\mX_{k+1}&=\mX_k + \tfrac{1}{n}\mOm_k( \vs_k - \mX_k \h \vw_k)\h \vw_k^\T \mA_k^{-1}
\charfn_{\mathcal{A}_k}
\label{eq:pet-mc-1}
\\
\mA_{k+1}&=\mA_k   + \tfrac{1}{n}
\big( \a \h \vw_k \h \vw_k^\T - \mu \mA_k) \charfn_{\mathcal{A}_k}, \label{eq:pet-mc-2}
\end{align}
where $\mu$ is the rescaled discount parameter such that $\gamma=1-\frac{\mu}{n}$.


In what follows, we  derive the limiting ODE for $\mA(t)$, $\mK(t)$ and $\mW(t)$, whose definitions are in \eqref{eq:def-AKW-t}, when $n$ goes to infinity. Following the same paradigm of the formal derivation as stated in Section~\ref{sec:ode} for GROUSE, we only need to find the leading terms in 
$\frac{ \mA_{k+1} - \mA_k}{1/n}$, $\frac{ \mK_{k+1} - \mK_k}{1/n}$, and $\frac{ \mW_{k+1} - \mW_k}{1/n}$. 

It is convenient to introduce the variables 
$\mZ_k$, $\t z_k$, $\vq_k$ and $\t \vq_k$ that are defined in \eqref{eq:def-k}.
Then, from \eqref{eq:def-AKW} and  \eqref{eq:pet-mc-1}--\eqref{eq:pet-hw}, we have
\begin{align} \label{eq:LA}
\mA_{k+1} - \mA_k &= \tfrac{1}{n} 
\big(
\a \mZ_k^{-1} \vq_k \vq_k^\T \mZ_k^{-1} - \mu \mA_k 
\big)
\charfn_{\mathcal{A}_k}
\\ \label{eq:LK}
\mK_{k+1} - \mK_k &= \tfrac{1}{n}
\big(
\t \vq_k \vq_k^\T - \t \mQ_k \mZ_k^{-1} \vq_k \vq_k^\T \big) \mZ_k^{-1} \mA_k^{-1}  \charfn_{\mathcal{A}_k}
\\ \nonumber
\mW_{k+1} - \mW_{k} &= 
\tfrac{1}{n}\t z_k \mA_k^{-1} \mZ_k^{-1} \vq_k \vq_k^\T \mZ_k^{-1} \mA^{-1}
\charfn_{\mathcal{A}_k} \\
& \quad - \tfrac{1}{n^2} \mA_k^{-1} \mZ_k^{-1} \vq_k \vq_k^\T \mZ_k^{-1} 
\vq_k \vq_k^\T \mZ_k^{-1} \mA_k^{-1}\charfn_{\mathcal{A}_k}.
\label{eq:LW}
\end{align}

Analogous to Lemma~\ref{lem:conv-var},  when $n\to \infty$, we know that
\begin{equation} \label{eq:conv-WK}
\begin{aligned}
\mZ_k & \to \a \mW_k\\
\widetilde\mQ_k & \to \a \mK_k \\
\t z_k & \to \a \s^2 \\
\vq_k\vq_k^\T &\to \a \big( \s^2 \mI_d + \a \mK_k^\T \mat \Lambda^2 \mK_k \big) \\
\t \vq_k \vq_k^\T &\to \a \big( \s^2 \mI_d + \a \mat \Lambda^2 \big)\mK_k,
\end{aligned}
\end{equation}
and  $\charfn_{\mathcal{A}_k}$ is equal to 1 with high probability.
The rigorous justification of the above claim will be presented in Lemma~\ref{lem:pet-conv-vars} in the next subsection.
Then, substituting \eqref{eq:conv-WK} into \eqref{eq:LA}, \eqref{eq:LK} and \eqref{eq:LW}, we  can derive the 
 limiting ODE  \eqref{eq:ODE-pet} as shown in the main text.

%

\subsection{Bounds and Moment Estimates}

\begin{lemma} \label{lem:bound-W}
For all positive integer $k$, the symmetric matrices $\mA_k$ and $\mW_k$ as defined in \eqref{eq:def-AKW} satisfying  
\begin{align} \label{eq:bound-mA}
\lambda_{\min}(\mA_k)  &\geq e^{-\frac{\mu k}{n}} \lambda_{\min}(\mA_0).
\\
\lambda_{\max}(\mA_k) & \leq \tfrac{\a}{n} \sum_{\t k=0}^k
\norm{\mZ_{\t k}^{-1} \vq_{\t k} \charfn_{\mathcal{A}_{\t k}} } \label{eq:amax}
\\
\lambda_{\min}(\mW_{k}) &\geq \lambda_{\min}(\mW_{0})  \label{eq:lmin}
\\
\lambda_{\max}(\mW_{k}) &\geq \max_{0< \t k \leq k}\lambda_{\max}(\mW_{\t k}). \label{eq:lmax}
\end{align}
\end{lemma}
\begin{IEEEproof}
From \eqref{eq:pet-mc-2}, we know that
\[
\lambda_{\min}(\mA_k) \geq (1-\tfrac{\mu}{n}) \lambda_{\min}(\mA_{k-1}) 
\geq (1-\tfrac{\mu}{n})^k  \lambda_{\min}(\mA_{0}) .
\]
Then, using the inequality that 
$(1-\tfrac{1}{x})^{-x}>e$ for all $x>1$, we  can prove \eqref{eq:bound-mA}.

From \eqref{eq:LW}, we know that $\mW_{k+1}-\mW_{k}$ is a positive semi-definite matrix.
Thus, using Weyl's inequality, we have
\begin{align*}
\lambda_{\min}(\mW_{k+1}) \geq \lambda_{\min}(\mW_{k}) \\
\lambda_{\max}(\mW_{k+1}) \geq \lambda_{\max}(\mW_{k}),
\end{align*}
which imply \eqref{eq:lmin} and \eqref{eq:lmax}.
\end{IEEEproof}
This Lemma ensures that $\norm{\mA_k^{-1}}$ is bounded by a constant $C(T)$ for any $k\leq nT$, where $T$ is any finite positive number,
and  $\mW_k$ is always invertible. Thus the overlap matrix
$\mQ_k = \mK_k \mW_k^{-\frac{1}{2}}$ is well-defined.

Similar to Lemma~\ref{lem:bound-vars}, we have the following lemma for the simplified PETRELS.
\begin{lemma} \label{lem:pet-bound-vars}
The variables defined in \eqref{eq:def-k}  satisfy
\begin{align} \label{eq:WZW}
\norm{ \mW_k^{-\frac{1}{2}} \mZ_{k} \mW_k^{-\frac{1}{2}}} & \leq 1 \\
\norm{\t \mQ_k \mW_k^{-\frac{1}{2}} } & \leq 1 \\
\EEk \t  z_k^\ell & \leq C(\ell) \\
\EEk \norm{\mW_k^{-\frac{1}{2}}\vq_k}^\ell &\leq C(\ell)  \label{eq:bound-Wq}\\
\EEk \norm{\t \vq_k}^\ell &\leq C(\ell),  \label{eq:pet-ql}
\end{align}
in which $\mW_k$ is defined in \eqref{eq:def-AKW}. In addition, if we replace $\mW_k^{-\frac{1}{2}}$  by $\mZ_k^{-\frac{1}{2}}\charfn_{\mathcal{A}_k}$, the above inequalities   still hold.
\end{lemma}
\begin{IEEEproof}
Note that here $\mX_k$ is not normalized. Instead, $\mX_k \mW_k^{-\frac{1}{2}}$ is normalized. 
We consider $\mX_k \mW_k^{-\frac{1}{2}}$ as a new, normalized $\mX_k$. Then, using Lemma~\ref{lem:bound-vars} we can  prove this lemma.

In addition, $\mOm_k \mX_k \mZ_k^{-\frac{1}{2}}$ is normalized. Thus, we can 
eplace $\mW_k^{-\frac{1}{2}}$  by $\mZ_k^{-\frac{1}{2}}\charfn_{\mathcal{A}_k}$, the  inequalities \eqref{eq:WZW}--\eqref{eq:pet-ql}   hold.
\end{IEEEproof}

\begin{lemma} \label{lem:bound-WA-2}
For $k\leq nT$ and any positive integer $\ell$, we have
\begin{align}
\EE \norm{\mA_k}^2 &\leq C(T),  \label{eq:bound-A-2} \\
\EE \norm{\mW_{k}}^2 &\leq C(T).  \label{eq:bound-W-2} 
\end{align}
\end{lemma}
\begin{IEEEproof}
From \eqref{eq:LA}, we have
\begin{align*}
&\EE \norm{\mA_{k+1}}^2 \leq 
\EE\big( \norm{\mA_k} 
+ \tfrac{a}{n} \norm{\mZ_k^{-1} \vq_k \vq_k^\T \mZ_k^{-1} } \big)^2 \\
& 
\leq \EE \norm{\mA_k}^2 + \tfrac{\a^2}{n^2} \EE  \norm{\mZ_k^{-1} \vq_k \vq_k^\T \mZ_k^{-1} }^2
\\
&\quad
+ 2\tfrac{\a}{n} \EE\norm{\mA_k}\cdot \norm{\mZ_k^{-1} \vq_k \vq_k^\T \mZ_k^{-1} }
\\
& 
\leq (1+\tfrac{\a}{2n}) \EE \norm{\mA_k}^2 
+ (\tfrac{\a^2}{n^2} +  \tfrac{a}{2n} )\EE \norm{\mZ_k^{-1} \vq_k \vq_k^\T \mZ_k^{-1}}^2
\\
& \leq (1+\frac{C}{n}) + \tfrac{C}{n},
\end{align*}
where in reaching the second last line we used Young's inequality, and the last line is due to Lemma~\ref{lem:pet-bound-vars}.
Then, using this iterative bound,  Lemma~\ref{lem:itr} implies  \eqref{eq:bound-A-2}.

Similarly, we can prove $\eqref{eq:bound-W-2}$. We omit the details here.
\end{IEEEproof}

Similar to Lemma~\ref{lem:conv-var}, we have the following lemma for the simplified PETRELS algorithm.
\begin{lemma} \label{lem:pet-conv-vars}
For all non-negative integer $k\leq nT$, the following inequalities hold.
Let $s_k\bydef\sum_{i=1}^n \sum_{j=1}^d [\mX_{k}]_{i,j}^4$.
Then,
\begin{align}
\EE \norm{ \EEk \mZ_k - \a \mW_k}^2
&\leq \tfrac{C(T)}{n}  \label{eq:XX}
\\
\EE \norm{\EEk \t \mQ_k - \a \mK}^2
&\leq \tfrac{C(T)}{n}
\\
\EE \norm{\EEk \t z_k - \a \s^2 }^2  
&\leq \tfrac{C(T)}{n}
\\ 
\EE \norm{ \EEk \vq_k \vq_k^\T 
- \a \big( \s^2 \mW_k + \a \mK_k^\T \mat \Lambda^2 \mK_k  \big)
} &
\leq \tfrac{C(T)}{n} 
\label{eq:XssX}
\\
\EE\norm{\EEk \t \vq_k \vq_k^\T -  \a \big( \s^2 \mI_d + \a \mat \Lambda^2 \big)\mK_k}
&\leq \tfrac{C(T)}{n} 
\end{align}
\end{lemma}

\begin{IEEEproof}
From Lemma~\ref{lem:conv-var}, we know that
\begin{align*}
\EE \norm{ \EEk \mZ_k - \a \mW_k} 
&\leq C s_k   
\\
\EE \norm{\EEk \t \mQ_k - \a \mK} 
&\leq C (s_k + \tfrac{1}{n}) 
\\
\EE \norm{\EEk \t z_k - \a \s^2 }^2  
&\leq C\tfrac{1}{n}
\\ 
\EE \norm{ \EEk \vq_k \vq_k^\T 
- \a \big( \s^2 \mW_k + \a \mK_k^\T \mat \Lambda^2 \mK_k  \big)
} 
&\leq C(s_k + \tfrac{1}{n}) 
\\
\EE\norm{\EEk \t \vq_k \vq_k^\T -  \a \big( \s^2 \mI_d + \a \mat \Lambda^2 \big)\mK_k} 
&\leq C(s_k + \tfrac{1}{n}) 
\end{align*}
The remaining task is to prove $\vs_k \leq \frac{C(T)}{n}$.

Let $\vb_{k,i}$ be the $i$th row of the matrix $\mX_k$. Since
$ \sum_{j=1}^{d} [\mX_k]_{i,j}^4 \leq \norm{\vb_{k,i}}$, it is sufficient to prove
\begin{equation} \label{eq:pet-b}
\EE \norm{\vb_{k,i}}^4 \leq C(T)/n^2.
\end{equation} 
In reaching this inequality, we first prove that
\begin{align} \label{eq:pet-Eb}
\norm{\EEk \vb_{k+1,i} - \vb_{k,i} } 
&\leq \tfrac{C}{n} \big( \norm{\vb_{k,i} } + \norm{\t \vu_i}  \big)
\\ \label{eq:pet-b4}
\EE \norm{\vb_{k+1,i} - \vb_{k,i} }^4 & \leq \tfrac{C}{n^4}.
\end{align}
Then,  with the same argument as shown in \eqref{eq:vb-itr}, we can get an 
iterative bound
\begin{equation}
 \EE  \norm{\vb_{k+1,i}}^4 \leq  \left(1+\tfrac{C}{n}\right)\EE\norm{ \vb_{k,i} }^4+\tfrac{C}{n^3}.
\end{equation}
Finally, we employ Lemma~\ref{lem:itr} to prove \eqref{eq:pet-b}.

In what follows, we establish \eqref{eq:pet-Eb} and \eqref{eq:pet-b4}.
From \eqref{eq:def-AKW}, \eqref{eq:pet-mc-1} and \eqref{eq:hw}, we have
\begin{equation} \label{eq:pet-update-b}
\vb_{k+1,i} = \vb_{k,i} + \tfrac{1}{n} v_{k,i} (s_{k,i} - \vb_{k,i} \mZ_k^{-1} \vq_k)
\vq_k^\T \mZ_k^{-1} \mA_k^{-1}   \charfn_{\mathcal{A}_k},
\end{equation}
where $v_{k,i}$ is the $i$the diagonal term of $\mOm_k$, and $s_{k,i}$ is the $i$th entry of $\vs_k$, and $\mZ_k$ and $\vq_k$ are defined in \eqref{eq:def-k}.
Then, we have
\begin{align*}
&\EEk \vb_{k+1,i} - \vb_{k,i} 
= \tfrac{1}{n} \EE_{\mOm_k} v_{k,i} \t \vu_i \mat \Lambda^2
\t \mQ_k \mZ_k^{-1} \mA_k^{-1} \charfn_{\mathcal{A}_k}
\\
&- \tfrac{\s^2}{n} \EE_{\mOm_k} v_{k,i} \vb_{k,i} 
\mZ_{k}^{-1} \mX_k^\T \mOm_k (\mat \Lambda^2 + \s^2 \mI_n) \mOm_k \mX_k
\mZ_{k}^{-1} \mA_k  \charfn_{\mathcal{A}_k},
\end{align*}
where $\t \vu_i$ is the $i$th row of $\mU$.
Since $ \norm{ \t \mQ_k \mZ_k^{-\frac{1}{2}} } \leq 1$, 
$\norm{\mZ_k^{-1}} \leq C$, $\norm{\mA_k^{-1}} \leq C$,
and $\norm{\mX_k \mZ_k^{-\frac{1}{2}}}\leq 1$, the above equation implies
\eqref{eq:pet-Eb}.

Next, we prove \eqref{eq:pet-b4}. From \eqref{eq:pet-update-b}, we have
\begin{align*}
&\EE \norm{\vb_{k+1,i} - \vb_{k,i} }^4 
\\
&\leq \tfrac{1}{n^4}
\EE \norm{ v_{k,i} (s_{k,i} - \vb_{k,i} \mZ_k^{-1} \vq_k)
\vq_k^\T \mZ_k^{-1} \mA_k^{-1}   \charfn_{\mathcal{A}_k}
}^4
\\
&\leq 
\tfrac{8}{n^4} \EE \norm{ 
v_{k,i} s_{k,i} \vq_k^\T \mZ_k^{-\frac{1}{2} }  \charfn_{\mathcal{A}_k} } 
\cdot
\norm{ \mZ_k^{-\frac{1}{2} }  \charfn_{\mathcal{A}_k} } \cdot \norm{\mA_k^{-1}} 
\\
&\quad
+\tfrac{8}{n^4} \EE \norm{ v_{k,i}  \vb_{k,i} \mZ_k^{-\frac{1}{2} }  \charfn_{\mathcal{A}_k} }  
\cdot \norm{\mZ_k^{-\frac{1}{2} }  \vq_k \vq_k^\T \mZ_k^{-\frac{1}{2} }  
\charfn_{\mathcal{A}_k} } 
\\
&\quad \quad\quad\quad
 \norm{ \mZ_k^{-\frac{1}{2} }  \charfn_{\mathcal{A}_k} } 
\cdot \norm{\mA_k^{-1} }
\\
& \leq \tfrac{C}{n^4}.
\end{align*}

\end{IEEEproof}

Now,  having completed all the preparations, we are ready to prove the convergence of 
$\mA_k, \mK_k$, and $\mW_k$.

\subsection{Convergence of the Simplified PETRELS}

In this section, we prove Theorem~\ref{thm:limit} and Theorem~\ref{thm:finite} by using Lemmas~\ref{thm:weak} and \ref{thm:gen-finite}. 

In order to use Lemmas~\ref{thm:weak} and \ref{thm:gen-finite}, 
we specialize  the general stochastic  process $\vq_k^{(n)}$ in Section~\ref{sec:limit} to be 
the tuple of $\mA_k$, $\mK_k$ and
$\mW_k$. More specifically, we  consider $\vq_k^{(n)}$ as a $3d^2$-dimensional vector that are the concatenation of the vectorized form of the three $d\times d$ matrices  $\mA_k$, $\mK_k$ and
$\mW_k$. Thus, we have 
\[
\norm{\vq_k^{(n)} }^2 = \normF{\mA_k}^2   + \normF{\mK_k}^2 + \normF{\mW_k}^2.
\]
In what follows, we check that Conditions~\ref{c:m}--\ref{c:init} are satisfied  for this specialized process on $\R^{3d^2}$.

We first show Condition~\ref{c:l} holds. 
From Lemma~\ref{lem:bound-W}, we know that $\norm{\mA_k^{-1}}$ and $\mW^{-1}$ is uniformly bounded, and $\norm{ \mK_k \mW^{-\frac{1}{2}}  }$ are also uniformly bounded. Furthermore, we can rewrite the functions $J_1$, $J_2$ and $J_3$ defined in \eqref{eq:def-pet-J} as
\begin{equation}
\begin{aligned}
 J_1 (\mA, \mK, \mW) &=   \mW^{-\frac{1}{2}}
\big( 
 \mW^{-\frac{1}{2}}\mK^\T \a\mat \Lambda^2 \mK \mW^{-\frac{1}{2}} 
\\
&\quad 
+ \s^2 \mI_d \big)
\mW^{-\frac{1}{2}}
-\mu\mA
\\
J_2(\mA, \mK, \mW) &= (\a \mat \Lambda^2 + \s^2 \mI_d)\mK \mW^{-\frac{1}{2}} 
 \mW^{-\frac{1}{2}} \mA^{-1}
\\
& \quad \;
- \mK  \mW^{-\frac{1}{2}}  \big(
  \mW^{-\frac{1}{2}} \mK^\T \a\mat \Lambda^2 \mK  \mW^{-\frac{1}{2}} 
\\
&\quad \;
+ \s^2 \mI_d
\big)
 \mW^{-\frac{1}{2}}
\mA^{-1}
\\
J_3 (\mA, \mK, \mW)&= \s^2 \mA^{-1} \mW^{-\frac{1}{2}}( 
\mW^{-\frac{1}{2}} \mK^\T \a \mat \Lambda^2 \mK \mW^{-\frac{1}{2}} 
\\
&\quad \;
+ \s^2 \mI_d) 
  \mW^{-\frac{1}{2}}
\mA^{-1}. 
\end{aligned}
\end{equation}
Then, it is clear that $J_1$, $J_2$ and $J_3$ are Lipschitz functions in the domain
where $\norm{\mA^{-1}}$, $\norm{\mW^{-1}}$ and $\norm{\mK \mW^{-\frac{1}{2}}}$ are uniformly bounded. 

We next show that Conditions~\ref{c:bound}  holds.
From \eqref{eq:lmax},  we have
\begin{align} \nonumber
& \mathbb{P}
  \Big(\max_{k=0,1,2,\nT} \norm{\mW_k } \geq b\Big)
   \leq 
   \mathbb{P}
  \Big( \norm{\mW_{\nT} } \geq b\Big)
  \\
  & \leq \frac{\EE \norm{ \mW_{\nT }} }{b}
 \leq \frac{C(T)}{b}, \label{eq:P-1}
\end{align}
where in reaching the last inequality, we used   \eqref{eq:bound-W-2}. Furthermore, 
from \eqref{eq:amax}, we have
\begin{align} \nonumber
& \mathbb{P}
\Big(\max_{k=0,1,2,\nT} \norm{\mA_k } \geq b\Big)
   \leq 
   \mathbb{P}
  \Big(\frac{\a}{n} \sum_{ k=0}^{\nT}
\EE \norm{\mZ_{ k}^{-1} \vq_{ k} \charfn_{\mathcal{A}_{ k}} }  \geq b\Big)
  \\
 &\leq \frac{\a}{nb} \sum_{ k=0}^{\nT}
\EE \norm{\mZ_{ k}^{-1} \vq_{ k} \charfn_{\mathcal{A}_{ k}} }  
\leq \frac{C(T)}{b}, \label{eq:P-2}
\end{align} 
 where in reaching the last inequality, we used Lemma~\ref{lem:pet-bound-vars}.
 In addition, we note that $\norm{\mK_k} \leq \norm{\mW_k^{\frac{1}{2}}}$. Thus,
 \begin{align} \label{eq:P-3}
& \mathbb{P}
  \Big(\max_{k=0,1,2,\nT} \norm{\mK_k } \geq b\Big)
   \leq 
   \mathbb{P}
  \Big( \norm{\mW^\frac{1}{2}_{\nT} } \geq b\Big)  \leq \frac{C(T)}{b}.
  \end{align} 
  Combining  \eqref{eq:P-1}, \eqref{eq:P-2}, and  \eqref{eq:P-3}, we proved that Conditions~\ref{c:bound}  holds.
  
Similarly, it is not difficult to check Conditions~\ref{c:init}  also holds from Lemmas~\ref{lem:bound-W}, \ref{lem:pet-bound-vars}, and \ref{lem:bound-WA-2}. We omit the details here.

In the remaining part of this subsection, we show that Conditions~\ref{c:m} and \ref{c:r} hold.
Following the recipe stated in Section~\ref{sec:limit}, we decompose each of $\mA_k, \mK_k$, and $\mW_k$ into three parts:
\begin{align*}
\mA_{k+1}-\mA_k &= \tfrac{1}{n} \mJ^{\mA}(\mA_k, \mK_k, \mW_k)+ 
\mM_k^{\mA} + \mR_k^{\mA} 
\\
\mK_{k+1}-\mK_k &= \tfrac{1}{n} \mJ^{\mK}(\mA_k, \mK_k, \mW_k)+ 
\mM_k^{\mK} + \mR_k^{\mK} 
\\
\mW_{k+1}-\mW_k &= \tfrac{1}{n} \mJ^{\mW}(\mA_k, \mK_k, \mW_k)+ 
\mM_k^{\mW} + \mR_k^{\mW},
\end{align*}
in which 
$\mM_k^{\mA} = \mat \Delta_k^{\mA} - \EEk \mat \Delta_k^{\mA}$,
$\mM_k^{\mK} = \mat \Delta_k^{\mK} - \EEk \mat \Delta_k^{\mK}$,
 and
 $\mM_k^{\mW} = \mat \Delta_k^{\mW} - \EEk \mat \Delta_k^{\mW}$.
 Here,
\begin{align} \label{eq:def-DA}
\mat \Delta_k^{\mA} &\bydef \tfrac{1}{n \a} \mW_k^{-1}\vq_k \vq_k^\T \mW_k^{-1} - \mu \mA_k 
\\ \label{eq:def-DK}
\mat \Delta_k^{\mK} &\bydef  \tfrac{1}{n\a} \big( \t \vq_k \vq_k - \mK_k \mW_k^{-1} \vq_k \vq_k^\T \big) \mW_k^{-1} \mA_k^{-1}
\\ \label{eq:def-DW}
\mat \Delta_k^{\mW} &\bydef \tfrac{1}{n\a} 
\mA_k^{-1} \mW_k^{-1} \vq_k \vq_k^\T \mW_k^{-1} \mA_k^{-1}.
\end{align}
The remainder terms $\mR_k^{\mA}$, $\mR_k^{\mK}$ and $\mR_k^{\mW}$ are
\begin{align*}
\mR_k^{\mA} \bydef & 
\mA_{k+1} - \mA_k - \Delta_k^{\mA}  \\
&+ \EEk  \Delta_k^{\mA} - \tfrac{1}{n}
\mJ^{\mA}(\mA_k, \mK_k, \mW_k) 
\\
\mR_k^{\mK} \bydef &
\mK_{k+1} - \mK_k - \Delta_k^{\mK}  \\
&+ \EEk  \Delta_k^{\mK} - \tfrac{1}{n}
\mJ^{\mK}(\mA_k, \mK_k, \mW_k) 
\\
\mR_k^{\mW} \bydef &
\mW_{k+1} - \mW_k - \Delta_k^{\mW}  \\
&+ \EEk  \Delta_k^{\mW} - \tfrac{1}{n}
\mJ^{\mW}(\mA_k, \mK_k, \mW_k). 
\end{align*}

It is sufficient to show that Condition \ref{c:m} holds by proving
\[
\EE \big[ \norm{\mM^{\mA}_k}^2 + \norm{\mM^{\mK}_k}^2 + \norm{\mM^{\mW}_k}^2 \big]
\leq C n^{-\frac{3}{2}},
\]
and Condition \ref{c:r} holds by proving
\[
\EE \big[ \norm{\mR^{\mA}_k} + \norm{\mR^{\mK}_k}+ \norm{\mR^{\mW}_k} \big]
\leq C n^{-\frac{3}{2}}.
\]

In what follows, we  prove that 
\begin{align}
\EE \norm{ \mM_k^{\mA} }^2 & \leq C n^{-2}  \label{eq:ineq-MA}\\
\EE \norm{ \mR_k^{\mA} } & \leq C n^{-\frac{3}{2}} \label{eq:ineq-MR},
\end{align}
and establish the counterpart inequalities for $\mK_k$ and $\mW_k$. 

We first note that \eqref{eq:ineq-MA} is straightforward from Lemma~\ref{lem:pet-bound-vars}, because
\begin{align*}
&\EE \norm{ \mM_k^{\mA} }^2
\leq \EE \norm{ \mat \Delta_k^{\mA} }^2
\\
&\leq
\tfrac{2}{n^2 \a^2} \EE \norm{\mW_k^{-\frac{1}{2}}}^4 \cdot
\norm{\mW_k^{-\frac{1}{2}} \vq_k }^4
+\tfrac{2 \mu^2}{n^2} \norm{\mA_k}^2 
\\
&\leq C(T) n^{-2}.
\end{align*}

Next, we prove \eqref{eq:ineq-MR}.
\begin{equation}
\begin{aligned}
\EE \norm{\mR_k^{\mA}}
\leq& 
\EE
\norm{\EEk \mat \Delta_k^{\mA} - \tfrac{1}{n}\mJ^{\mA} (\mA_k, \mK_k, \mW_k) }
\\ &+ 
\EE \norm{\mA_{k+1} - \mA_k - \mat \Delta_k^{\mA} }.
\end{aligned} \label{eq:bound-RA}
\end{equation}
The first term is bounded by
\begin{align*}
&\EE
\norm{\EEk \mat \Delta_k^{\mA} - \tfrac{1}{n}\mJ^{\mA} (\mA_k, \mK_k, \mW_k) }
\\
&\leq 
\tfrac{1}{n}\EE 
\Big \Vert
 \mW_k^{-1} 
\big[
\tfrac{1}{\a} \EEk \vq_k \vq_k^\T 
\\
& \quad\quad \quad\quad -
\a \mK_k^\T \mat \Lambda^2 \mK_k - \s^2 \mW_k
\big]
\mW_k^{-1}
\Big \Vert
\\
&\leq \tfrac{C}{n} \EE 
\Big \Vert
\tfrac{1}{\a} \EEk \vq_k \vq_k^\T - 
\a \mK_k^\T \mat \Lambda^2 \mK_k - \s^2 \mW_k
\Big \Vert
\\
&\leq C(T)n^{-\frac{3}{2}}.
\end{align*}
In reaching the second inequality, we used the fact from Lemma~\ref{lem:bound-W} that 
\[
\norm{\mW_k^{-1}}\leq \norm{ \mW_0^{-1}} \leq C,
\]
and the last line is due to \eqref{eq:XssX}.

From \eqref{eq:LA} and \eqref{eq:def-DA},  the second term on the right hand side of \eqref{eq:bound-RA} is bounded by
\begin{align}\nonumber
&
\EE \norm{\mA_{k+1} - \mA_k - \mat \Delta_k^{\mA} }
 \\ \nonumber
& \leq 
\tfrac{1}{n} 
\EE \norm{ \big(
\a \mZ_k^{-1} \vq_k \vq_k^\T \mZ_k^{-1}  
 - \a^{-1} \mW_k^{-1} \vq_k \vq_k^\T \mW_k^{-1} \big) 
\charfn_{\mathcal{A}_k}
}
\\
&\quad + \tfrac{1}{n} \EE
\norm{ \big( \a^{-1} \mW_k^{-1} \vq_k \vq_k^\T \mW_k^{-1} - \mu \mA_k \big) \big( 1- \charfn_{\mathcal{A}_k} \big)}, \label{eq:bound-AAD}
\end{align}
where $\mZ_k$ and $\vq_k$ are  defined in \eqref{eq:def-k}.
The first term in \eqref{eq:bound-AAD} is bounded by
\begin{align} \nonumber
&
\EE \norm{ \big(
\a \mZ_k^{-1} \vq_k \vq_k^\T \mZ_k^{-1}  
 - \a^{-1} \mW_k^{-1} \vq_k \vq_k^\T \mW_k^{-1} \big) 
\charfn_{\mathcal{A}_k} }
\\ \nonumber
& 
\leq \EE \Big \Vert
( \mZ_{k}^{-1} \vq_k \vq_k^\T \mZ_k^{-1}  
(\a \mW_k - \mZ) \mW^{-1}_k
\charfn_{\mathcal{A}_k}
 \Big \Vert
 \\ \nonumber
 & \quad
 + \EE 
 \Big \Vert
\mZ_k^{-1} (\a \mW_k - \mZ_k) 
\mW_k^{-1} \vq_k \vq_k^\T \mW_k^{-1}
 \charfn_{\mathcal{A}_k}
 \Big \Vert
 \\ \nonumber
 & \leq C \EE \norm{\mZ_k^{-\frac{1}{2}} \vq_k \vq_k^\T \mZ_k^{-\frac{1}{2}}  }
 \cdot \norm{\a \mW_k - \mZ_k}
 \\ \nonumber
 & \quad + C \EE  \norm{\a \mW_k - \mZ_k} \cdot
  \norm{\mW_k^{-\frac{1}{2}} \vq_k \vq_k^\T \mW_k^{-\frac{1}{2}}  \charfn_{\mathcal{A}_k} }
 \\ \nonumber
 & \leq C \Big( \EE \norm{\a \mW_k - \mZ_k}^2 \Big)^{\frac{1}{2}  } 
 \\
 & \leq Cn^{-\frac{1}{2}}, \label{eq:bound-AA1}
\end{align}
where in reaching the  second  last line, we used Lemma~\ref{lem:pet-bound-vars} and  H\"older's inequality, 
and the last line is due to \eqref{eq:XX}.
The second term on the right hand side of \eqref{eq:bound-AAD} is bounded by
\begin{align} \nonumber
&\EE
\norm{ \big( \a^{-1} \mW_k^{-1} \vq_k \vq_k^\T \mW_k^{-1} - \mu \mA_k \big) \big( 1- \charfn_{\mathcal{A}_k} \big)}
\\ \nonumber
& \leq \a^{-1} 
 \Big[
 \EE \norm{\mW_k^{-1} \vq_k \vq_k^\T \mW_k^{-1}}^2
 \EE\big( 1- \charfn_{\mathcal{A}_k} \big)^2 
 \Big]^{\frac{1}{2}}
 \\ \nonumber
 & \quad \quad
  + \mu 
   \Big[
  \EE \norm{\mA_k}^2 
  \EE\big( 1- \charfn_{\mathcal{A}_k} \big)^2 
  \Big]^{\frac{1}{2}}
\\ \nonumber
&\leq C(T) \big[ \EE\big( 1- \charfn_{\mathcal{A}_k} \big)^2 \big]^{\frac{1}{2}}
\\ \nonumber
& \leq C(T)  \big[ 1 - \EE \charfn_{\mathcal{A}_k} \big]^{\frac{1}{2}}
\\
& \leq C(T)n^{-\frac{1}{2}},  \label{eq:bound-AA2}
\end{align}
where in reaching the second inequality, we used Lemma~\ref{lem:pet-bound-vars}.
Now, we finish the proof of \eqref{eq:ineq-MR}.

Next, we prove that
\begin{equation}\label{eq:ineq-MK}
\begin{aligned}
\EE \norm{ \mM_k^{\mK} }^2 & \leq C n^{-2}  \\
\EE \norm{ \mR_k^{\mK} } & \leq C n^{-\frac{3}{2}}.
\end{aligned}
\end{equation}

Similar to the proof of \eqref{eq:ineq-MR},  establishing \eqref{eq:ineq-MK} is straightforward.
\begin{align} \nonumber
&\EE \norm{ \mM_k^{\mK} }^2 
\leq \EE \norm{ \mat \Delta_k^{\mK} }^2 
\\ \nonumber
&\leq 
\tfrac{1}{n^2 \a^2} \EE \norm{
\big( \t \vq_k \vq_k^\T - \tfrac{1}{\a} \mK_k \mW_k^{-1}
\vq_k \vq_k^\T \big) \mW^{-1}_k \mA_k^{-1} }
\\ \nonumber
&\leq
\tfrac{1}{n^2 \a^2} 
\EE \norm{ \t \vq_k} \cdot \norm{\vq_k^\T \mW_k^{-\frac{1}{2}} } \cdot
\norm{  \mW_k^{-\frac{1}{2} } } \cdot
\norm{\mA_k^{-1} }
\\ \nonumber
&\quad + \tfrac{1}{n^2 \a^2}
\EE \norm{\mK_k \mW_k^{-\frac{1}{2}} } \cdot
\norm{\mW_k^{-\frac{1}{2}} \vq_k }^2 \cdot
\norm{  \mW_k^{-\frac{1}{2} } } \cdot
\norm{\mA_k^{-1} }
\\ \nonumber
&\leq C(T)n^{-2}.
\end{align}
In reaching the last line we used Lemma~\ref{lem:pet-bound-vars}.
Following the same strategy as what did for  \eqref{eq:ineq-MR}, we can prove the second inequality in \eqref{eq:ineq-MK}.
Specifically, we first decompose $\EE \norm{ \mR_k^{\mK} } $ into two terms
\begin{equation}
\begin{aligned}
\EE \norm{ \mR_k^{\mK} }
\leq &
\EE
\norm{\EEk \mat \Delta_k^{\mK} - \tfrac{1}{n}\mJ^{\mK} (\mA_k, \mK_k, \mW_k) }
\\
&+ 
\EE \norm{\mK_{k+1} - \mK_k - \mat \Delta_k^{\mK} } 
\end{aligned} \label{eq:bound-RK}
\end{equation}
The first term is bounded by
\begin{align*}
&\EE
\norm{\EEk \mat \Delta_k^{\mK} - \tfrac{1}{n}\mJ^{\mK} (\mA_k, \mK_k, \mW_k) }
\\
& \leq
\tfrac{1}{n}
\EE  \Big[ \norm{\tfrac{1}{\a}\EEk \t \vq_k \vq_k^\T - \big( \s^2 \mI_d + \a \mat \Lambda^2 \big)\mK_k }
\cdot \norm{\mW_k^{-1}} \cdot  \norm{\mA_k^{-1}} 
\\
&\quad 
+
 \norm{\mK_k \mW_k^{-\frac{1}{2} }} \cdot 
\norm{\mW_k^{-\frac{1}{2}}} \cdot
 \norm{\tfrac{1}{\a} \EEk  \vq_k \vq_k^\T - \big( \s^2 \mI_d + \a \mat \Lambda^2 \big)}
 \\
 & \quad \quad  \;\;
  \norm{\mW_k^{-1}} \cdot  \norm{\mA_k^{-1}} \Big]
 \\
 & \leq C(T)n^{-1}.
\end{align*}
The second term on the right-hand side of \eqref{eq:bound-RK} can be bounded by
\begin{align*}
&\EE \norm{\mK_{k+1} - \mK_k - \mat \Delta_k^{\mK} }
\\
&
= \tfrac{1}{n}
\EE 
\Big \Vert
 \big( \t \vq_k \vq_k^\T - \t \mQ_k \mZ_k^{-1} \vq_k \vq_k^\T \big) \mZ_k^{-1} \mA^{-1}
\charfn_{\mathcal{A}_k} \\
&\quad \quad
- \tfrac{1}{\a} \big( \t \vq_k \vq_k - \t \mK_k \mW_k^{-1} \vq_k \vq_k^\T \big) \mW_k^{-1} \mA_k^{-1}
\Big \Vert
\\
&\leq
\tfrac{1}{n}
\EE 
\Big \Vert
 \big( \t \vq_k \vq_k^\T - \t \mQ_k \mZ_k^{-1} \vq_k \vq_k^\T \big) \mZ_k^{-1} \mA^{-1}
\charfn_{\mathcal{A}_k} \\
&\quad \quad
- \tfrac{1}{\a} \big( \t \vq_k \vq_k - \mK_k \mW_k^{-1} \vq_k \vq_k^\T \big) \mW_k^{-1} \mA_k^{-1}\charfn_{\mathcal{A}_k} 
\Big \Vert
\\
&\quad \quad
+  \tfrac{1}{\a n}\EE \norm{
\big( \t \vq_k \vq_k -  \mK_k \mW_k^{-1} \vq_k \vq_k^\T \big) \mW_k^{-1}
(1-\charfn_{\mathcal{A}_k} )
}
\\
& \leq 
C(T)n^{-2}.
\end{align*}
In reaching the last line, we used Lemmas~\ref{lem:pet-bound-vars} and \ref{lem:pet-conv-vars}.
The details are similar to \eqref{eq:bound-AA1} and \eqref{eq:bound-AA2}.
Now, we proved \eqref{eq:bound-RK}.

Finally, we  prove that
\begin{align}
\EE \norm{ \mM_k^{\mW} }^2 & \leq C n^{-2}  \label{eq:ineq-MW}\\
\EE \norm{ \mR_k^{\mW} } & \leq C n^{-\frac{3}{2}} \label{eq:ineq-RW}.
\end{align}

Establishing \eqref{eq:ineq-MW} follows the way as what we did in proving \eqref{eq:ineq-MA} and \eqref{eq:ineq-MK}. Thus, we omit the details here.

We next prove \eqref{eq:ineq-RW}.
Note that 
\begin{equation}
\begin{aligned}
\mW_{k+1} - \mW_{k} =& 
\tfrac{1}{n}\t z_k \mA_k^{-1} \mZ_k^{-1} \vq_k \vq_k^\T \mZ_k^{-1} \mA^{-1}
\charfn_{\mathcal{A}_k} 
\\
& - \tfrac{1}{n^2} \mA_k^{-1} \mZ_k^{-1} \vq_k \vq_k^\T \mZ_k^{-1} 
\vq_k \vq_k^\T \mZ_k^{-1} \mA_k^{-1}\charfn_{\mathcal{A}_k} 
\end{aligned}
\end{equation}
Same as what we did in proving 
\eqref{eq:ineq-MA} and \eqref{eq:ineq-MK}, we decompose  $\EE \norm{ \mR_k^{\mW} }$ into two part:
\begin{equation}
\begin{aligned}
\EE \norm{ \mR_k^{\mW} }
\leq &
\EE
\norm{\EEk \mat \Delta_k^{\mW} - \tfrac{1}{n}\mJ^{\mW} (\mA_k, \mK_k, \mW_k) }
\\
&+ 
\EE \norm{\mW_{k+1} - \mW_k - \mat \Delta_k^{\mW} } 
\end{aligned} \label{eq:bound-RWW}
\end{equation}
The first term is bounded by Lemmas~\ref{lem:pet-bound-vars} and \ref{lem:pet-conv-vars}.
Specifically, we have
\begin{align*}
&\EE
\norm{\EEk \mat \Delta_k^{\mW} - \tfrac{1}{n}\mJ^{\mW} (\mA_k, \mK_k, \mW_k) }
\\
&\leq
\tfrac{\s^2}{n\a}\EE 
\Big \Vert
\mA_k^{-1} \mW^{-1}_k 
\big( 
\a \mK^\T_k \mat \Lambda^2 \mK_k + \s^2 \mW_k - \EE \vq_k \vq_k^\T
\big) \\
& \quad \quad\quad \quad
\mW_k^{-1} \mA_k^{-1} 
\Big \Vert
\\
& \leq C(T)n^{-2}.
\end{align*}

The second term on the right-hand sides of \eqref{eq:bound-RWW} can be bounded by
\begin{align*}
&\EE \norm{\mW_{k+1} - \mW_k - \mat \Delta_k^{\mW} } 
\\
& \leq
\tfrac{1}{n} \EE
\Big \Vert
\big (
\t z_k \mA_k^{-1} \mZ_k^{-1} \vq_k \vq_k^\T \mA_k^{-1}
\\
&\quad\quad
- \tfrac{1}{\a}\s^2 \mA_k^{-1} \mW_k^{-1} \vq_k \vq_k^\T \mW_k^{-1} \mA_k^{-1}
\big)\charfn_{\mathcal{A}_k}  
\Big \Vert
\\
& \quad
+ \tfrac{1}{n^2} \EE
\norm{\mA_k^{-1} \mZ_k^{-1} \vq_k \vq_k^\T \mZ_k^{-1} \vq_k \vq_k^\T
\mZ_k^{-1} \mA_k^{-1} \charfn_{\mathcal{A}_k}
}
\\
& \leq C n^{-2}.
\end{align*}
Now we complete the proof of Condition \ref{c:m} and \ref{c:r}.

Using  Lemma~\ref{thm:weak}, we can prove that $\mA^{(n)}(t)$,
$\mK^{(n)}(t)$, and $\mW^{(n)}(t)$ converge weakly to the unique solution of the ODE \ref{eq:ODE-pet}. In addition, noting that $\mQ^{(n)}(t)=\mK^{(n)}(t) \big( \mW^{(n)}(t) \big)^{-\frac{1}{2} } $, it is straightforward to show that $\mQ^{(n)}(t)$ converges weakly to $\mQ(t)$, where
$\mQ(t)=\mK(t) \big( \mW(t) \big)^{-\frac{1}{2} } $. Thus, we proved Theorem \ref{thm:limit}.

Using Lemma~\ref{thm:gen-finite}, we can prove that 
$\norm{\mK^{(n)}(t)- \mK(t)} \leq \frac{C(T)}{\sqrt{n}}$ 
and 
$\norm{\mW^{(n)}(t)- \mW(t)} \leq \frac{C(T)}{\sqrt{n}}$.
Noting that $\mK(t)$, $\mW^{-\frac{1}{2}}(t)$ and $\big(\mW^{(n)}(t) \big)^{-\frac{1}{2}}$ are uniformly bounded, we have
\begin{align*}
&
\EE \norm{\mQ^{(n)} (t) - \mQ(t) }
\\
&
\leq 
\EE \norm{\big(\mK^{(n)}(t)- \mK(t)  \big) \big(\mW^{(n)}(t) \big)^{-\frac{1}{2}} }
\\
&\; + \EE \norm{ \mK(t)  \mW^{-\frac{1}{2}}(t)
\big[  \mW^{\frac{1}{2}} (t) - \big( \mW^{(n)}(t) \big)^\frac{1}{2} \big]   \big(\mW^{(n)}(t) \big)^{-\frac{1}{2}}}
\\
& \leq C(T) \EE \norm{ \mK^{(n)}(t)- \mK(t) } \\
&\quad 
+ C(T)\EE
\norm{   \mW^{\frac{1}{2}}(t) - \big(\mW^{(n)}(t) \big)^{\frac{1}{2}} }
\\
&\leq \tfrac{C(T)}{\sqrt{n}}.
\end{align*}
Now, we proved Theorem \ref{thm:finite} for the simplified PETRELS.

\section{Proofs of the Deterministic Limit of Stochastic Processes}
\label{sec:gen-proof}
In this section, we provide the proofs of Lemmas \ref{thm:weak} and \ref{thm:gen-finite} claimed in Section \ref{sec:gen-conv}.
 
 We first prove Lemma \ref{thm:gen-finite}.
 \begin{IEEEproof}[Proof of Lemma \ref{thm:gen-finite}]
The proof utilizes the coupling trick \cite{Bossy1997}.

We first define a  stochastic process $\t \vq_k^{(n)}$  that is coupled with the original process $\vqn_k$ as
\begin{equation} \label{eq:def-tvq}
\tvqn_{k+1}= \tvqn_k + \tfrac{1}{n} L(\tvqn_k) + \vmn_k, \mbox{ with } \tvqn_0 = \vq_0.
\end{equation}
Compared with the stochastic process $\vqn_k$, the process $\tvqn_k$ has two different points.
One is that the initial state $\tvqn_0$  is a deterministic vector. 
Another one is that the remainder term $\vrn_k$ is removed.
The two process $\vqn_k$ and $\tvqn_k$ are coupled via the same martingale term $\vmn_k$.

Then, we can show that 
\begin{equation} \label{eq:tvq}
\EE \norm{\vqn_k - \tvqn_k} \leq C(T)n^{-\min\{\e_2, \e_3\}}.
\end{equation} 
The proof is the following. From \eqref{eq:q-decomp} and \eqref{eq:def-tvq}, we have
\begin{align*}
&\EE \norm{\vqn_{k+1} - \tvqn_{k+1}} \\
&\leq  \EE  \norm{\vqn_{k} - \tvqn_{k}} 
+ \tfrac{1}{n} \EE  \norm{L(\vqn_{k}) - L(\tvqn_{k}) }
 + \EE  \norm{\vrn_k}
\\
&\leq (1+\frac{C}{n}) \norm{\vqn_{k} - \tvqn_{k}} + C(T)/n^{1+\e_2},
\end{align*}
where in reaching the last line we used Conditions \ref{c:r} and \ref{c:l}.
Knowing  $\EE \norm{\vq^{(n)}_0 -\vq_0} \leq C/n^{\e_3}$, and we can prove \eqref{eq:tvq} using Lemma \ref{lem:itr}.

Next, we define a deterministic process $\hvqn_k$
\begin{equation}\label{eq:def-hvq}
\hvqn_{k+1} = \hvqn_k + \tfrac{1}{n}L(\hvqn_k) \mbox{ with } \hvqn_0=\vq_0.
\end{equation}
And, we can show that 
\begin{equation} \label{eq:hvq}
\EE \norm{\tvqn_k - \hvqn_k}^2 \leq C(T)n^{-\e_1}.
\end{equation} 
The proof is the following.
From \eqref{eq:def-tvq} and \eqref{eq:hvq}, we have
\begin{align*}
&\EE \norm{\tvqn_{k+1} - \hvqn_{k+1}}^2 \\
&= \EE \norm{\tvqn_{k} - \hvqn_{k}}^2 + \tfrac{1}{n^2} \EE \norm{L(\tvqn_k) - L(\hvqn_k)  }^2 
\\
&\quad + \tfrac{2}{n}\EE  \big(L(\tvqn_k) - L(\hvqn_k) \big)^\T (\tvqn_{k} - \hvqn_{k}) + \EE \norm{\vmn_k}^2\\
&\leq (1+\tfrac{C}{n}) \norm{\tvqn_{k} - \hvqn_{k}}^2 + C(T)/n^{1+\e_1},
\end{align*}
where in reaching the second line, we applied the independent increment property of a martingale to $\vmn_k$, 
and in reaching the last line, we used Condition \ref{c:m} and \ref{c:l}.

Since $\hvqn_k$ is simply a finite-difference approximation of the ODE \eqref{eq:ode-q} with a step size $\frac{1}{n}$, it is straightforward that
\begin{equation} \label{eq:hq-vq}
\norm{\hvqn_k - \vq(\tfrac{k}{n})} \leq C(T)/n.
\end{equation}

Finally, combining \eqref{eq:tvq}, \eqref{eq:hvq} and \eqref{eq:hq-vq}, we have
\begin{align*}
&\EE \norm{\vqn_k - \vq(\tfrac{k}{n})} \\
&\leq \EE \norm{\vqn_k -\tvqn_k} + \EE \norm{\tvqn_k - \hvqn_k} +  \norm{\hvqn_k - \vq(\tfrac{k}{n})} \\
&\leq \EE \norm{\vqn_k -\tvqn_k} +  \sqrt{\EE \norm{\tvqn_k - \hvqn_k}^2} \\
&\quad+  \norm{\hvqn_k - \vq(\tfrac{k}{n})} \\
&\leq C(T)/\sqrt{n}.
\end{align*}
\end{IEEEproof}
 
 \begin{IEEEproof}[Proof of Lemma \ref{thm:weak}]
The proof contains three steps: First we show that any 
sequence of random processes $\{\{ \vq^{(n)}(t)\}_{0\leq t \leq T} \}_{n=1,2,\ldots}$ 
indexed by $n$ is tight. 
The tightness indicates that any sequence must contain a 
converging sub-sequence. Second, we prove that  any converging (sub)-sequence 
$\{\{ \vq^{(n)}(t)\}_{0\leq t \leq T} \}_{n=1,2,\ldots}$ converges weakly to a 
solution of the ODE \eqref{eq:ode-q}. Third,  Condition \ref{c:l} implies the ODE  
\eqref{eq:ode-q} has the unique solution.
Combining these three steps, we conclude that any sequence of 
$\{\{ \vq^{(n)}(t)\}_{0\leq t \leq T} \}_{n=1,2,\ldots}$ must converge to the unique solution of the ODE \eqref{eq:ode-q}.
 
We first prove  that the sequence $\{\{ \vq^{(n)}(t)\}_{0\leq t \leq T} \}_{n=1,2,\ldots}$ is tight
 in $D(\R^d,[0,T])$, where $D$ is the space of c\'adl\'ag process.  
 According to Billingsley \cite[Theorem 13.2, pp.\,139 - 140]{billingsley2013convergence} (with a slight extension to $D(\R^d, [0,T])$ from 
 $D(\R, [0,T])$ using $L2$ metric in $\R^d$), this
is equivalent to checking the following two conditions.

1. $ \lim_{b \to \infty} \limsup_{n} \mathbb{P}( \sup_{t \in[0,T]} \norm{ \vq^{(n)}(t) } \ge b)=0$,
and

2. for each $ \epsilon$, $ \lim_{ \delta \to0} \limsup_{n} \mathbb{P}( \omega_{n}^{ \prime}( \delta; \vqn(\cdot)) \ge \epsilon)=0$. 
Here, $ \omega_{n}^{ \prime}( \delta;\vqn(\cdot) )$ is the modulus of continuity
of the function $\vqn(t)$,
defined as 
\[
\omega_{n}^{ \prime}( \delta) \bydef \inf_{ \{t_{i} \}} \max_{i} \sup_{s,t \in[t_{i-1},t_{i})} \norm{ \vq^{(n)}(t)- \vq^{(n)}(s)},
\]
 where $ \left \{ t_{i} \right \} $ is a partition of $ \left[0,T \right]$
such that $ \min_{i} \{t_{i}-t_{i-1} \} \geq \delta$.

The first condition in the tightness criterion is ensured by Condition \ref{c:bound}.

For the second one,
we can prove it by using the uniform partition $ \{t_{i} \}$. 
Let $ \{t_{i} = \frac{iT}{K} \}_{0 \leq i \leq K}$ be a
uniform partition of the interval $[0,T]$. 
Since $\norm{ \vq^{(n)}(t)- \vq^{(n)}(s)} =\norm{ \sum_{k= \ns+1}^{ \nt}\vq_{k}^{(n)} }$,
and \eqref{eq:q-decomp} we only need to prove  that  
%
\begin{align*}
&\lim_{K \to \infty} \limsup_{n \to \infty} \PP \biggl( \max_{1<i \leq K} \sup_{t,s \in[t_{i-1},t_{i}]} 
\norm{ \sum_{k= \ns}^{ \nt-1}\vr_{k}^{(n)} } 
\geq \frac{\e}{3} \biggr)=0
\\
&\lim_{K \to \infty} \limsup_{n \to \infty} \PP \biggl( \max_{1<i \leq K} \sup_{t,s \in[t_{i-1},t_{i}]} 
\norm{ \sum_{k= \ns}^{ \nt-1}\vm_{k}^{(n)} } 
\geq \frac{\e}{3} \biggr)=0
\\
&\lim_{K \to \infty} \limsup_{n \to \infty} \PP \biggl( \max_{1<i \leq K} \sup_{t,s \in[t_{i-1},t_{i}]} 
\norm{ \sum_{k= \ns}^{ \nt-1}\frac{L( \vq_{k}^{(n) }) }{n}} 
\geq \frac{\e}{3} \biggr) \nonumber \\
&=0
\end{align*}
The first equation  trivially holds because of Condition \ref{c:r}, since $\sum_{k= \ns}^{ \nt-1} \norm{\vr_{k}^{(n)}} \leq C/n^{\e_2}$.
Based on  Lemma \ref{lem:tight-det}, which is stated after this proof,
and Conditions \ref{c:m} and \ref{c:init},  the second and third equations also hold.
Combining these three terms we finish the proof of tightness.

Next, we prove that any converging subsequence must converge to the solution of the ODE \eqref{eq:ode-q}.
It is sufficient to show that
\begin{align*}
& \EE \norm{\vqn(t)-\vqn(0) - \int_0^t L(\vqn(s)) ds}\\
&\leq \EE \norm{\vqn_{\nt}-\vqn_{0} - \tfrac{1}{n} \sum_{k=1}^{\nt} L(\vqn_k)}  \\
&=  \EE \norm{ \sum_{k=1}^{\nt} \vmn_{k}
+ \sum_{k=1}^{\nt} \vrn_{k} }
\\
&\leq \Big( d\EE \norm{ \sum_{k=1}^{\nt} {\vmn_k} }^2 \Big)^{1/2}
+ \sum_{k=1}^{\nt} \norm{ \vrn_{k} }
\\
&=
\Big( 
\sum_{k=1}^{\nt} \EE  \norm{\vmn_k}^2
\Big)^{1/2}
+ \sum_{k=1}^{\nt} \norm{ \vrn_{k} }
\\
&\leq Cn^{-\min\{ \epsilon_1/2, \epsilon_2\}}.
\end{align*}

And finally, as mentioned, Condition \ref{c:l} that $L(\vq)$ is a Lipschitz function guarantee that 
the uniqueness of the solution of the ODE \eqref{eq:ode-q}.  
 \end{IEEEproof}

\begin{lemma}
\label{lem:tight-det}Let $(\vz_{k}^{(n)})_{k \geq0}$ be a $d$-dimensional discrete-time stochastic
process parametrized by $n$ and let $ \{t_{i} = \frac{iT}{K} \}_{0 \leq i \leq K}$ be a
uniform partition of the interval $[0,T]$. If $\EE \norm{\vz_k^{(n)}}^2 \le C(T) n^{-2}$ for all $k \le nT$, then for any $ \epsilon>0$, we have 
\[
\lim_{K \to \infty} \limsup_{n \to \infty} \PP \biggl( \max_{1<i \leq K} \sup_{t,s \in[t_{i-1},t_{i}]} \norm{ \sum_{k= \ns}^{ \nt-1}\vz_{k}^{(n)} } \geq \e \biggr)=0.
\]
\end{lemma}
This is a simple extension of Lemma 8 in \cite{Wang2017c} from a 1-dimensional process to a $d$-dimensional process.
\begin{IEEEproof}
It follows from Markov's inequality that
\begin{align}
&\PP \biggl( \max_{1<i \leq K} \sup_{t,s \in[t_{i-1},t_{i}]} \norm{ \sum_{k= \ns}^{ \nt-1}\vz_{k}^{(n)} } \geq \e \biggr)\nonumber\\
 &\qquad \leq \frac{1}{ \e} \EE \max_{1<i \leq K} \sup_{t,s \in[t_{i-1},t_{i}]} \norm{ \sum_{k= \ns}^{ \nt-1}\vz_{k}^{(n)}} \leq \frac{1}{ \e}M_{n,K,T}, \label{eq:inq-1}
\end{align}
where $M_{n,K,T}= \EE \max_{1<i \leq K} \sum_{k= \left \lfloor nt_{i-1} \right \rfloor }^{ \left \lfloor nt_{i} \right \rfloor -1} \norm{ \vz_{k}^{(n)} }.$

For any positive number $B$, we have
\begin{align}
M_{n,K,T} & \leq\EE\max_{1<i\leq K}\sum_{k=\left\lfloor nt_{i-1}\right\rfloor }^{\left\lfloor nt_{i}\right\rfloor -1}[B+(\,\norm{\vz_{k}^{(n)}}-\norm{\vz_k^{(n)}} \wedge B)]\nonumber\\
 & \leq \frac{nTB}{K}+\EE\max_{1<i\leq K}\sum_{k=\left\lfloor nt_{i-1}\right\rfloor }^{\left\lfloor nt_{i}\right\rfloor -1}(\,\norm{\vz_{(k)}^{(n)}}-\norm{\vz_k^{(n)}} \wedge B) \nonumber\\
 & \leq \frac{nTB}{K}+\sum_{k=1}^{\lfloor nT \rfloor}\EE(\,\norm{\vz_{k}^{(n)}}-\norm{\vz_k^{(n)}} \wedge B).\label{eq:tight_d}
\end{align}
Next, we bound the expectations on the right-hand side of \eref{tight_d} as
\[
\EE(\,\norm{\vz_{k}^{(n)}}-\norm{\vz_k^{(n)}} \wedge B) \le \frac{\EE \norm{\vz_k^{(n)}}^2}{B} \le \frac{C(T)}{Bn^2},
\]
where the second inequality follows from the assumption that $\EE \norm{\vz_k^{(n)}}^2 \le C(T)n^{-2}$. Substituting this bound into \eref{tight_d} gives us
\[
M_{p,K,T} \le \frac{nTB}{K} + \frac{TC(T)}{Bn}.
\]
Choosing $B = \sqrt{K}/n$ and using \eref{inq-1}, we are done.
\end{IEEEproof}

\end{document}